\documentclass[twoside,11pt]{article}
\usepackage{jair, theapa, rawfonts}

\usepackage{amssymb}
\usepackage{amsfonts}
\usepackage{amsmath}
\usepackage{blkarray}
\usepackage{booktabs}
\usepackage{dsfont}
\usepackage{psfrag}
\usepackage{rotating}
\usepackage{url}
\usepackage{bm}
\usepackage{algorithm}
\usepackage{algorithmic}
\usepackage{color,soul}
\usepackage{lscape}
\usepackage{graphicx,epstopdf}

\usepackage{multirow}
\usepackage{microtype}
\usepackage{cleveref}
\usepackage[authoryear]{natbib}

\usepackage{tikz}
\usepackage{tikz-qtree}
\usepackage{array}
\usepackage{subfig}
\usepackage{multirow}

\usepackage{algorithm}
\usepackage{algorithmic}

\usepackage{enumerate}

\usetikzlibrary{fit,positioning}

\newcommand{\transp}{^\top}

\newcommand{\proba}{P}

\newcommand{\complexSpace}{\mathbb{C}}
\renewcommand{\Re}{\mathbb{R}}
\newcommand{\C}{\complexSpace} 
\newcommand{\R}{\Re} 
\newcommand{\real}{\mathrm{Re}}
\newcommand{\imag}{\mathrm{Im}}

\providecommand{\U}[1]{\protect\rule{.1in}{.1in}}
\newcommand{\be}{\begin{equation}}
\newcommand{\ee}{\end{equation}}
\newcommand{\bd}{\begin{definition}}
\newcommand{\ed}{\end{definition}}
\newcommand{\ba}{\begin{algorithm}}
\newcommand{\ea}{\end{algorithm}}
\newcommand{\br}{\begin{problem}}
\newcommand{\er}{\end{problem}}
\newcommand{\bex}{\begin{example}}
\newcommand{\eex}{\end{example}}
\newcommand{\bt}{\begin{theorem}}
\newcommand{\et}{\end{theorem}}

\def\ifa{\iffalse}
\def\ifappendix{\iftrue}

\newcommand{\ObsTensor}{\mathbf{Y}}
\newcommand{\EntitySpace}{\mathcal{E}}
\newcommand{\RelationSpace}{\mathcal{R}}
\newcommand{\TripleSpace}{\mathcal{T}}
\newcommand{\score}{s}

\renewcommand{\cite}{\citep}

\newcommand{\T}{\transp}
\newcommand{\conj}{\overline}

\newtheorem{thm}{Theorem}

\newcommand{\Ne}{N_e} 
\newcommand{\Nr}{N_r} 
\newcommand{\OmegaSpace}{\Omega} 
\newcommand{\sampval}{y_{rso}}  
\newcommand{\setent}{\mathcal{E}} 
\newcommand{\rank}{K}
\newcommand{\rk}{\rank}
\renewcommand{\j}{\mathbf{j}}
\newcommand{\fourm}{{\mathrm{4main}}}
\newcommand{\thiro}{{\mathrm{13other}}}
\newcommand{\train}{{\mathrm{train}}}

\def\tt{\texttt}

\graphicspath{{figures/}}

\usepackage[textsize=small]{todonotes}

\jairheading{42}{1993}{1-33}{4/17}{00/00}
\ShortHeadings{On Inductive Abilities of Latent Factor Models for Relational Learning}
{Trouillon, Gaussier, Dance \& Bouchard}
\firstpageno{1}

\begin{document}

\title{On Inductive Abilities of Latent Factor Models\\for Relational Learning}

\author{\name Th\'eo Trouillon \email theo.trouillon@imag.fr \\
       \addr Univ. Grenoble Alpes, 700 avenue Centrale, 38401 Saint Martin d'H\`eres, France
       \AND
       \name \'Eric Gaussier \email eric.gaussier@imag.fr \\
       \addr Univ. Grenoble Alpes, 700 avenue Centrale, 38401 Saint Martin d'H\`eres, France
       \AND
       \name Christopher R. Dance \email chris.dance@naverlabs.com \\
       \addr NAVER LABS Europe, 6 chemin de Maupertuis, 38240 Meylan, France
       \AND
       \name Guillaume Bouchard \email g.bouchard@cs.ucl.ac.uk \\
       \addr Bloomsbury AI, 115 Hampstead Road, London NW1 3EE, United Kingdom \\
       \addr University College London, Gower St, London WC1E 6BT, United Kingdom
       }

\maketitle
\begin{abstract}
Latent factor models are increasingly popular for modeling
multi-relational knowledge graphs.
By their vectorial nature, it is not only hard to interpret 
why this class of models works so well,
but also to understand where they fail and 
how they might be improved. We conduct an experimental
survey of state-of-the-art models, not towards
a purely comparative end, but as a means to get insight 
about their inductive abilities.
To assess the strengths and weaknesses of each model, we create simple tasks 
that exhibit first, atomic properties of binary relations, 
and then, common inter-relational inference through synthetic genealogies.
Based on these experimental results, we propose new research 
directions to improve on existing models.
\end{abstract}


\section{Introduction}

In many machine learning fields, research is drifting away
from first-order logic methods. Most of the time, this
drift is justified by better predictive performances and scalability of
the new methods. It is especially true with
link prediction, a core problem of
statistical relational learning \cite{Getoor2007},
where latent factor models became more popular than logic-based
models \cite{Nickel2011,bordes2013translating,trouillon2016}.

Link prediction in knowledge graphs---also known as knowledge
graph completion---operates on predicates
of pairs of \emph{entities}: the objects of knowledge graphs.
Each different predicate symbol is called a
\emph{relation}, and a grounded relation is called a \emph{fact}.
For example, given the entities \tt{Alice}, \tt{Eve}
and \tt{Bob} and the relations \tt{mother} and \tt{grandmother},
if \tt{mother(Alice,Eve)} and \tt{mother(Eve,Bob)} are
true facts, then \tt{grandmother(Alice,Bob)} is also true.
Inferring this last fact from the first two however, 
requires knowing that the mother of one's mother is
one's grandmother, which can be expressed by the first-order formula: 
$\mathtt{\forall x \forall y \forall z\,mother(x,y)\wedge mother(y,z) \Rightarrow
grandmother(x,z)}$.

Logic-based link prediction consists in using both observed facts
and logical rules to infer the truth of unobserved facts.
It can be achieved deterministically by logical deduction,
or probabilistically to cope with uncertainty of the data 
\cite{richardson2006markov,kersting2001towards}. 
Beyond known problems such as complexity or brittleness, 
an obvious limitation arises in this setup: 
logical rules over the knowledge graph relations are required
for inference, and many knowledge graphs only provide
observed facts \cite{dong2014_knowledgevault,dbpedia}.
In this case one must either handcraft rules, or learn them, generally
through inductive logic programming (ILP) methods
\cite{muggleton1994inductive,dzeroski1994inductive}.

Latent factor models do not suffer this limitation,
as the learned model is never represented explicitly in a symbolic way,
but rather as vectorial embeddings of the entities and relations. 
Such representations can make the model difficult to interpret,
and although they show better predictive abilities, 
it has not yet been explored how well
those models are able to overcome this absence
of logical rules, and how their inference abilities
differ from logic-based models. 

To do so, we evaluate state-of-the-art 
latent factor models for relational learning on 
synthetic tasks, each designed to target a specific inference ability,
and see how well they discover structure in the data.
As we are only interested in evaluating inductive abilities of 
these models, and not their ability to cope
with uncertainty, we design synthetic experiments with 
noise-free deterministic data. 
The choice of this very favorable setup for deterministic logical inference
clarifies the approach followed in this
paper and its very purpose: we \emph{do not} evaluate latent factor models
as an \emph{end}, but as a \emph{means} to point out their weaknesses
and stimulate research towards models that 
do not suffer from combinatorial complexity---as advocated
by \citet{bottou2014machine}. Computational complexity, and namely
polynomiality, could turn out to be the very criterion
for machine intelligence \cite{aaronson2011philosophers}.
Beyond complexity, one could also argue against
explicitly learning logical expressions to tackle 
knowledge graph completion that,
``when solving a given problem, try to avoid 
solving a more general problem as an intermediate step''
\cite{vapnik1995nature}.

We first evaluate the models on the three main properties of binary
relations: reflexivity, symmetry and transitivity, and their combinations. 
We do so by experimentally testing their ability to learn these patterns 
from facts, and their robustness to missing data.
Then we set up tasks that represent real reasoning over family
genealogies. On this data, we explore different types of training/testing splits
that map to different types of inference.

The remainder of the paper is organized as follows.
We first review the literature in \Cref{sec:related_work}, before
presenting formally the link-prediction task, the evaluated latent
factor models and the optimization procedure in \Cref{sec:lp_task}. 
Experiments on learning properties of relations are presented along
with their results in \Cref{sec:rel_props}, and experiments
description and results for family genealogies are reported in 
\Cref{sec:family_rels}. Finally, we propose
new research directions from these results in \Cref{sec:research_dir}.

\section{Related Work}
\label{sec:related_work}

Artificial intelligence is becoming more driven by its empirical successes
than by the quest for a principled formalisation of reasoning, making it
more of an empirical science than a theoretical one.
Experimental design is a key skill of empirical scientists, and
a well-designed experiment should expose model limitations 
to enable improving on them. Indeed, seeking falsification is
up to now the best definition of science \cite{popper1934}. 
In machine learning, it is extremely simple
to come up with an experiment that will fail. However it is less easy to think
of one that brings an informative failure---when one thinks of 
a failing experiment at all. The bAbI data set
\cite{weston2015towards}, proposing a set of 20 prerequisite tasks for reasoning 
over natural language, is an example of an informative experiment,
by the specific reasoning type that each task targets.
Inspired by the idea of this work, we designed
simple tasks for relational learning that assess basic properties
of relations, as well as simple reasonings such as kinship relations.

Learning from knowledge graphs and more generally relational data
is an old problem of artificial intelligence \cite{davis1993knowledge}.
Many contributions have been made using inductive logic programming
for relational data during the last decades \cite{muggleton1995inverse,lisi2010inductive,galarraga2015fast}.
Handling inference probabilistically gave birth 
to the statistical relational learning field \cite{Getoor2007},
and link prediction has always been one of the main problems
in that field.
Different probabilistic logic-based inference models have
been proposed \cite{ngo1997answering,wellman1992knowledge,kersting2001towards}.
The main contribution along this line of research is probably
Markov Logic Networks (MLNs) \cite{richardson2006markov}.
MLNs take as input a set of first-order rules
and facts, build a Markov random field between facts co-occuring in
possible groundings of the formulae, from which they learn a weight over each 
of these rules that represents
their likeliness of being applied at inference time.
Some other proposals followed a purely probabilistic approach \cite{friedman1999learning,heckerman2007probabilistic}.

The link-prediction problem has recently drawn attention from a wider community. Driven by the W3C standard data representation for the semantic web, the resource description framework \cite{cyganiak2014rdf}, various knowledge graphs---also called knowledge bases---have been collaboratively or automatically 
created in recent years such as DBpedia~\cite{dbpedia}, 
Freebase~\cite{Bollacker2008} or the Google Knowledge Vault~\cite{dong2014_knowledgevault}.
Since the Netflix challenge \cite{koren_netflix},
latent factor models have taken the advantage over probabilistic and symbolic 
approaches in the link-prediction task.
In terms of prediction performances first, but also in scalability. 
This rise of predictive performances and speed 
enabled many applications including
automated personal assistants and recommender systems
\cite{ma2015knowledge,koren2}.

Statistical models for learning in knowledge graphs
are summarized in a recent review
\cite{nickel_2016_review}, and among them latent factor models. 
We discuss these models in detail in the following section.
One notable latent factor model that is not tested in this paper is the holographic
embeddings model \cite{nickel_2016_holographic},
as it has been shown to be equivalent to the \textsc{ComplEx} model \cite{trouillon2017comparison,hayashi2017equivalence}.
The \textsc{ComplEx} model \cite{trouillon2016} is detailed in the next section.
Also, the latent factor model proposed by \citet{Jenatton2012}
is not included as it is a combination of uni-, bi- and trigram 
terms that will be evaluated in separate models to understand
the contribution of each modeling choice in different situations.

Not all latent models are actually factorization models. Among these 
are a variety of neural-network models, including the neural tensor networks
\cite{socher2013reasoning}, or the multi-layer perceptron used
in \citet{dong2014_knowledgevault}. We did not survey these models in
this work and focus on latent \emph{factor} models, that is models
that can be expressed as a factorization of the knowledge graph
represented as a tensor.

Similarly to our approach, \citet{bowman2015recursive}  
learned some \emph{natural logic} operations 
directly from data with recurrent neural tensor networks,
to tackle natural language processing
tasks such as entailment or equivalence. 
Natural logic is a theoretical framework for natural 
language inference that uses natural language strings as the logical symbols.
\citet{bouchard2015approximate} compared the squared and logistic
losses when learning transitive and sequential synthetic relations, 
and \citet{Singh2015} also investigated a few synthetic examples for relational
learning on different latent factor models.

Following a different goal, other approaches formalised
the encoding of logical operations as tensor operations.
\citet{smolensky2016basic} applied it to the bAbI data set
reasoning tasks, and \citet{grefenstette2013towards} to general Boolean operations.

Advances in bringing both worlds together include the
work of \citet{Rocktaschel2015,rocktaschel2014low} and \citet{demeester2016lifted}, 
where a latent factor model is used, as well
as a set of logical rules. An error-term over the rules is added to the classical
latent factor objective function. In \citet{rocktaschel2016learning}, a fully differentiable
neural theorem prover is used to handle both facts and rules, whereas
\citet{minervini2017adversarial} use adversarial training to do so.
\citet{wang2016learning} learned first-order logic embeddings
from formulae learned by ILP.
Similar proposals for integrating logical knowledge in distributional representations
of words include the work of \citet{lewis2013combining}. 
Conversely, \citet{Yang2015} learn a latent factor model
over the facts only, and then try to extract rules from the learned embeddings.
\cite{yoon2016translation} proposed to use projections of the subject and
object entity embeddings that conserve transitivity and symmetry.

\section{The Link-Prediction Task and Models}
\label{sec:lp_task}

This section formally defines the link-prediction problem in knowledge
graphs, as well as the notations that will be used throughout this paper.
We then introduce the state-of-the-art models that will be tested
in the experimental sections.

\subsection{Link-Prediction in Knowledge Graphs}

A knowledge graph stores facts about a set of entities $\EntitySpace$,
and a set of relations $\RelationSpace$, in the form of facts
$r(s,o)$, that we also write as triples $(r,s,o)$,
where the relation $r \in \RelationSpace$ and the subject
and object entities $s,o \in \EntitySpace$.
Each fact is associated with its truth value $\sampval \in \{-1,1\}$.
For example, the fact \tt{first\_used}(\tt{Cardano}, \tt{imaginary\_numbers}) is true \cite{cardano1545}, 
thus it has a corresponding truth value $\sampval = 1$.
To false facts we attribute the value $-1$.
We denote the set of all possible triples for a given entity set
and relation set with $\TripleSpace = \RelationSpace \times \EntitySpace \times \EntitySpace$.

In the link-prediction task we observe the truth values of a given set of 
training triples $\TripleSpace_{\OmegaSpace}\subseteq\TripleSpace$, 
that together form the observed facts set $\OmegaSpace = \{((r,s,o), \sampval) \,|\, (r,s,o) \in \TripleSpace_{\OmegaSpace}\}$. 
The task then consists in predicting the truth values of 
a disjoint set of unobserved triples 
$(r',s',o')\in \TripleSpace \setminus \TripleSpace_{\OmegaSpace}$.

Each model is defined by its scoring function $\phi(r,s,o;\Theta)$,
where $\Theta$ are the latent parameters of this model---the entity
and relation embeddings---and 
$\phi(r,s,o;\Theta): \C^{|\Theta|} \rightarrow \R$ assigns a real-valued
score to the fact $r(s,o)$. As some models are real-valued and some
other models are complex-valued, we define the space of 
the parameters $\C^{|\Theta|}$
directly over the complex space.

We define the probability of a given fact $r(s,o)$ to be true as
\begin{equation}
    \proba(\sampval=1) =  \sigma(\phi(r,s,o;\Theta))
    \label{observation-model}
\end{equation}
where $\sigma$ is a sigmoid function. We here use
the classical logistic function $\sigma(x) = \frac{1}{1+\mathrm{e}^{-x}}$.

In the remainder of the paper, the following notations
will also be used: the number of entities is denoted by $N_e = |\EntitySpace|$,
and the number of relations by $N_r = |\RelationSpace|$.
The $i^{\mathrm{th}}$ row of a complex matrix $X \in \C^{n \times m}$
is written $x_i \in \C^m$. By a slight abuse of notation,
for entities $i \in \EntitySpace$ and relations $r \in \RelationSpace$,
we will write their corresponding rows in the embedding matrices
as $x_i$ or $x_r$, where $x_i, x_r \in \C^m$.

Let us also define the trilinear product of three vectors over 
the complex space:
\begin{eqnarray}
   \left<a,b,c\right> &=& \sum\limits_{j=1}^\rk a_j b_j c_j\notag\\
   &=& a\T (b \odot c)
\end{eqnarray}
where $a,b,c \in \C^\rk$, and $\odot$ is the Hadamard product, 
that is the element-wise product between two vectors of same length.

\subsection{The Models}
\label{sec:models}

In the following we present in detail the model scoring functions and
parameters that we compare in this experimental survey.
We chose to compare only the most popular 
and best-performing link-prediction models. 
The models' scoring functions
and parameters are summarized in \Cref{tab:scoring}.

\begin{table*}[t]
    \centering
    
    
    \begin{tabular}{|m{6.5cm}|m{3.6cm}|m{3.8cm}|@{}m{0pt}@{}}
        \hline
        \textbf{Model} &
        \textbf{Scoring Function} $\phi$&
        \textbf{Parameters} $\Theta$
        &\\[5pt]
        \hline
        \textsc{CP} \cite{hitchcock-sum-1927} &  
        $\left<w_r, u_s, v_o\right>$ &
        $w_r, u_s, v_o \in\Re^{K}$
        &\\[5pt]
        \hline
        \textsc{RESCAL} \cite{Nickel2011} &
        $e_s^T W_r e_o$ &
        $W_r\in\Re^{K^2}$, $ e_s, e_o \in\Re^{K}$
        &\\[5pt]
        \hline
        \textsc{TransE}  \cite{bordes2013translating}&
        $-||(e_s + w_r) - e_o||_q$ &
        $w_r, e_s, e_o \in\Re^{K}$
        &\\[5pt]
        \hline
        F model  \cite{riedel_2013_univschema}&
        $u_d\T w_r $ &
        $w_r, u_d \in \Re^{K}$
        &\\[5pt]
        \hline
        \textsc{DistMult} \cite{Yang2015}&  
        $\left<w_r, e_s, e_o\right>$ &
        $w_r, e_s, e_o \in\Re^{K}$
        &\\[5pt]
        \hline
        \textsc{ComplEx} \cite{trouillon2016} & 
        $\real(\left<w_r, e_s, \bar{e}_o\right>)$ &
        $w_r, e_s, e_o \in\C^{K}$
        &\\[5pt]
        \hline
    \end{tabular}
    \caption{
     Scoring functions of the evaluated latent factor models for a given fact $r(s,o)$, along with the representation of their parameters. In the F model, $d$ indexes all possible pairs of entities: $d = (s,o) \in \EntitySpace \times \EntitySpace$.}
    \label{tab:scoring}
\end{table*}

Each of the following models uses latent representations of variable length,
that is controlled by the hyper-parameter $\rk \in \mathbb{Z}_{++}$.
We start by introducing the most natural model, 
a general decomposition
for tensors: the Canonical-Polyadic (\textsc{CP})
decomposition \cite{hitchcock-sum-1927}, also know as CANDECOMP \cite{cc-candecomp-1970},
and PARAFAC \cite{harshman-parafac-1970}.

\paragraph{Canonical-Polyadic Decomposition (\textsc{CP}) }

A natural way to represent a knowledge graph mathematically
is to frame it as a $3^{\mathrm{rd}}$-order, partially observed, binary tensor
$\ObsTensor \in \{-1,1\}^{\Nr \times \Ne \times \Ne}$,
where the value at index $(r,s,o)$ is  the truth value of
the corresponding triple: $\sampval$.

The Canonical-Polyadic decomposition involves one latent matrix for each dimension
of the decomposed tensor, so in our case we have three latent matrices as $\ObsTensor$
is a $3^{\mathrm{rd}}$ order tensor. 
The dimension of the learned representations $\rk$
is also often called the rank of the decomposition. 
The scoring function is
\begin{eqnarray}
    \phi(r,s,o;\Theta) = \left<w_r, u_s, v_o\right>
\end{eqnarray}
where $U,V \in \mathbb{R}^{\Ne \times \rk}$ are the embedding matrices of entities
depending on whether they appear as subject ($U$) of the triple or as object ($V$),
and $W \in \mathbb{R}^{\Nr \times \rk}$ is the embedding matrix of the relations.

This model is a very general tensor decomposition, though it is not
really tailored to our problem, since our tensor is a stack
of $\Nr$ square matrices where rows and columns 
represent the same underlying objects:
the entities. Indeed, its completely decorrelated representations 
$u_i$ and $v_i$ of the same entity $i \in \EntitySpace$
make it harder for this model to generalize, as we will see in 
the experiments.

\paragraph{\textsc{RESCAL}}

\textsc{RESCAL} \cite{Nickel2011} differs from the CP decomposition in two points:
there is only one embedding per entity instead of having one embedding 
for entities as subject and another one for entities as objects; 
and each relation is represented by a matrix embedding instead of a vector.
Its scoring function is
\begin{eqnarray}
    \phi(r,s,o;\Theta) = e_s\T W_r e_o 
    \label{eq:rescal}
\end{eqnarray}
where $E \in \mathbb{R}^{\Ne \times \rk}$ is the embedding matrix of the entities,
and $W \in \mathbb{R}^{\Nr \times \rk \times \rk}$ the embedding tensor of the relations.
Thus $W_r \in \mathbb{R}^{\rk \times \rk}$ is the embedding matrix of the relation $r$.

\textsc{RESCAL} was the first model to propose unique embeddings for entities---simultaneously
with \citet{bordes2011learning}---which yielded
significant performance improvement, and since then unique entity embeddings 
have been adopted by most
of the subsequent models.
However, its matrix representations of relations makes its scoring
function time and space
complexity quadratic in the rank $\rk$ of the decomposition.
This also leads to potential overfitting.

\paragraph{F model}

This model proposed by \citet{riedel_2013_univschema} maps all possible 
subject and object entity pairs 
$d = (s,o) \in \EntitySpace \times \EntitySpace$ to a single dimension. 
Each row in the entity embedding matrix corresponds to 
one pair of entities. The scoring function computes the dot
product of the embedding of the pair $d$ with the embedding
of the relation $r$:
\begin{eqnarray}
    \phi(r,s,o;\Theta) = e_d\T w_r 
\end{eqnarray}
where $ E \in \mathbb{R}^{\Ne^2 \times \rk}$ is the embedding matrix of the
pairs of entities,
and $W \in \mathbb{R}^{\Nr \times \rk}$ the embedding matrix of the relations.
It is actually a decomposition of the matrix
that results from a specific unfolding of the $\ObsTensor$ tensor.

Its pairwise nature gives this model an advantage over non-compositional
pairs of entities. However, its memory complexity is quadratic in
the number of entities $\Ne$. In practice, unobserved pairs
of entities are not stored in memory as they are useless.
Though this is also the weak point of this model: it
cannot predict scores for unobserved pairs of entities 
since it only learns pairwise representations.

\paragraph{\textsc{TransE}}

The \textsc{TransE} model \cite{bordes2013translating} imposes a geometrical structural bias on the model: the subject entity vector should be close to the object entity vector once translated by the relation vector. For a given $q$-norm (generally
$q=1$ or $q=2$) over the embedding space:
\begin{eqnarray}
    \phi(r,s,o;\Theta) = - ||(e_s + w_r) - e_o||_q
\end{eqnarray}
where $E \in \mathbb{R}^{\Ne \times \rk}$ is the embedding matrix of the entities,
and $W \in \mathbb{R}^{\Nr \times \rk}$ is the embedding matrix of the relations.
Deriving the norm in the scoring function exposes another perspective on the model
and unravels its factorial nature,
as it gives a sum of bilinear terms as explored by \citet{garcia2014effective}:
\begin{eqnarray}
    \phi(r,s,o;\Theta) \approx e_s\T e_o + e_o\T w_r - e_s\T w_r
    \label{eq:transe2}
\end{eqnarray}
where constant multipliers and norms of the embeddings have been ignored here.
These bigram terms will help in some specific situations as
shown in \Cref{sec:family_rels}.

It is difficult to capture symmetric relations with this model in a
multi-relational setting.
Indeed, having $\phi(r,s,o;\Theta) = \phi(r,o,s;\Theta)$ implies either
$e_s = e_o$, or $w_r\T(e_o - e_s) = 0$.
Since $e_s \neq e_o$ in general for $s \neq o$, and $w_r$ is in
general not the zero vector---in order to share latent dimensions' 
information with the
other relation embeddings---modeling symmetric 
relations such as \tt{similar}, \tt{cousin}, or \tt{related}
implies a strong geometrical constraint on entity embeddings:
their difference must be orthogonal to the relation embedding $w_r$.
The model thus has to make a trade-off between (i) correctly modelling
the symmetry of the relation $r$, (ii) not zeroing its relation embedding
$w_r$, and (iii) not altering too much the
entity embeddings to meet the orthogonality requirement between
$w_r$ and $(e_o - e_s)$ for all $e,o \in \EntitySpace$.

\paragraph{\textsc{DistMult}}

The \textsc{DistMult} model \cite{Yang2015} can be seen as 
a simplification of the \textsc{RESCAL} model,
where the unique representation of entities is kept, while the representation
of the relations is brought back to vectors instead of matrices:
\begin{eqnarray}
    \phi(r,s,o;\Theta) = \left<e_s, w_r, e_o\right>
\end{eqnarray}
where $E \in \mathbb{R}^{\Ne \times \rk}$ is the embedding matrix of the entities,
and $W \in \mathbb{R}^{\Nr \times \rk}$ the embedding matrix of the relations.

The major drawback of this model is its symmetry over the subject and object
entity roles. Indeed we have $\phi(r,s,o;\Theta) = \phi(r,o,s;\Theta) $, 
for all $s,o \in \EntitySpace$. But
many antisymmetric relations appear in knowledge graphs such as 
\texttt{older}, \texttt{partOf}, \texttt{hypernym}. One does not
want to assign the same score to \tt{older(a,b)} as to \tt{older(b,a)}!

\paragraph{\textsc{ComplEx}}

The \textsc{ComplEx} model \cite{trouillon2016,trouillon2017knowledge} 
can be seen as a complex-valued
version of the \textsc{DistMult} model. The latent matrices of the entities
and relations are in the complex domain instead of the real domain.
The scoring function is the real part of the trilinear product
of the entities and relation embeddings:

\begin{eqnarray}
    \phi(r,s,o;\Theta) = \real(\left<e_s, w_r, \conj{e}_o\right>)
\end{eqnarray}
where $E \in \C^{\Ne \times \rk}$ is the embedding matrix of the entities,
and $W \in \C^{\Nr \times \rk}$ the embedding matrix of the relations.
We write $\real(a)$ the real part of the complex number $a \in \C$,
and $\imag(a)$ its imaginary part:
$a = \real(a) + i \imag(a)$, where $i = \sqrt{-1}$ and $\real(a), \imag(a)
\in \R$.
Note that the conjugate of the object entity embedding is used, and
introduces asymmetry between the subject and object entities.
We use the same notations for complex vectors:
$\conj{e}_o = \real(e_o) - i \imag(e_o)$, where $\real(e_o), 
\imag(e_o) \in \R^\rk$.

This model solves the symmetry problem of \textsc{DistMult} by having slightly
different representations of entities as subject or object through
the use of the complex conjugate. These representations are still tightly
linked which yields good generalization properties, unlike \textsc{\textsc{CP}}. 
Yet this slight difference allows the model to retain a vectorial representation
of the relations, and thus a linear time and space complexity, 
unlike \textsc{RESCAL}, and to do so without any loss of expressiveness---\textsc{ComplEx} is able 
to decompose exactly all possible knowledge graphs
\cite{trouillon2017knowledge}.

\subsection{Training the Models}

All models have been reimplemented within the same framework
for experimental fairness. We describe their
common optimization scheme in this section.

As previously mentioned, we used the logistic function for $\sigma$,
as it provides better 
results compared to squared error \cite{nickel2013logistic,bouchard2015approximate}.
We minimized the negative log-likelihood with $L^2$ regularization applied
entity-wise and relation-wise over their
vector embeddings of the considered model:
\begin{equation}
    \mathcal{L}(\Omega;\Theta) = \sum_{((r,s,o), y_{rso}) \in \Omega} \log( 1 + \exp(-y_{rso}\phi(r,s,o;\Theta))) + \lambda ||\Theta_{\{r,s,o\}}||^2_2\enspace,
    \label{eq:objective_func}
\end{equation}
as we found it to perform better than the ranking 
loss used in previous studies
\cite{bordes2013translating,nickel_2016_holographic}.
For \textsc{RESCAL}'s relation embeddings $W_r \in \R^{\rank \times \rank}$,
the Frobenius norm is used for regularization $||W_r||_F$.

The loss is optimized through stochastic gradient descent with mini-batches
(10 batches for the relation properties experiment, 
and 100 for the families experiment), AdaGrad \cite{duchi2011adaptive} with
an initial learning rate of $0.1$, and early stopping 
when average precision decreased on the validation set calculated
every 50 epochs.
The $\lambda$ regularization parameter was validated over the values $\{0.3, 0.1, 0.03, 0.01, 0.003, 0.001, 0.0003, 0.0\}$
for each model for each factorization rank $\rk$.
Parameters are initialized from a centered unit-variance Gaussian
distribution. The complete algorithm is detailed in 
\Cref{app:algo}.

Models are evaluated using average precision as it is
classically used in the field \cite{richardson2006markov,Nickel2011}.
We also computed the average precision of a deterministic
logic inference engine, by applying the corresponding rules
that were used to generate each data set.
For each fact $r(s,o)$ in the test set, its probability
$ \proba(\sampval=1) $ is set to 1 if the fact can be logically deduced
true from the facts of the training and validation sets, 0 if it can be deduced to be false, and 0.5 otherwise.

For the \textsc{TransE} model, we trained it with $L^1$ and $L^2$ norms, 
results are reported for the best performing
one in each experiment. As in the original paper, we did not use
regularization over the parameters but instead we enforced
entity embeddings to have unit norm $||e_i||_2 = 1$
for all $i \in \EntitySpace$ \cite{bordes2013translating}.
With the F model, prediction of unobserved entity
pairs in the training set is handled through random Gaussian 
embeddings, similarly to their initialization.

Results are evaluated with average precision,
where for each factorization rank the models with 
best validated $\lambda$ are kept.
Average precisions are macro-averaged over 10 runs, 
and error bars show the standard deviation over these 10 runs.

To assess whether latent factor models are able to generalize
from data without any first-order logic rules, we
generate synthetic data from such rules, and evaluate
the model in a classical training, validation and test splitting
of the data.
The proportion of positives and negatives is respected
across the sets.

We first consider rules corresponding to
relation properties, then rules corresponding 
to inter-relations reasonings
about genealogical data. We also
explore robustness to missing data, as well as different
training/testing splits of the data.
Keeping the data deterministic and simple also allows us 
to write the corresponding logical rules of each experiment,
and simulate test metrics of what perfect induction would yield
to get an upper-bound on the performance of any method.
All data sets are made available\footnote{\url{https://github.com/ttrouill/induction_experiments}}.

\section{Learning Relation Properties}
\label{sec:rel_props}

In this section we define the three main properties of
binary relations, and devise different experimental
setups for learning them individually or jointly,
and with more or less observed data.

\subsection{Experimental Design}
\label{sec:rel_props_design}

Relations in knowledge graphs have different names in the
different areas of mathematics. Logicians call them
binary \emph{predicates}, as they are Boolean-valued functions of
two variables. For set theorists, they are binary \emph{endorelations},
as they operate on two elements of a single set, in our case the
set of entities $\EntitySpace$. In set theory, relations
are characterized by three main properties: reflexivity/irreflexivity, symmetry/antisymmetry and transitivity. 
The definitions of these properties are given in first-order logic
in \Cref{tab:rel_properties}.

Different combinations of these properties define basic building blocks of
set theory such as equivalence relations that are reflexive,
symmetric and transitive relations, or partial orders
that are reflexive, antisymmetric and transitive relations
\cite{halmos1998naive}. Examples are given in \Cref{fig:wiki_bin_rel}.

\begin{table}[h]
\begin{center}
	\begin{tabular}{|r|l|}
		\hline
		Property & Definition \\
		\hline \hline
		Reflexivity & $\forall a\enspace r(a,a)$ \\ \hline
		Irreflexivity & $\forall a\enspace \neg r(a,a)$ \\ \hline
		Symmetry & $\forall a \forall b\enspace r(a,b) \Rightarrow r(b,a)$ \\ \hline
		Antisymmetry & $\forall a \forall b\enspace r(a,b) \wedge r(b,a) \Rightarrow a = b$ \\ \hline
		Transitivity & $\forall a \forall b \forall c\enspace r(a,b) \wedge r(b,c)  \Rightarrow r(a,c)$ \\ \hline
	\end{tabular}
	\caption{Definitions of the main properties of binary relations.}
	\label{tab:rel_properties}
\end{center}
\end{table}

\begin{table}[ht]
	\centering
		\includegraphics[width=0.8\linewidth]{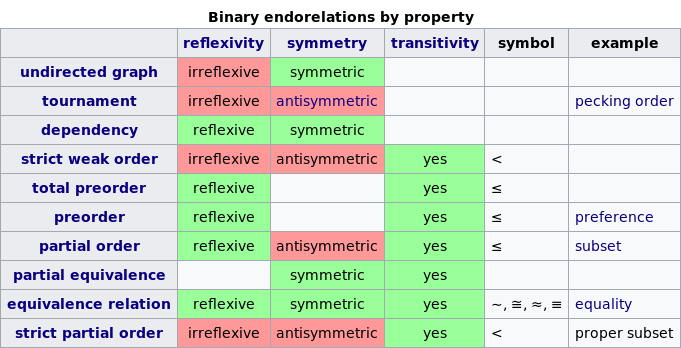}
    \caption{Different types of binary relations in set theory. From Wikipedia
    page on binary relations \cite{wiki:binary_relations}.}
	\label{fig:wiki_bin_rel}
\end{table}

There are many such common examples of these combinations in
knowledge graphs, as there are many hierarchical and similarity relations.
For example, the relations \tt{older} 
and \tt{father} are 
both strict hierarchies, thus antisymmetric and irreflexive.
But one is transitive (\tt{older}) whereas the other is not, 
and that makes all the difference at inference time.
Similarly for symmetric relations, such as \tt{has-the-same-parents-as} 
and \tt{friend}, your sibling's parents are 
also yours which makes the
first relation transitive, whereas your friend's friends are not necessarily yours. 
Note that this makes the \tt{has-the-same-parents-as} relation 
reflexive---it is thus an equivalence relation.

Relational learning models must be able to handle relations that exhibit 
each of the possible combinations of these
properties, since they are all very common, but imply different
types of reasoning, as already acknowledged by \citet{bordes2013irreflexive}.
Given that a relation can be reflexive, irreflexive, or neither;
symmetric, antisymmetric, or neither; and transitive or not, we end up
with 18 possible combinations. However we will not
address the cases of little interest where (i) none of these
properties are true, (ii) only reflexivity or irreflexivity is true,
(iii) the irreflexive, symmetric and transitive case 
as the only consistent possibility is that all facts are false,
and (iv) the irreflexive transitive case that again must be
either all false, or antisymmetric---and thus corresponds to an
already existing case---to be consistent.
Indeed, if one observes two true facts $r(s,o)$ and
$r(o,s)$, by application of the transitivity rule, $r(s,s)$
and $r(o,o)$ must be true, which explains the inconsistency
of cases (iii) and (iv), as they are irreflexive.
This leaves us with 13 cases of interest.
To evaluate the ability of models to learn these properties,
we generate random $50 \times 50$ matrices that exhibit each
combination.

To do so, we sample random square sign matrices 
$Y \in \{-1,1\}^{\Ne \times \Ne}$. 
First we fill the diagonal with
$1$, $-1$ or missing depending on reflexivity/irreflexivity or none.
Then we make successive passes over the data to make it [anti-]symmetric
and/or transitive, until all of the properties are true over the whole matrix.
A pass to make a matrix symmetric consists in assigning
$y_{ji} \leftarrow y_{ij}$ for all $i,j \in 1,\ldots, \Ne$ where $ i < j$,
and $y_{ji} \leftarrow -y_{ij}$ to make it antisymmetric.
A pass to make a matrix transitive consists in assigning 
$y_{ij} \leftarrow 1$ if there exists a $k \in 1,\ldots, \Ne$
such that $y_{ik} = y_{kj} = 1$, for all $i,j \in 1,\ldots, \Ne$. 
When no more assignment is made
during the passes it means the desired properties are true,
and the relation generation is finished.

We also sample each matrix under the constraint of having a balanced
number of positives and negatives up to $\pm 1 \%$. Though
there are many more negatives than positives in real knowledge
graphs, in practice negatives are generally subsampled or generated 
to match the number of positive facts
\cite{bordes2013translating,nickel_2016_holographic}.

We first learn each
relation individually as in a single relation knowledge graph,
and then jointly.
In the joint case, note that since each relation is generated independently,
there is no signal shared across the relations that would help
predicting facts of one relation from facts of another relation, 
thus only the ability to learn each relation patterns is tested.
The proportion of observed facts is generally very small
in real knowledge graphs. To assess models robustness
to missing data, we also reduce the proportion 
of the training set when learning the different relations 
jointly.

Results are averaged over a 10-fold cross-validation,
with 80\% training, 10\% validation and 10\% test
in the individual learning case.
In the joint learning case, the training percentage varies
between 80\% and 10\%, the validation set size is kept constant at 10\%,
and the test set contains the remaining samples---between 10\% and 80\%.

\subsection{Results}
\label{sec:rel_props_res}

Results are first reported on each relation, then
jointly and with decreasing proportion of training data.

\subsubsection{Individual Relation Learning}
\label{sec:rel_props_indiv_res}


First of all, results were identical for all models whether the 
relations were reflexive, irreflexive, or 
neither (unobserved). This tells us that 
reflexivity and irreflexivity are not so important in practice as
they do not add or remove any quality in the prediction of latent
factor models. We report
only results for different combinations of symmetry/antisymmetry 
and transitivity in the main text. Results of combinations including
reflexivity and irreflexivity are reported in \Cref{app:refl_irrefl}.

\begin{figure}[t]
	\centering
	\includegraphics[width=0.422\linewidth]{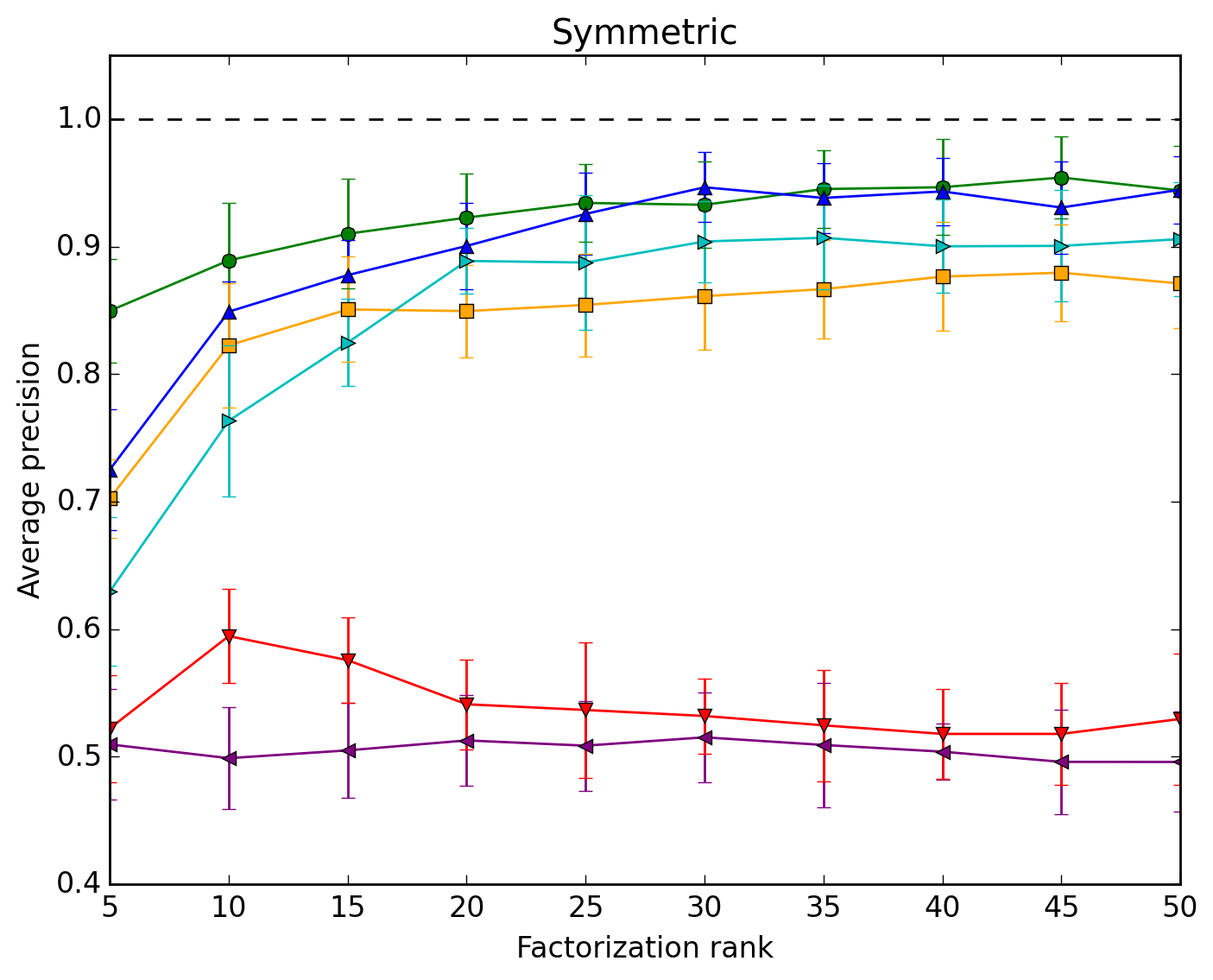}
	\includegraphics[width=0.56\linewidth]{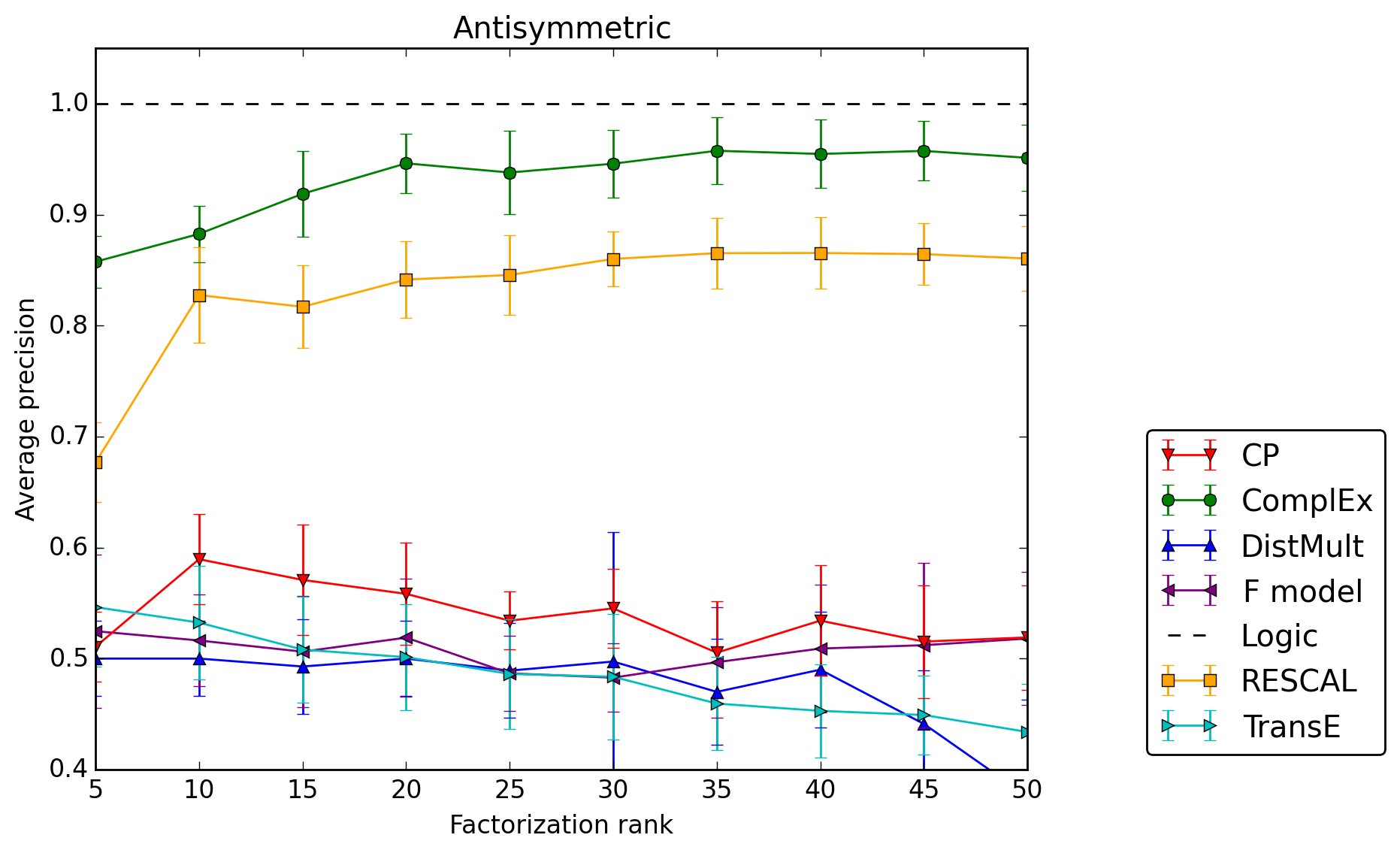}
	\caption{Generated symmetric (left) and antisymmetric (right) relations with 50 entities. Average precision for each rank ranging from 5 to 50 for each model.}
	\label{fig:exp_symm}
\end{figure}

\begin{figure}[t]
	\centering
	\includegraphics[width=0.56\linewidth]{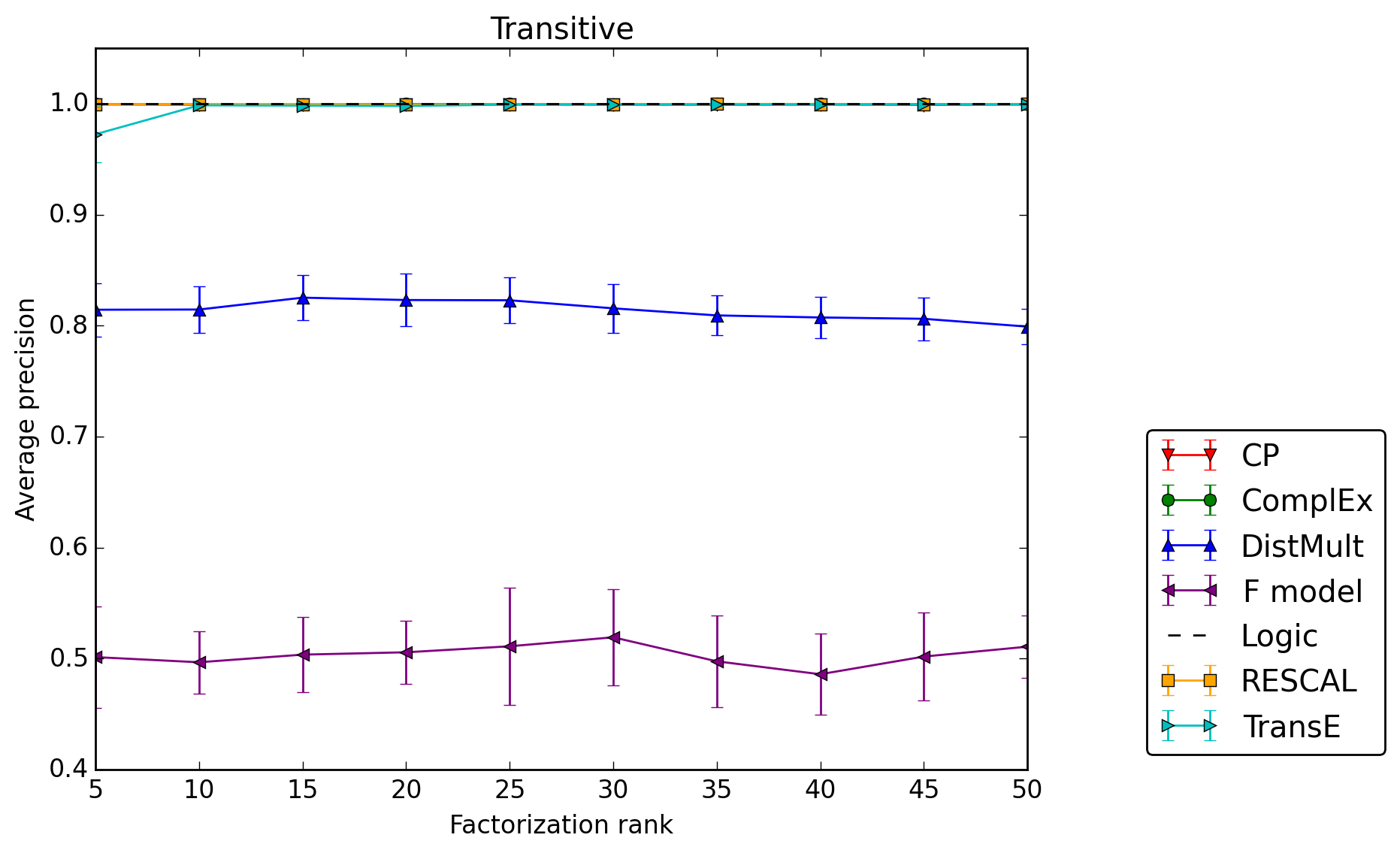}
	\caption{Generated transitive relation with 50 entities. Average precision for each rank ranging from 5 to 50 for each model.}
	\label{fig:exp_trans}
\end{figure}

\begin{figure}[t]
	\centering
	\includegraphics[width=0.422\linewidth]{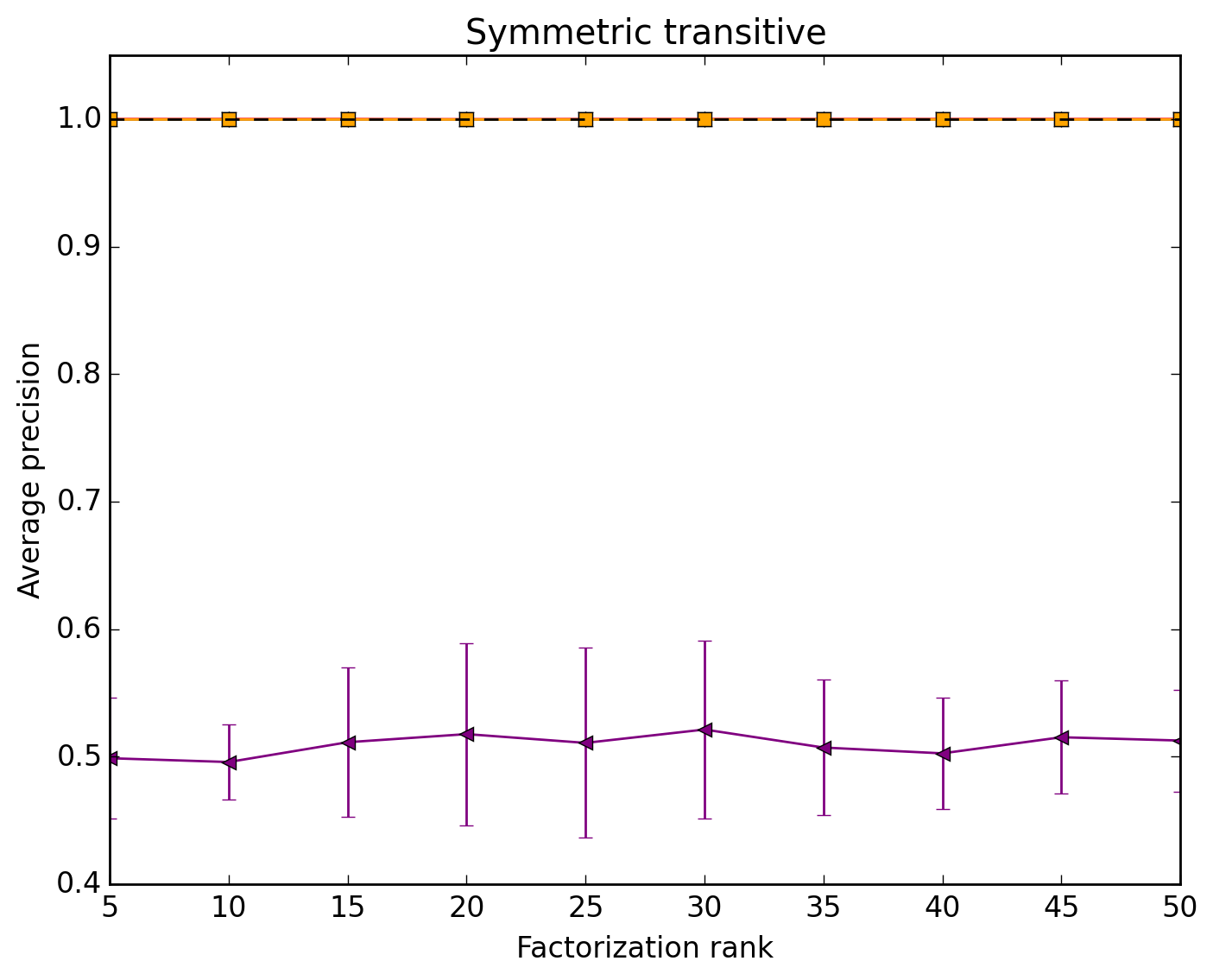}
	\includegraphics[width=0.56\linewidth]{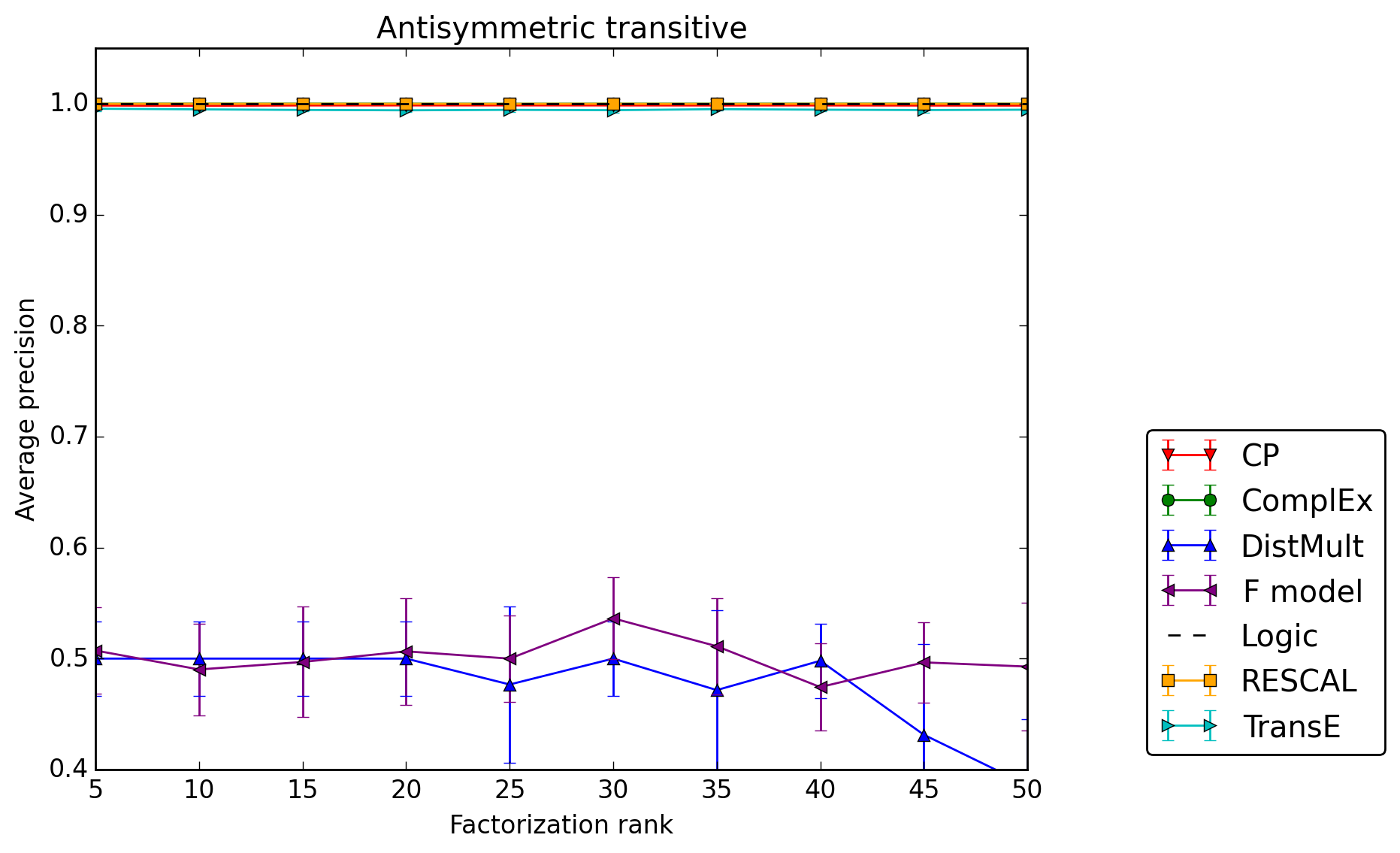}
	\caption{Generated symmetric and transitive (left) and antisymmetric and transitive (right) relations with 50 entities. Average precision for each rank ranging from 5 to 50 for each model.}
	\label{fig:exp_symm_trans}
\end{figure}

\Cref{fig:exp_symm} shows the average precision for 
each model over the
generated symmetric and antisymmetric relations.
Surprisingly, on such simple relations with 80\%
of observed data, only \textsc{ComplEx} and \textsc{RESCAL} manage
to learn from the symmetric and antisymmetric patterns,
with a non-negligible advantage for the \textsc{ComplEx} model.
Moreover, we can see that with higher ranks, \textsc{RESCAL}
starts overfitting as its average precision decreases
presumably due to its quadratic number of parameters
with respect to the rank, since each relation $r$
is represented by a matrix $W_r \in \mathbb{R}^{\rk \times \rk}$,
as showed in \Cref{eq:rescal}.

The \textsc{CP} model probably fails due to its uncorrelated
representations of entities as subject and as objects,
which makes it unable to model symmetry and antisymmetry.
\textsc{DistMult} unsurprisingly fails in the non-symmetric cases, 
due to the symmetric nature of its scoring function,
and thus succeeds in the symmetric case.
More unexpectedly, the \textsc{TransE} model has a hard time on antisymmetry,
but performs well
on the symmetric relation, by zeroing its relation
embedding, as explained in \Cref{sec:models}.
The F model, cannot actually
generalize in a single relation case, as it has one
single embedding for each (ordered) entity pair.
For any fact $r(s,o)$ in the test set, the
entity pair $(s,o)$ has never been seen in the training
set, and thus has a random embedding vector.

\Cref{fig:exp_symm_trans} shows results 
for the symmetric transitive and
antisymmetric transitive relations, and \Cref{fig:exp_trans} for
the transitive only relations. Almost all models, except the F model
and \textsc{DistMult} in the non-symmetric cases, perfectly generalize with very low-rank.
This tells us that transitivity actually 
brings so much structure in the data that it makes the problem very easy 
for latent factor models when most of the data is observed. 

Most state-of-the-art latent factor models are surprisingly
unable to model all the basic properties of binary relations.
Though most of the time those relations are learnt 
together, but also with much less observed facts.
We next assess the models ability to learn these five relations
together, and their robustness to sparse observations
by gradually decreasing the size of the training set.


\subsubsection{Joint Learning}

\begin{figure}[t]
	\centering
	\includegraphics[width=0.422\linewidth]{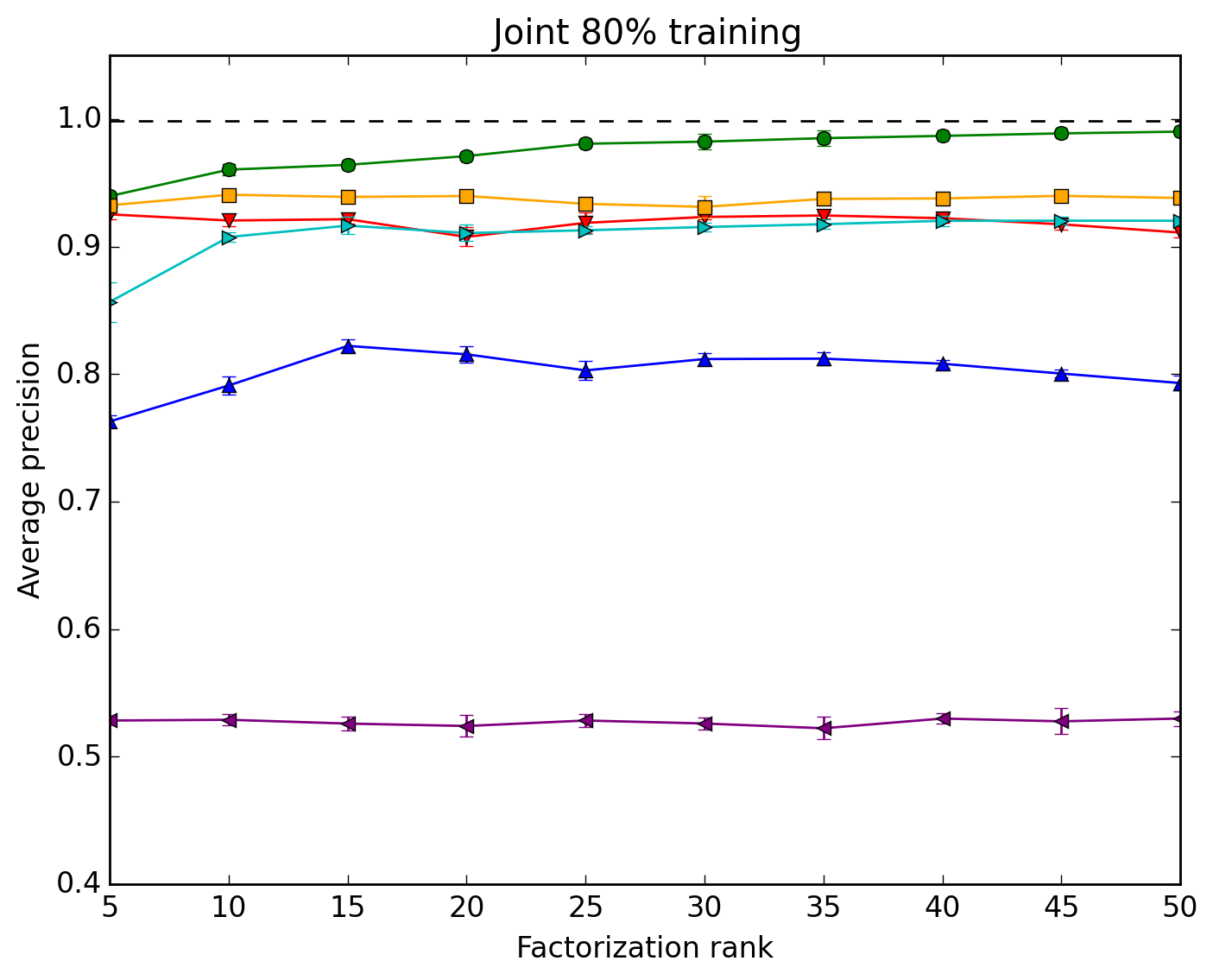}
	\includegraphics[width=0.56\linewidth]{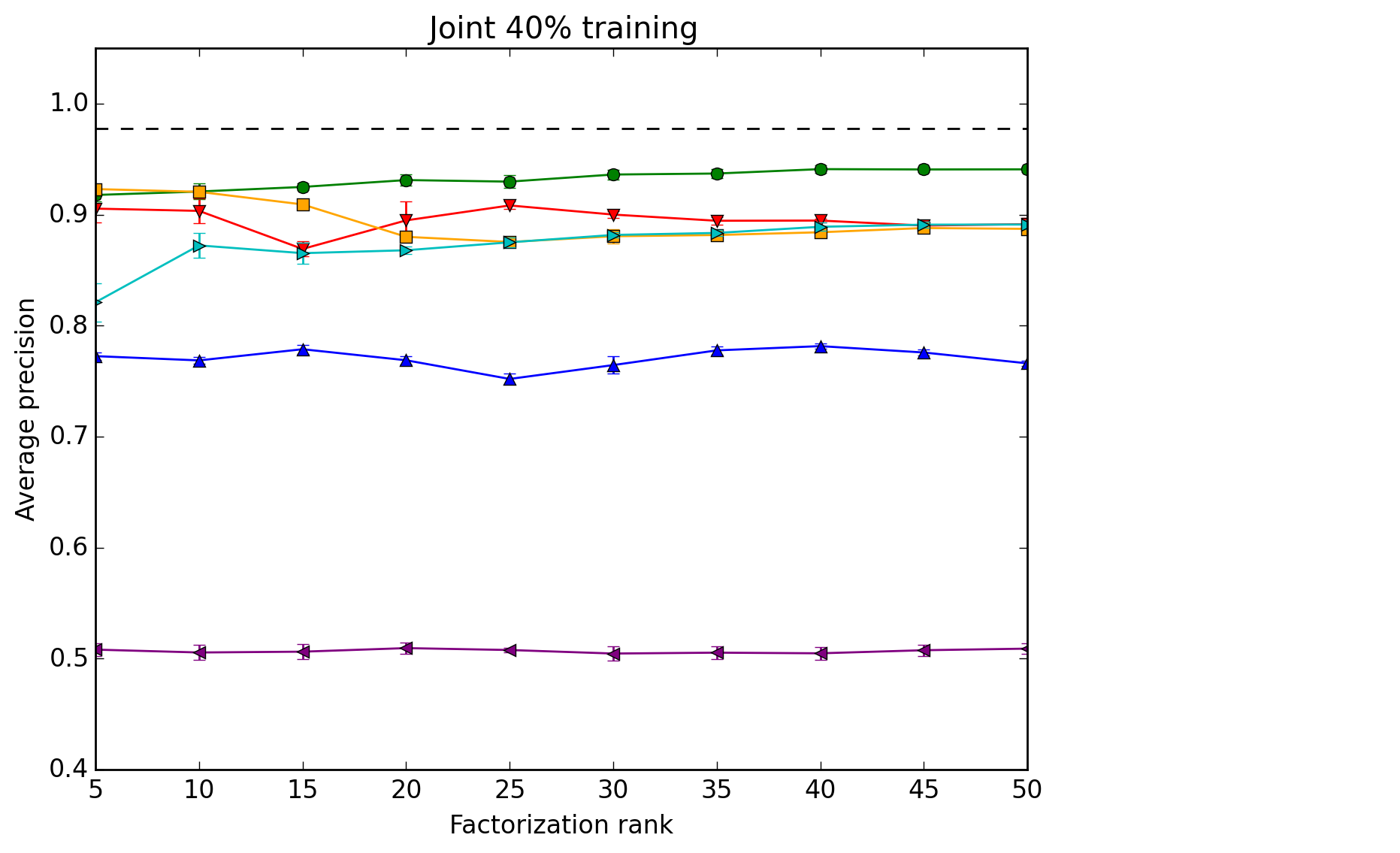}
	\includegraphics[width=0.422\linewidth]{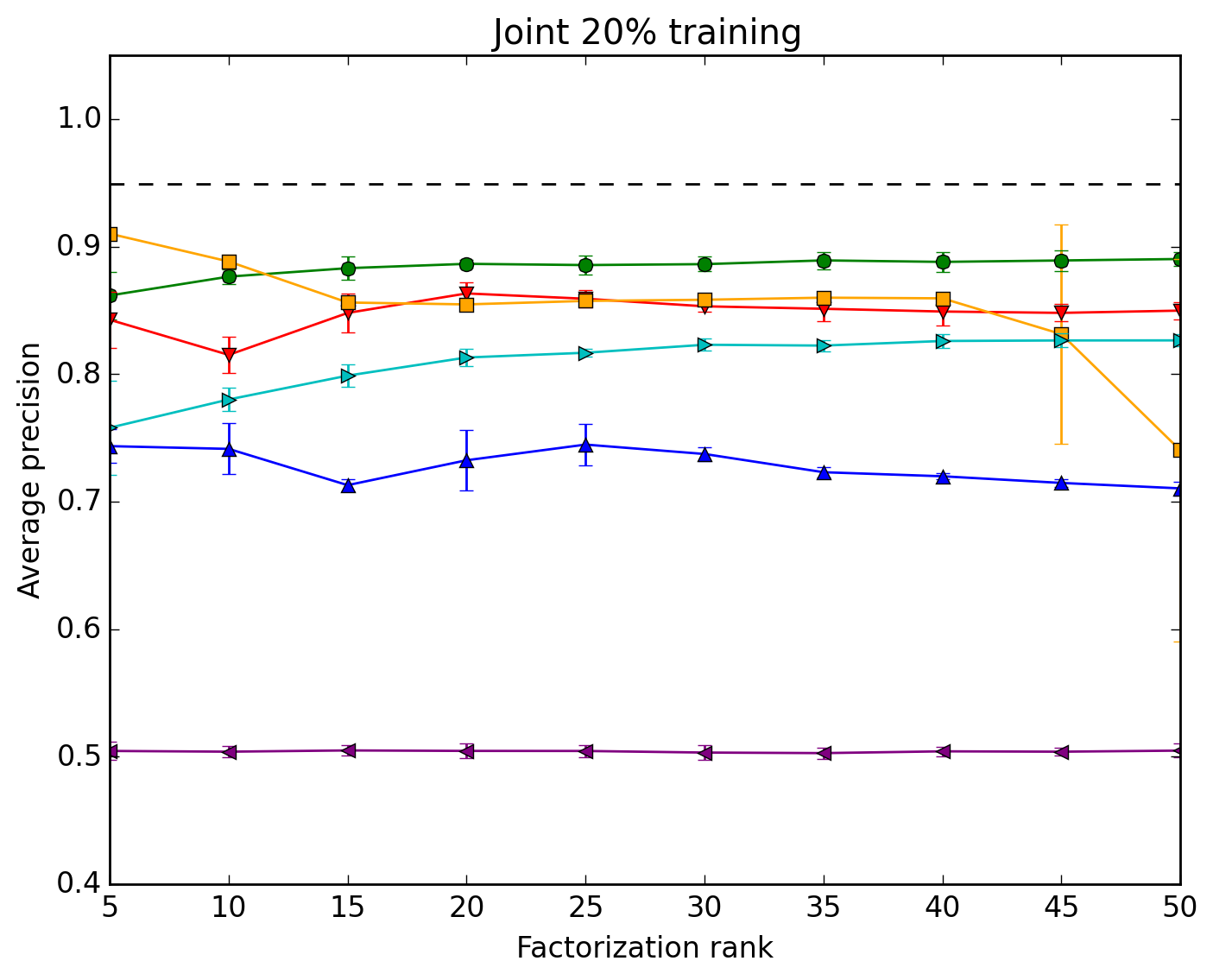}
	\includegraphics[width=0.56\linewidth]{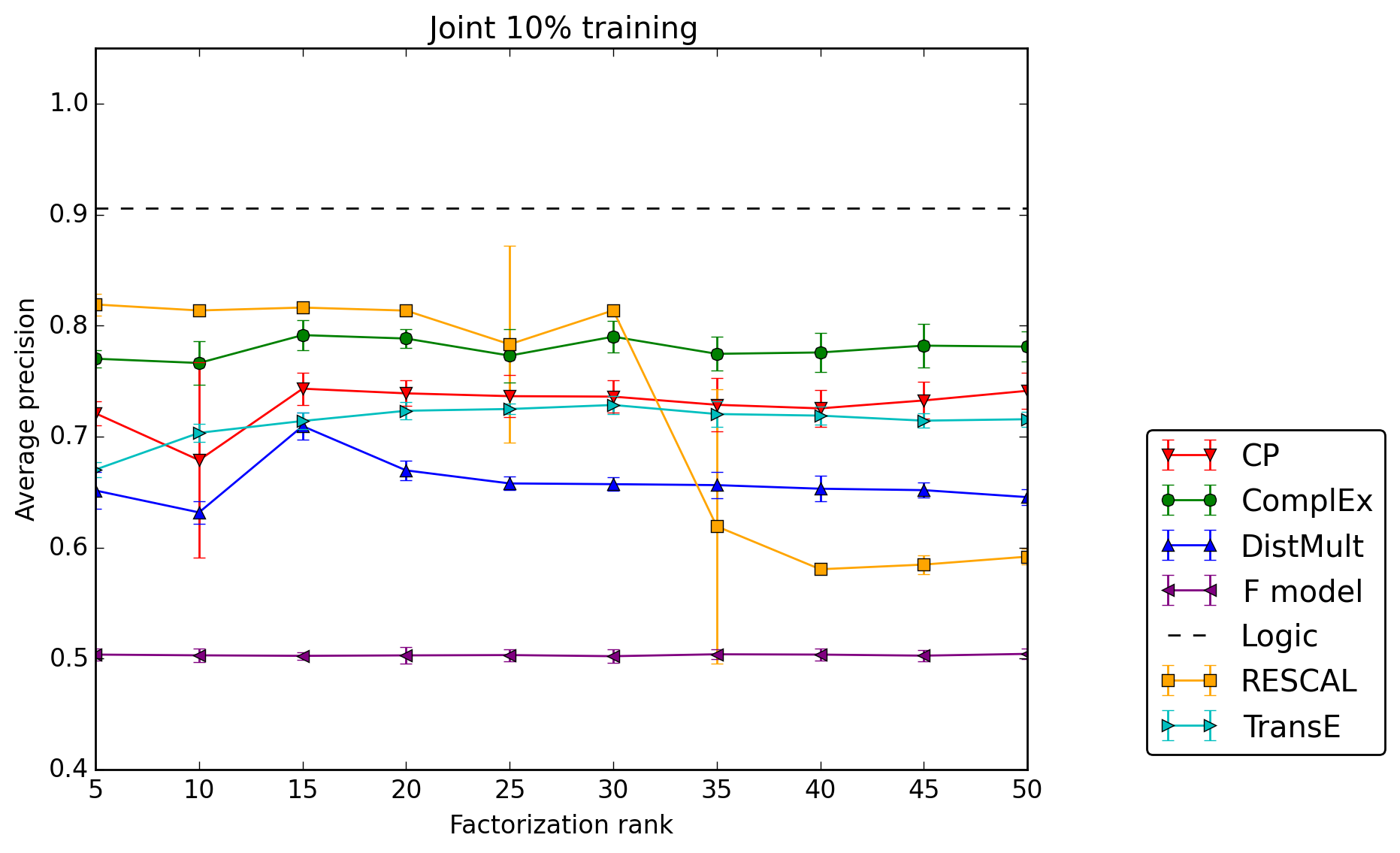}
	\caption{Joint learning of the 5 relations with 50 entities: one symmetric, one antisymmetric, one transitive, one symmetric and transitive, and one antisymmetric and transitive. Average Precision for each factorization rank ranging from 5 to 50 of each model. Top-Left: 80\% train set, Top-Right: 40\% train set, Bottom-Left: 20\% train set, Bottom-Right: 10\% train set.}
	\label{fig:exp_joint_toy}
\end{figure}

\Cref{fig:exp_joint_toy} shows the results when all five above relations
are jointly learned, for different proportions of the training set: 
80\%, 40\%, 20\%, 10\%. As expected the scores drop,
and the gap between the---deterministic logic---upper-bound 
and latent factor models widen with the decrease of training data.
\textsc{ComplEx} proves to be the most robust to 
missing data down to 20\%, but match logical inference
only with 80\% of training data. 

\textsc{RESCAL} again overfits
with the rank increasing, but is the best performing model
with 10\% of the training set, up to rank $\rk=30$.
This suggests that having richer relation representations
than entity representations, that is with more parameters,
can be profitable for learning relation properties from
little data. However the reason why the variance of
\textsc{RESCAL}'s average precision decreases again for 
$K \geq 40$ remains mysterious.

The \textsc{CP} and \textsc{TransE} models seem to be more sensitive to
missing data as their curves progressively get away from
\textsc{RESCAL}'s one with the percentage of observed data decreasing.
\textsc{DistMult}, being a symmetric model, is below
the other models in the four settings as some
of the relations are not symmetric.

Since each relation is generated independently,
having observed the entity pair $(s,o)$ in
the other relations does not help the F model,
and it thus fails here too.
At 10\%, we see that the latent factor models 
cannot match logical inference,
suggesting that the number of examples is not 
sufficient to learn these properties.

Finally, in the last setting with 10\% of the training set,
the best models are still 10 points
below the best achievable average precision,
showing that they need a large amount of training
data to correctly learn these basic properties 
of binary relations.

These results should be taken cautiously as this experiment does 
\emph{not} state that in general at least 80\% of the facts should be
observed in order to learn these properties correctly.
Indeed, here the 5 relations are completely
uncorrelated, while in real knowledge graphs they generally are correlated
and thus share information. Also, as often in machine learning,
the ratio between the number of parameters and the number 
of data points is more informative about generalization
than the number of data points alone.\\

\textbf{Relation Properties Experiments Summary:}
\begin{itemize}
    \item Only \textsc{ComplEx} manages to learn each combination near perfectly.
    \item \textsc{RESCAL} is the most robust to missing data with low ranks.
    \item There is room for improvement on all models when a lot of data is missing.
\end{itemize}

\section{Learning Inter-Relational Patterns: Family Relationships}

\label{sec:family_rels}

We generated family trees and their corresponding kinship
relations and facts, and designed three different splits 
of the data.
The three splits try to assess
different inductive properties of the latent models, by
giving more or less supporting facts in the training set. 

\subsection{Experimental Design}

Predicting family relationships is an old task in AI, popularised by 
Hinton's kinship data set \cite{Hinton1986}.
Generated synthetic families for testing link-prediction models
have also been recently proposed \cite{garcia2014effective}.
In this public dataset, generated families are all
intertwined with each other in it. We here want each family to be 
disjoint from the other ones, such that there is no true
fact between entities of two different families, and
we will see why below.

We propose here to generate family relations from synthetic
family trees, namely: 
\tt{mother}, \tt{father}, \tt{husband}, \tt{wife}, \tt{son}, \tt{daughter}, \tt{brother},
\tt{sister}, \tt{uncle}, \tt{aunt}, \tt{nephew}, \tt{niece}, \tt{cousin}, \tt{grandfather},
\tt{grandson}, \tt{grandmother} and \tt{granddaughter}.


\begin{figure}[t]
\centering
	\includegraphics[width=0.7\linewidth]{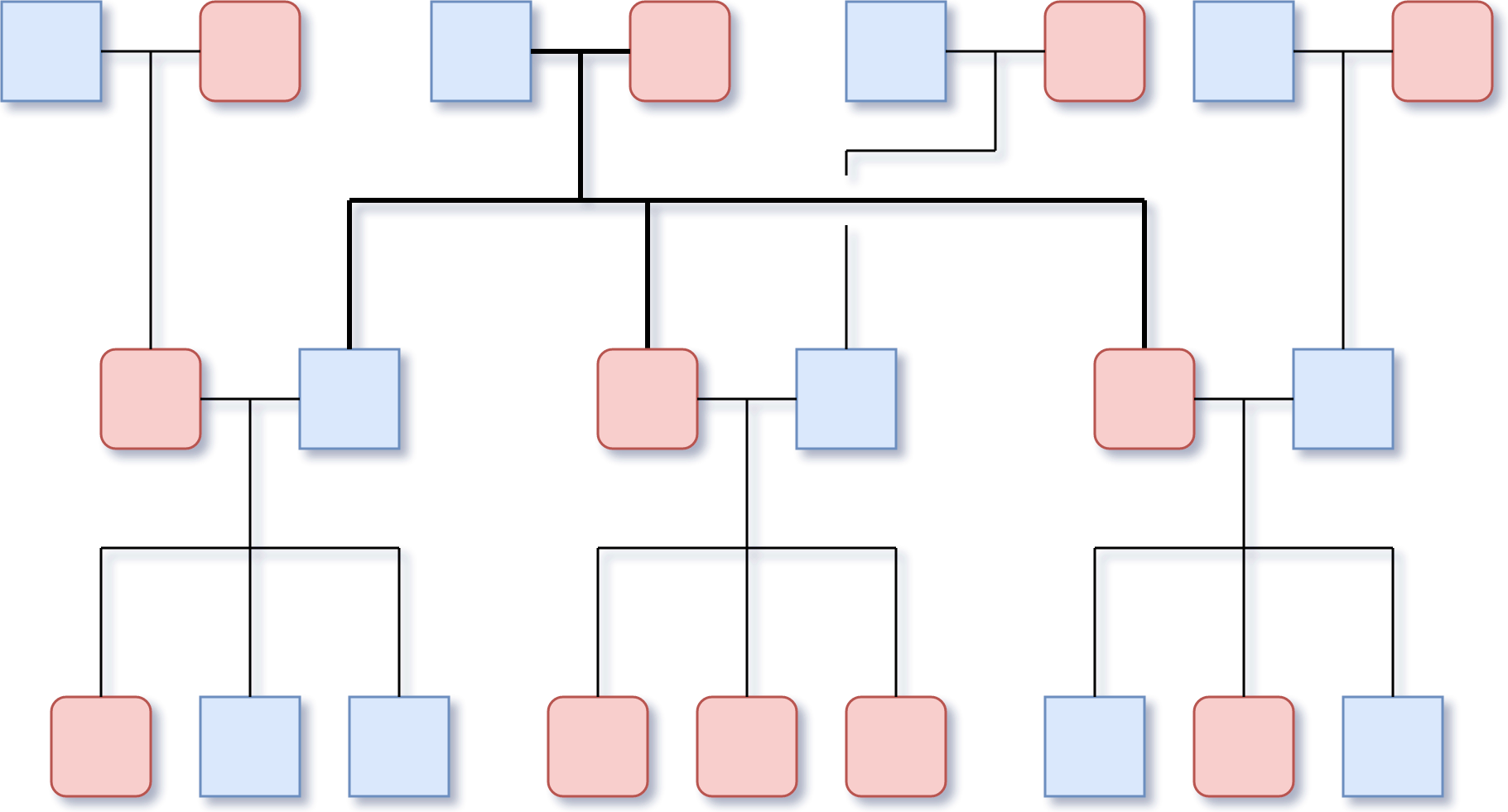}
	\caption{Example of a generated family tree.}
	\label{fig:family_tree}
\end{figure}


We sample five families independently that span over three generations
from a unique couple, with three children per couple of random sex,
where husbands, wives and their parents were added to wed the middle generation.
\Cref{fig:family_tree} shows an example of such a family tree.
The whole data set totals 115 entities---23 persons per family---and
the 17 relations mentioned above. Each family thus consists
in 8993 true and false facts.

\begin{table}[t]
\begin{center}
	\begin{tabular}{|c|} \hline
		$\mathtt{\forall a \forall b \forall c\enspace father(a,c) \wedge mother(b,c) \Rightarrow husband(a,b)}$ \\ \hline
		$\mathtt{\forall a \forall b \forall c\enspace father(a,c) \wedge mother(b,c) \Rightarrow wife(b,a)}$ \\ \hline
		
		$\mathtt{\forall a \forall b \forall c\enspace daughter(a,c) \wedge son(b,c) \Rightarrow sister(a,b)}$ \\ \hline
		$\mathtt{\forall a \forall b \forall c\enspace daughter(a,c) \wedge son(b,c) \Rightarrow brother(b,a)}$ \\ \hline
		
		$\mathtt{\forall a \forall b \forall c\enspace father(a,b) \wedge father(b,c) \Rightarrow grandfather(a,c)}$ \\ \hline
		$\mathtt{\forall a \forall b \forall c\enspace son(a,b) \wedge son(b,c) \Rightarrow grandson(a,c)}$ \\ \hline

		$\mathtt{\forall a \forall b \forall c\enspace mother(a,b) \wedge mother(b,c) \Rightarrow grandmother(a,c)}$ \\ \hline
		$\mathtt{\forall a \forall b \forall c\enspace daughter(a,b) \wedge daughter(b,c) \Rightarrow granddaughter(a,c)}$ \\ \hline
		
		$\mathtt{\forall a \forall b \forall c \forall d\enspace son(a,b) \wedge daughter(b,c) \wedge son(d,c) \Rightarrow uncle(d,a)}$ \\ \hline
		$\mathtt{\forall a \forall b \forall c \forall d\enspace daughter(a,b) \wedge son(b,c) \wedge daughter(d,c) \Rightarrow aunt(d,a)}$ \\ \hline
		
		$\mathtt{\forall a \forall b \forall c \forall d\enspace son(a,b) \wedge daughter(b,c) \wedge son(d,c) \Rightarrow nephew(a,d)}$ \\ \hline
		$\mathtt{\forall a \forall b \forall c \forall d\enspace daughter(a,b) \wedge son(b,c) \wedge daughter(d,c) \Rightarrow niece(a,d)}$ \\ \hline
		
		$\mathtt{\forall a \forall b \forall c \forall d \forall e\enspace son(a,b) \wedge daughter(b,c) \wedge son(d,c) \wedge daughter(e,d) \Rightarrow cousin(a,e)}$ \\ \hline

	\end{tabular}
	\caption{Examples of rules to deduce all relations from the four relations: \tt{mother}, \tt{father}, \tt{son} and \tt{daughter}.}
	\label{tab:rel_prop_examples}
\end{center}
\end{table}

Within these traditional families that feature only married heterosexual
couples that do not divorce and have children, one can figure 
out that the relations \tt{mother}, \tt{father}, \tt{son} and \tt{daughter}
are sufficient to deduce the 13 remaining ones.
Examples of rules that allow deducing these 13 relations
from the 4 main ones are shown in \Cref{tab:rel_prop_examples}. 
From this fact, we devise three different splits of the data.

\begin{figure}[t]
	\centering
	\includegraphics[width=0.5\linewidth]{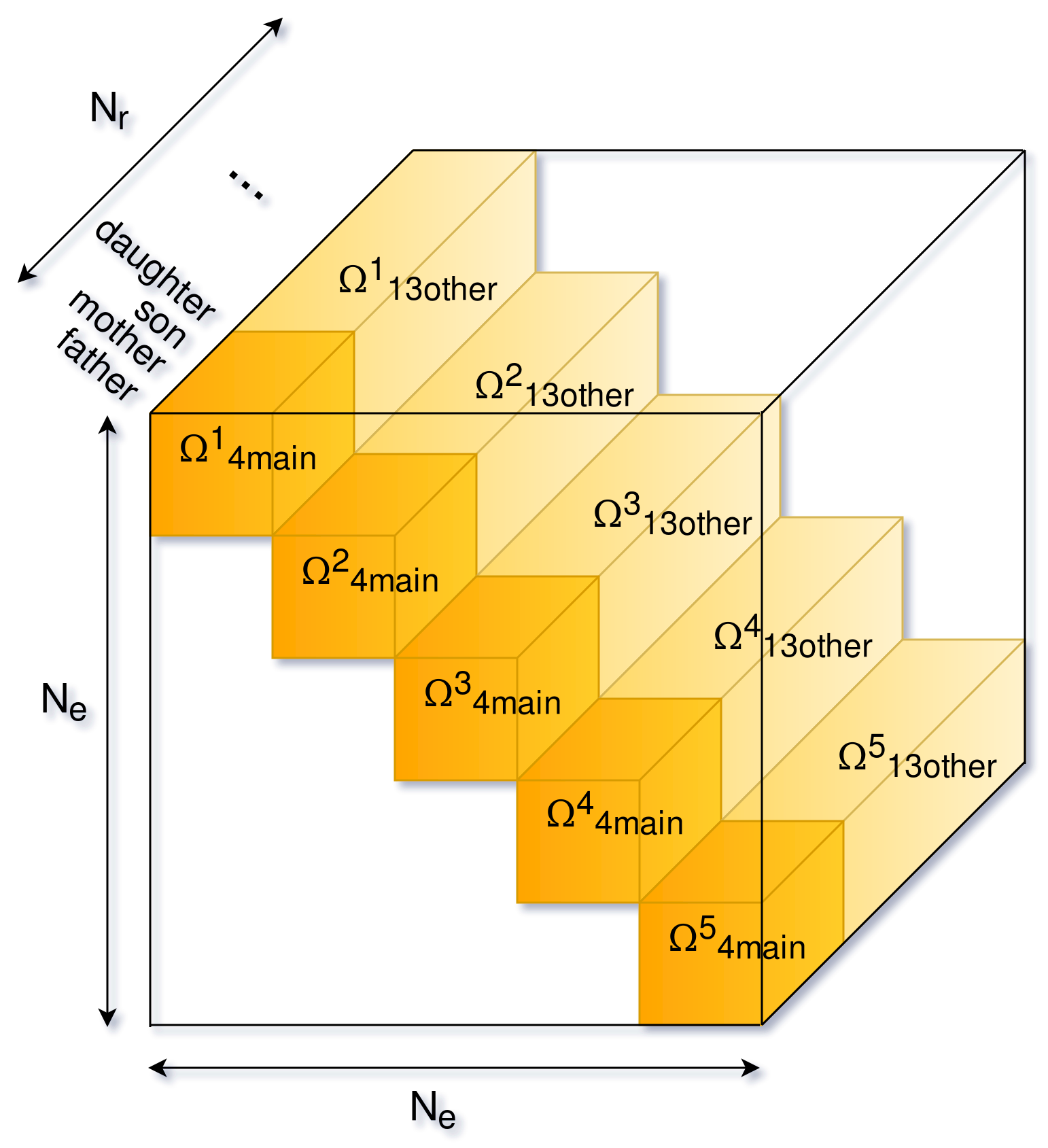}
    \caption{Tensor representation of the observed subsets for the family experiments. The part in dark orange represents the sets containing the 
    four relations \tt{mother}, \tt{father}, \tt{son} and \tt{daughter},
    while the part in light orange represents the 13 other relations.}
	\label{fig:family_sets}
\end{figure}

Let us first introduce notations for each subset of the observed
facts set $\Omega$. 
As each family is independent from the four others,
we will denote each entity set of each family from 1 to 5:
$\setent^1,\ldots,\setent^5$, where $\setent^i \cap \setent^j = \emptyset$
with $ i \neq j$. Accordingly, we will write the observed facts of each family
$\Omega^1,\ldots,\Omega^5$, where for all $((r,s,o),y_{rso}) \in \Omega^i$
we have $s,o \in \setent^i$. Observed fact sets that contain
only the 4 main relations \tt{mother}, \tt{father}, 
\tt{son} and \tt{daughter} are denoted by $\Omega_\fourm$, 
and the facts involving the 13 other
relations by $\Omega_\thiro$.
We thus have for each family $i$: $\Omega^i = \Omega^i_\fourm \cup 
\Omega^i_\thiro$.
\Cref{fig:family_sets} draws the corresponding tensor 
with each subset of the observed facts.
Finally, let the sampling function $\mathcal{S}_p(\Omega)$
be a uniformly random subset of $\Omega$
of size $|\mathcal{S}_p(\Omega)| = \lceil p|\Omega| \rceil$, with 
$0 \leq p \leq 1$, $p$ being the proportion of the set that
is randomly sampled.

We propose to split the data in three different ways
to explore inductive abilities of the models.
The first split is the classical \emph{random split} between training,
validation
and test sets, it will mainly serve as a control experiment for
the other splits.
In the second split, we aim at evaluating whether 
latent factor models are able to leverage this information.
To do so, we ensure that all the relations \tt{mother}, 
\tt{father}, \tt{son} and \tt{daughter} of the five families
are in the training set, and we split the 13 remaining ones
between training, validation and test set.
Formally: $\Omega_\train = \Omega_\fourm \cup \mathcal{S}_p(\Omega_\thiro)$.
We will call this splitting scheme
the \emph{evidence split}, as the training set always contains the
sufficient evidence to deduce the 13 other relations---that is
the four main ones.

\begin{figure}[t]
	\centering
    \subfloat[Random split]{\includegraphics[width=0.33\linewidth]{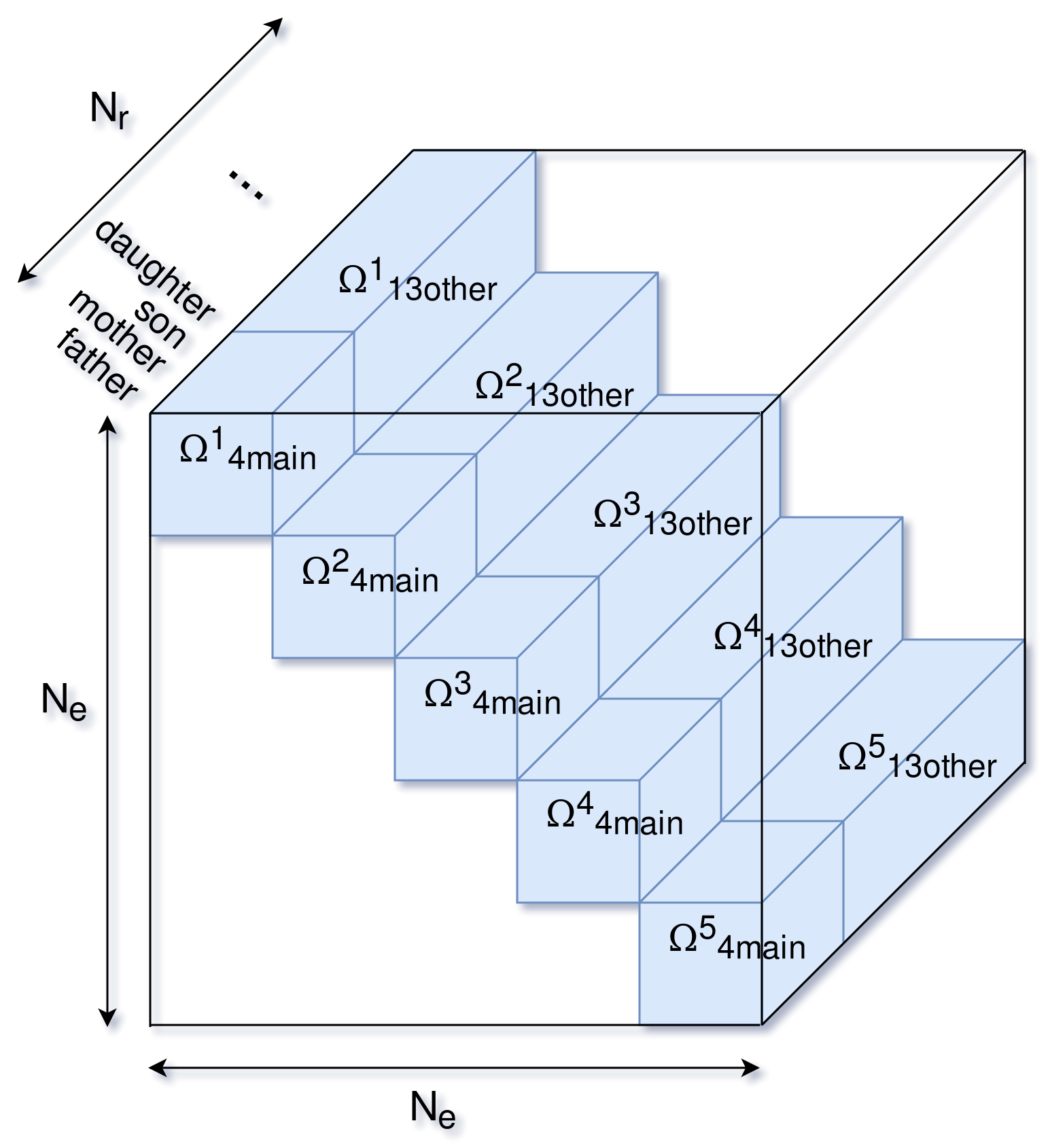}}
    \subfloat[Evidence split]{\includegraphics[width=0.33\linewidth]{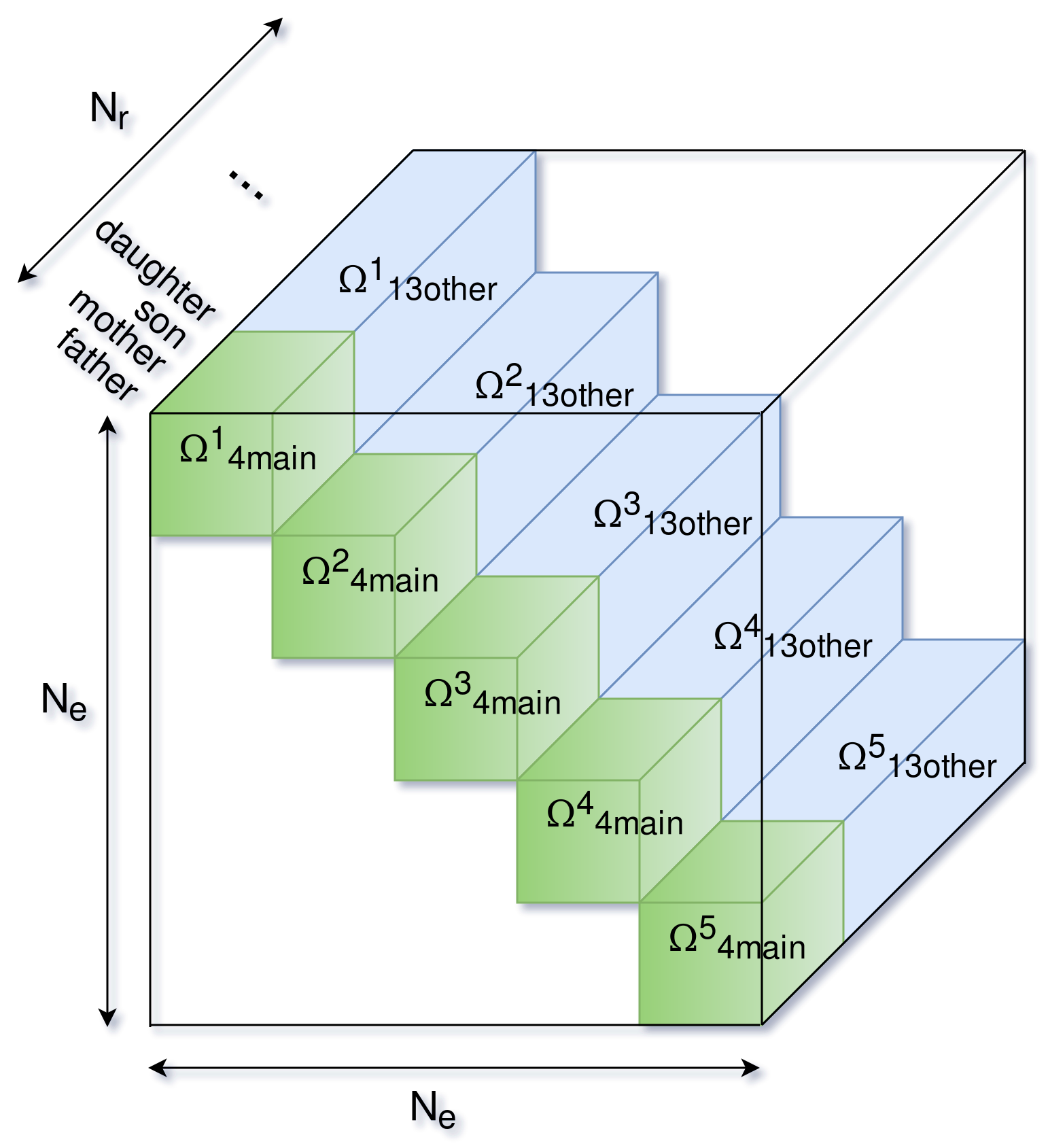}}
    \subfloat[Family split]{\includegraphics[width=0.33\linewidth]{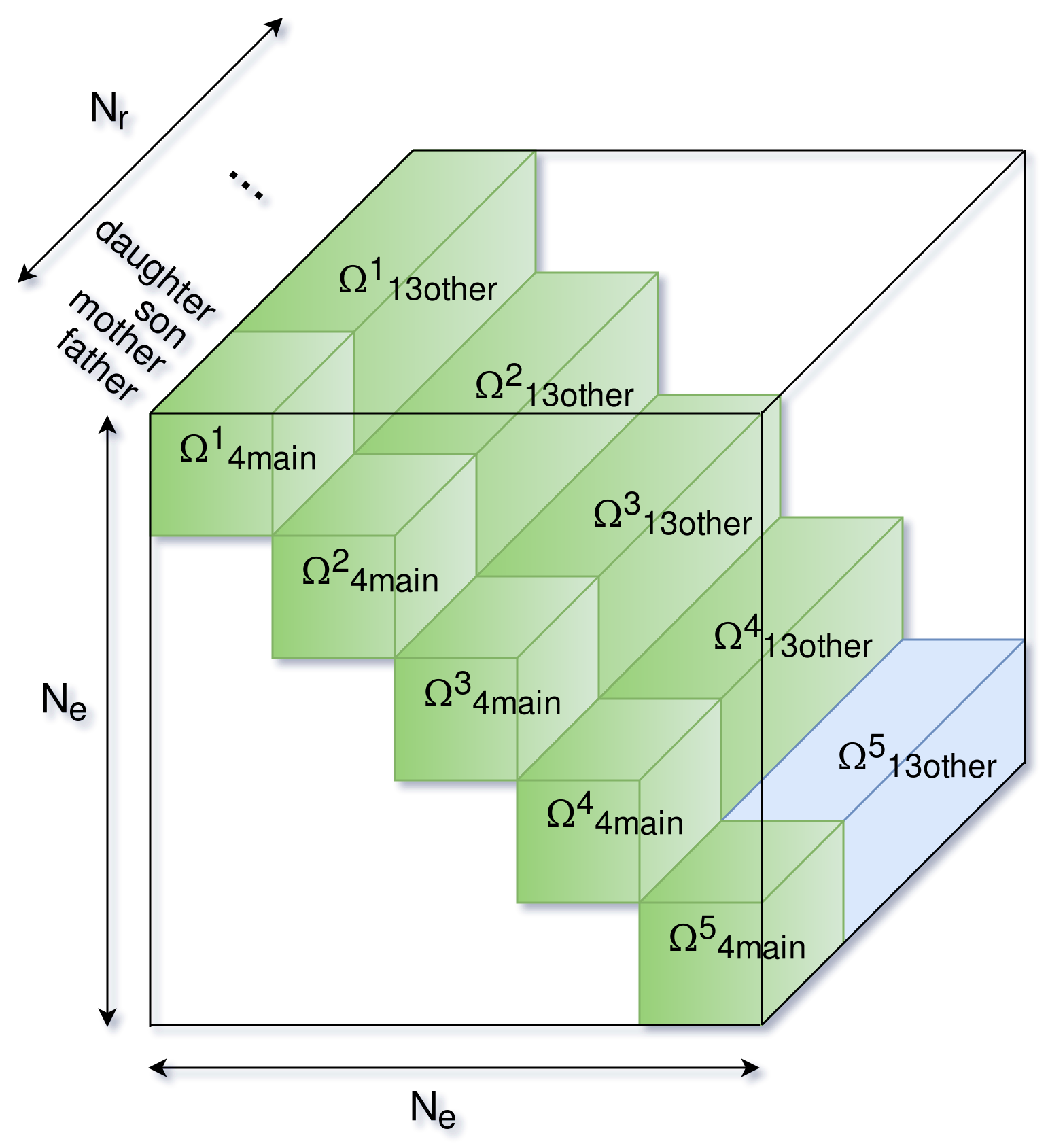}}
    \caption{Tensor representation of the three different splits. Green sets
    are always contained in the training set $\Omega_\train$, whereas blue
    sets are split among training, validation and test sets.}
	\label{fig:family_splits}
\end{figure}

Thirdly, we assess the ability 
of latent factor models to transfer knowledge learnt from 
a family to another, 
that is between disjoint set of entities.
In this split, the training set always contains all the relations 
for four out of the five families
plus all the \tt{mother}, \tt{father}, \tt{son} and \tt{daughter} of the fifth family, while the 13 other relations of this fifth family are split 
between training, validation and test set. 
Formally: $\Omega_\train = \Omega^{1-4}\cup \Omega^{5}_\fourm 
\cup \mathcal{S}_p(\Omega^5_\thiro)$.
We will call it the \emph{family split}.
\Cref{fig:family_splits} shows
tensor drawings of the three splits.

For each split we explore different values of $p \in \{0.8, 0.4,
0.2, 0.1\}$. We also run with $p=0$ in the last (family) split, which
corresponds to $\Omega_\train = \Omega^{1-4}\cup \Omega^{5}_\fourm$,
that is 4 entirely observed family, plus the 4 main relations
of the fifth one. Observe that it only makes sense to have $p=0$
in this last split. If latent factor models have expected inductive abilities,
they would be able to understand genealogical reasoning from
the four first families, and use this learned information 
to correctly predict the 13 other relations of the fifth family
from its four main ones.
Note that in the last two splits,
a deterministic logic inference system makes perfect
predictions---given rules such as the ones shown 
in \Cref{tab:rel_prop_examples}---for 
any value of $p$.
The number of facts in the training, validation and test sets
of each split are summarized in \Cref{tab:split_sizes}.

\begin{table}[t]
\begin{center}
    \resizebox{\columnwidth}{!}{%
    \begin{tabular}{ l|l||l|l|l|l|l }
        \centering
        & & \multicolumn{5}{c}{Size with $p=$} \\
        Split & Set & 0.8 & 0.4 & 0.2 & 0.1 & 0 \\
        \hline
        \multirow{3}{*}{Random} & $\Omega_\train = \mathcal{S}_p(\Omega)$ & 35973 & 17987 & 8994 & 4496 & -\\
         & $\Omega_{\mathrm{valid}}= \mathcal{S}_{0.1}(\Omega)$ & 4496 & 4496 & 4496 & 4496 & -\\
         & $\Omega_{\mathrm{test}}= \mathcal{S}_{(0.9-p)}(\Omega)$ & 4496 & 22482 & 31475 & 35973 & -\\
        \hline
        \multirow{3}{*}{Evidence} & $\Omega_\train= \Omega_\fourm \cup \mathcal{S}_p(\Omega_\thiro)$ & 38089 & 24334 & 17457 & 14019 & -\\
         & $\Omega_{\mathrm{valid}} = \mathcal{S}_{0.1}(\Omega_\thiro)$ & 3438 & 3438 & 3438 & 3438 & -\\
         & $\Omega_{\mathrm{test}}= \mathcal{S}_{(0.9-p)}(\Omega_\thiro)$ & 3438 & 17193 & 24070 & 27508 & -\\
        \hline
        \multirow{3}{*}{Family} & $\Omega_\train = \Omega^{1-4}\cup \Omega^{5}_\fourm \cup \mathcal{S}_p(\Omega^5_\thiro)$ & 43589 & 40839 & 39463 & 38776 & 38088\\
         & $\Omega_{\mathrm{valid}}= \mathcal{S}_{0.1}(\Omega^5_\thiro)$ & 688 & 688 & 688 & 688 & 688\\
         & $\Omega_{\mathrm{test}} = \mathcal{S}_{(0.9-p)}(\Omega^5_\thiro)$ & 688 & 3438 & 4814 & 5501 & 6189\\
    \end{tabular}
    }
    \caption{Training, validation and test set numbers for each split for each value of $p$.}
    \label{tab:split_sizes}
\end{center}
\end{table}

Similar splits of data have already been proposed
to evaluate rule-based inference models (for
example the UW-CSE dataset \cite{richardson2006markov}),
which are able of such transfer
of reasoning between disjoint sets of entities. 
Interestingly, such data sets have rarely been reused
in the subsequent latent factor model literature.
Results reported next might
give us a hint why this is the case.

\subsection{Results}
\label{sec:family_res}

Results are reported for each split separately.
In each of them
we again decrease progressively the amount of training data.

\subsubsection{Random Split}

\begin{figure}[ht]
	\centering
	\includegraphics[width=0.422\linewidth]{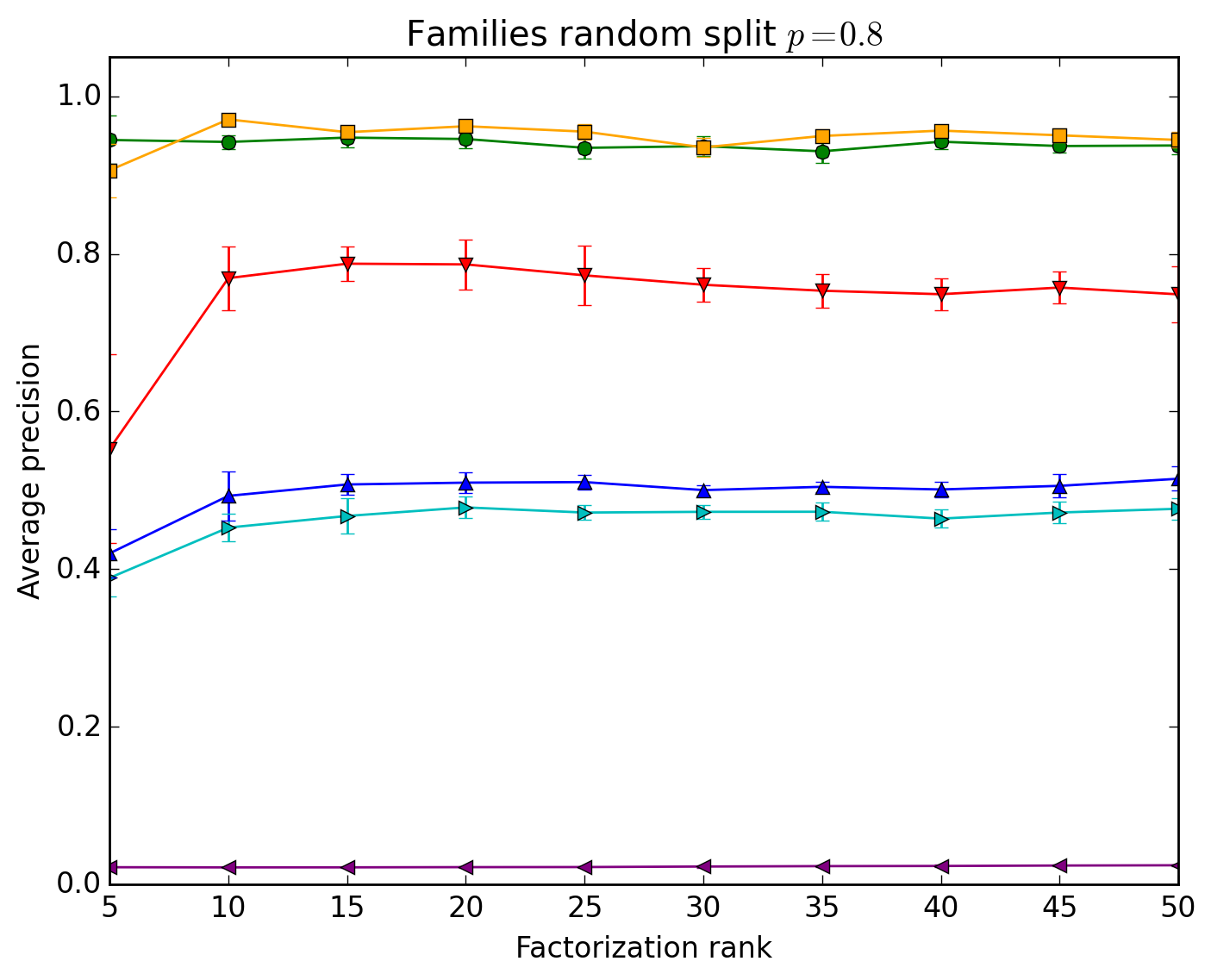}
	\includegraphics[width=0.56\linewidth]{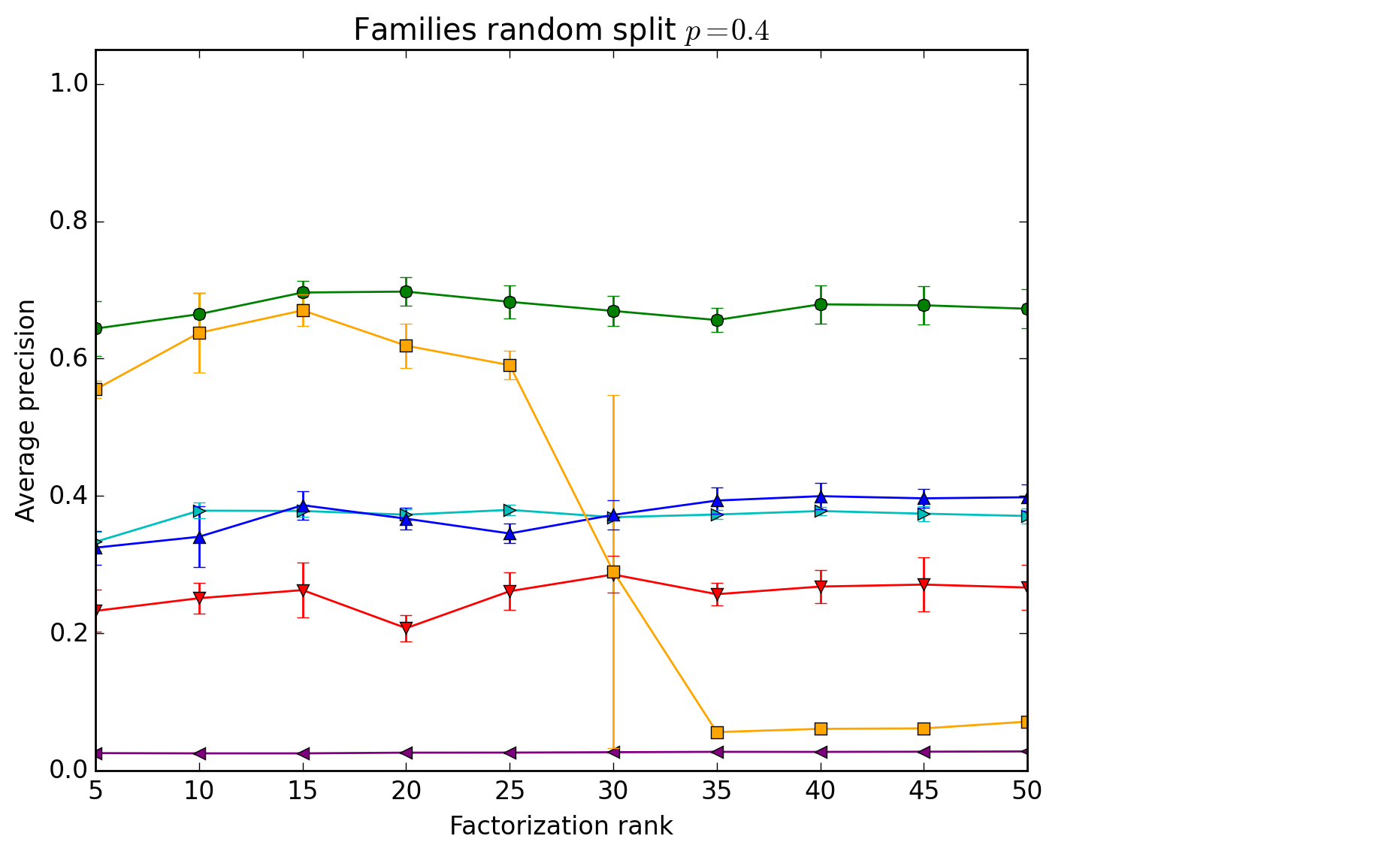}
	\includegraphics[width=0.422\linewidth]{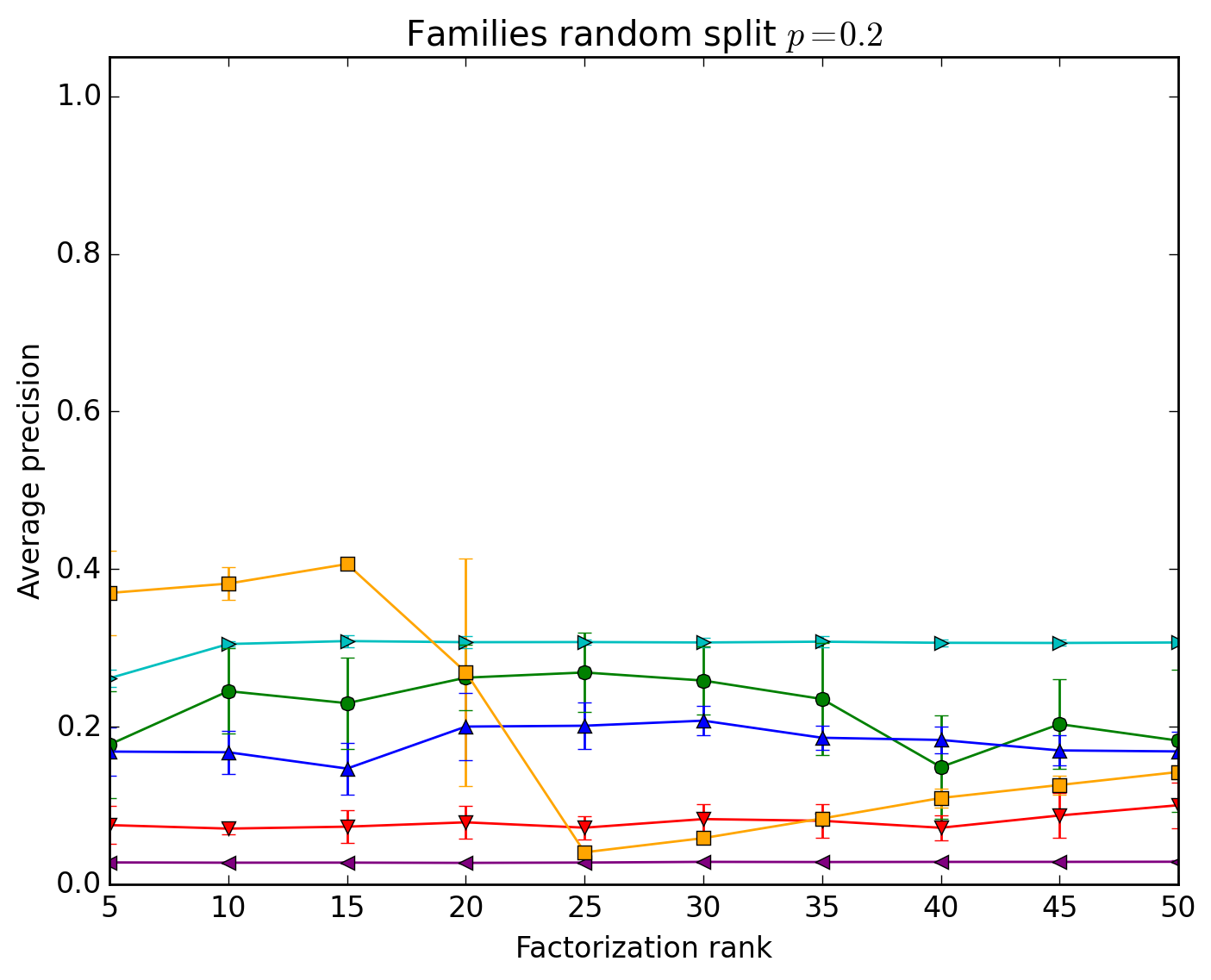}
	\includegraphics[width=0.56\linewidth]{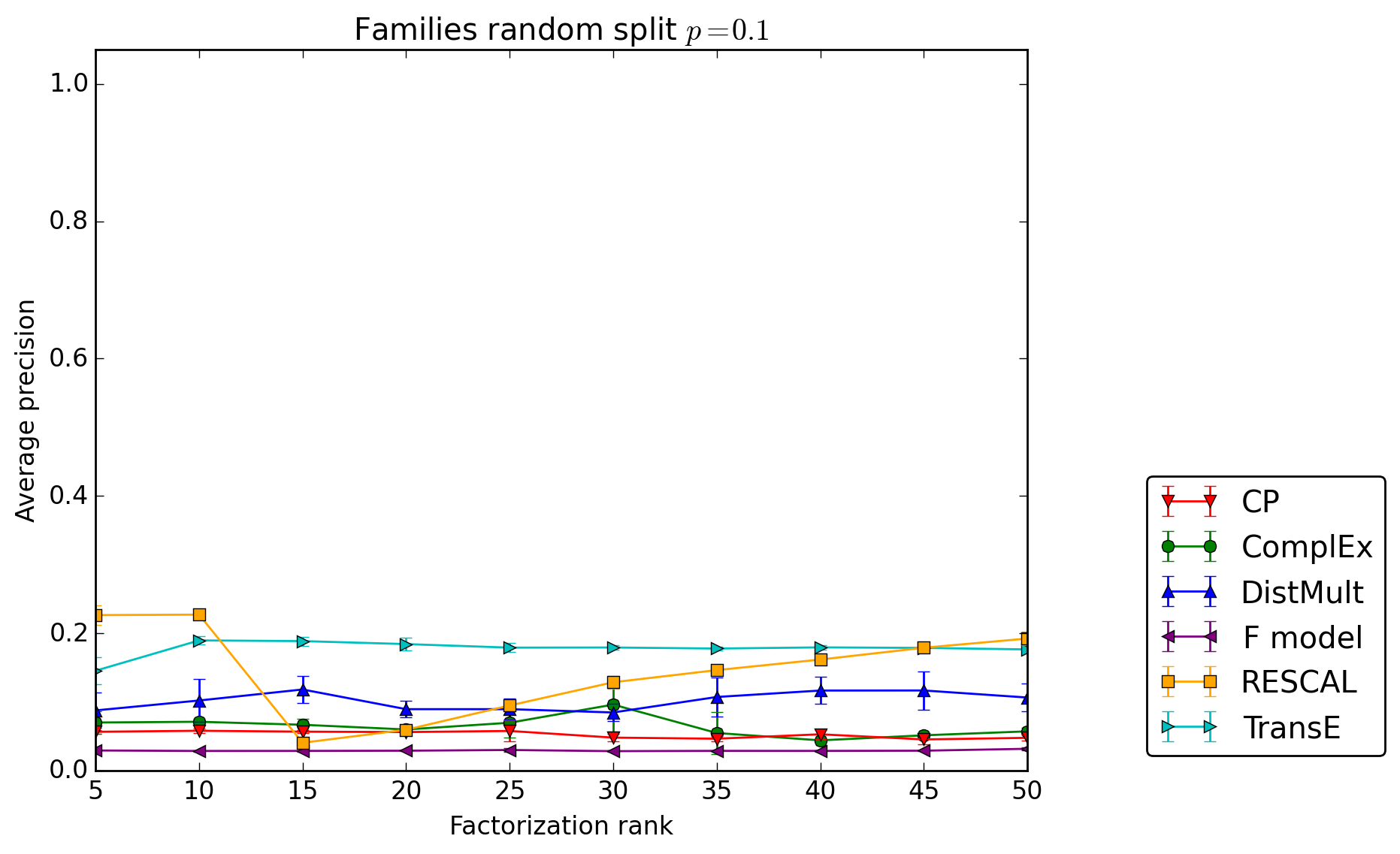}
	\caption{Average Precision for each factorization rank ranging from 5 to 50 of each model on the families experiment with the random split. Top-left: $p=0.8$, top-right: $p=0.4$, bottom-left: $p=0.2$, bottom-right: $p=0.1$.}
	\label{fig:exp_families_random}
\end{figure}

In the first random split, we try to evaluate the 
quantity of training data needed to 
learn to reason in genealogies. 
\Cref{fig:exp_families_random} shows the average precision of each
model for ranks ranging from 5 to 50, for each value of $p$.
Only \textsc{ComplEx} and \textsc{RESCAL} are able to generalize almost perfectly with 80\%
of observed data, which first tells us that these models are indeed
capable to learn such genealogical reasonings. 
As many relations are antisymmetric, it is
no surprise that \textsc{DistMult} and \textsc{TransE} cannot reach perfect predictions,
as they already failed in the antisymmetric synthetic relation.

The \textsc{ComplEx} model generalizes quickly with small ranks, 
but is outrun by \textsc{RESCAL}---with small
ranks---and \textsc{TransE} when the percentage of observed data decreases
below $p=0.2$. We conjecture that \textsc{TransE}'s robustness 
is due to its bilinear terms, and
especially the one that involves the subject and the object
embeddings---$e_s\T e_o$---as shown in \Cref{eq:transe2}, 
that can give high scores to pairs of entities belonging 
to the same family.
For \textsc{RESCAL}, its richer representations of relations
by matrices probably help here, as long as the rank
is not too big which clearly causes overfitting.

The \textsc{CP} decomposition scores drop quickly with the proportion
of observed data, because of its unrelated subject and object
representations.
The F model here fails again, for a simple reason:
these relations are exclusive between themselves for a given pair 
of entities $(s,o)$. Indeed, if \tt{father($s,o$)} is true for example,
then none of the other relations between $s$ and $o$ will be
true---at least not in these families. Hence if the F model 
has to predict the score
of the fact $r(s,o)$, it has no other true triple involving $(s,o)$
to support its decision. It will also have troubles 
on the two next splits for the same reason.
Note that in this split, the logic upper-bound is not given as one
would need to know all possible rules to deduce the 17 relations from
each of them---and not only from the four main ones---to compute
this upper-bound.

\subsubsection{Evidence Split}

In this split, we recall that all the \tt{mother}, \tt{father}, \tt{son} and \tt{daughter}
relations are always in the train set for the 5 families. The value of
$p$ ranging from 0.8 to 0.1 corresponds here to the proportion
of the 13 other relations that are also put into the training set.
The test and validation sets are only composed of these
13 relations.

\begin{figure}[ht]
	\centering
	\includegraphics[width=0.422\linewidth]{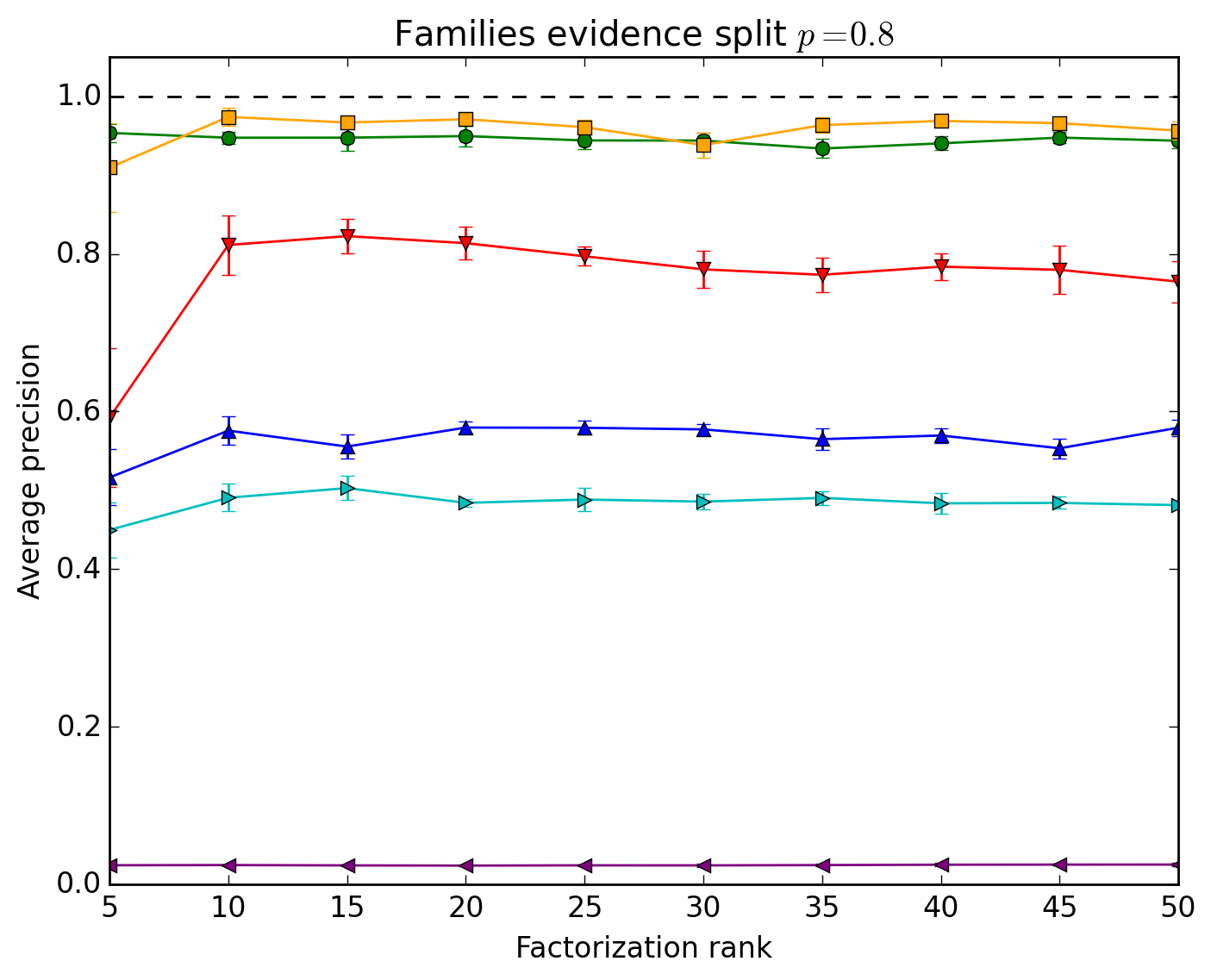}
	\includegraphics[width=0.56\linewidth]{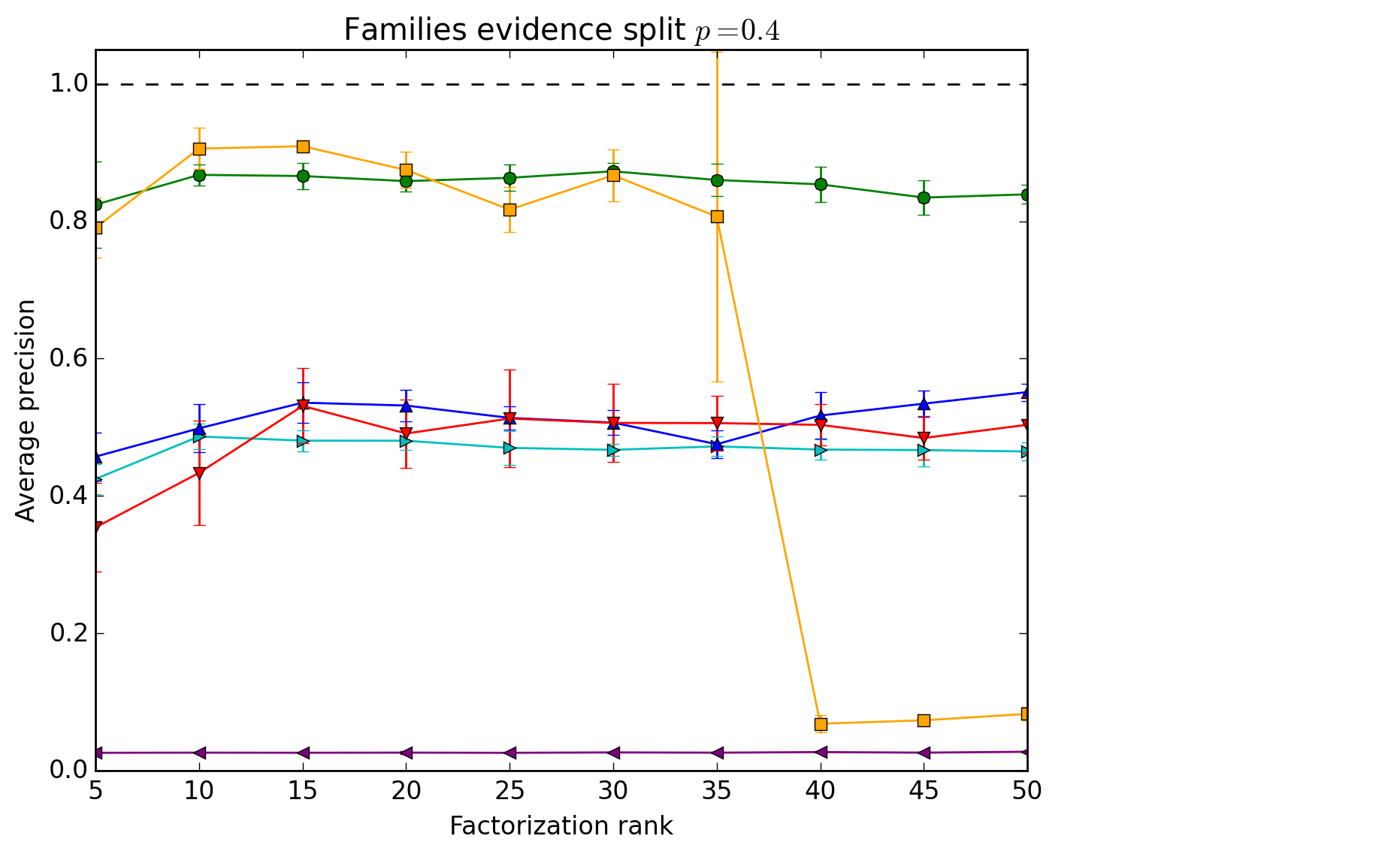}
	\includegraphics[width=0.422\linewidth]{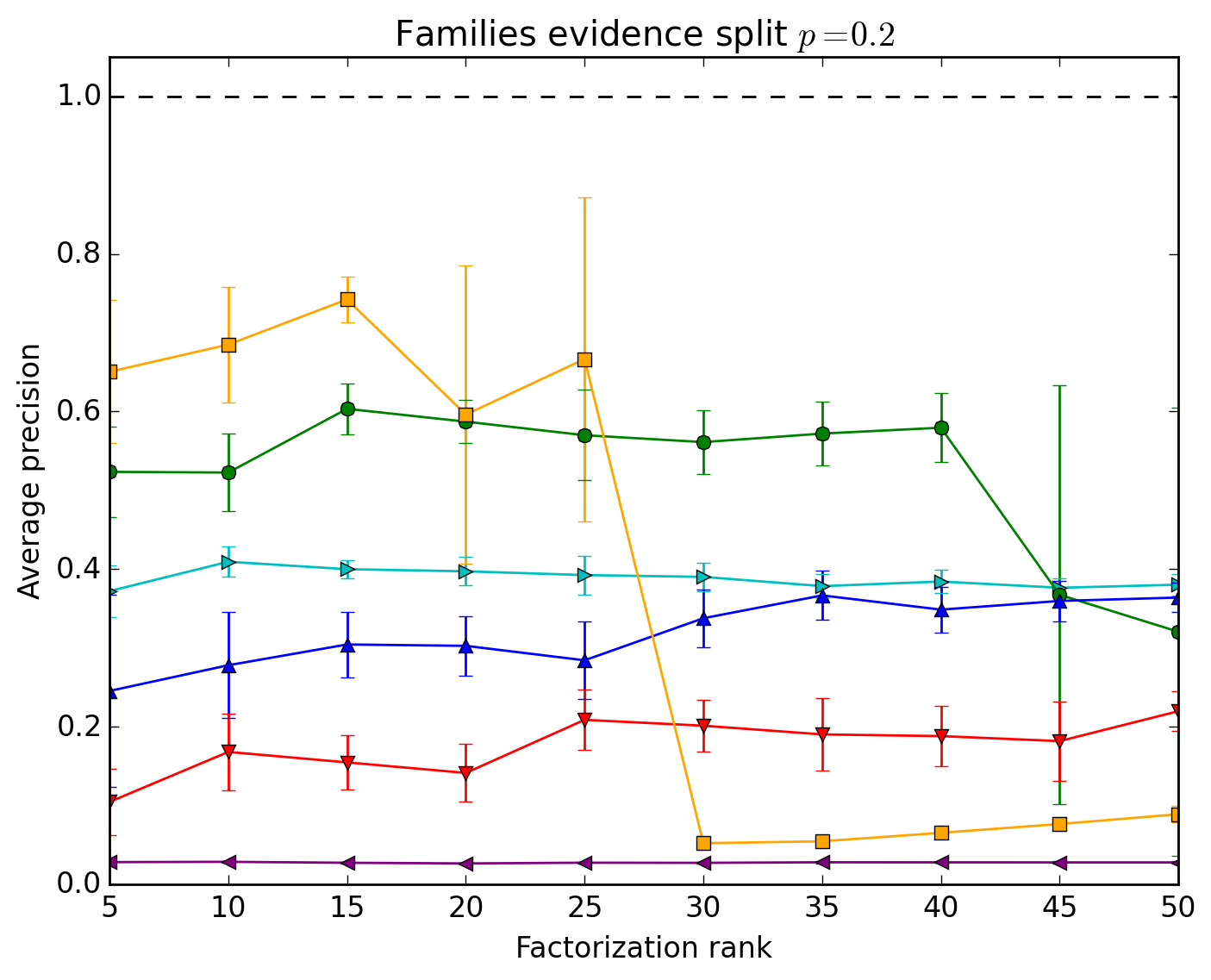}
	\includegraphics[width=0.56\linewidth]{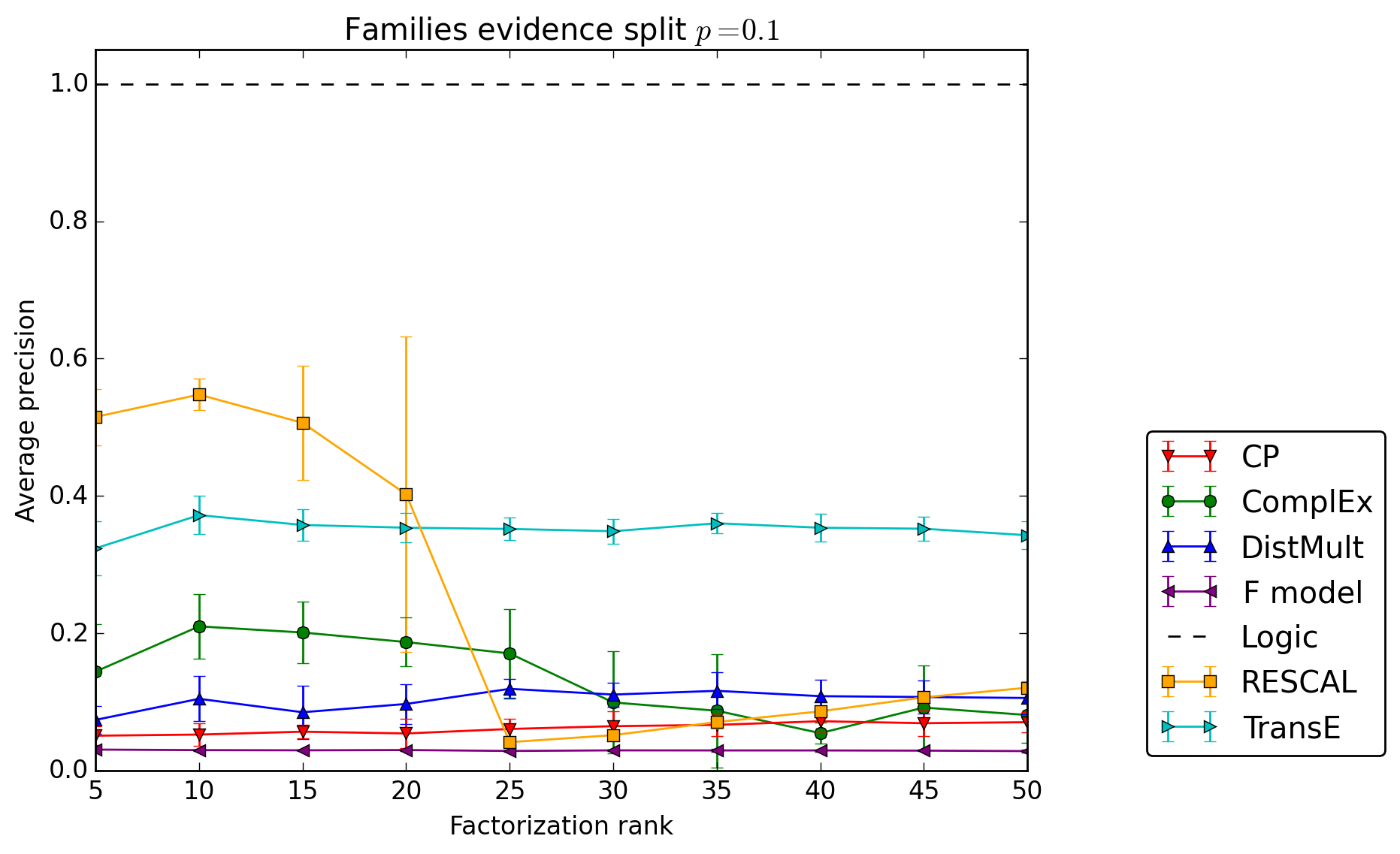}
	\caption{Average Precision for each factorization rank ranging from 5 to 50 of each model on the families experiment with the evidence split. Top-left: $p=0.8$, top-right: $p=0.4$, bottom-left: $p=0.2$, bottom-right: $p=0.1$.}
	\label{fig:exp_families_evidence}
\end{figure}

Compared to the random split setting, we can see in
\Cref{fig:exp_families_evidence}
that the performances
of the models decrease more slowly with the percentage of observed data.
This shows that latent factor models are able to use the 
information provided by these four relations from which all 
of the others can be deduced. 

\textsc{RESCAL} is here clearly the best model for all values
of $p$, as long as $K$ is not too big. It exhibits again
a behavior that seems to have two equilibria
distributed around a pivotal $K$ at which average precision
suddenly drops, with high variance of the predictions around that $K$.
\textsc{ComplEx} also seems to show a lighter overfitting with high 
values of $K$ when $p \leq 0.2$.
\textsc{TransE} confirms an advantage with $p=0.1$ with a notable 
rise of average precision compared to the random split.
\textsc{CP}, \textsc{DistMult} and the F model fail again for the same
reasons as in the random split.

However, given the rules to deduce the 13 other relations from the four main ones,
recall that a logical inference engine is able to reach 
an average precision of one.
Though improvement compared to the random split setup is large,
the gap with logical inference is still wide with $p=0.1$
and $p=0.2$, showing
that latent factor models have troubles making the link between
the four main relations and the 13 other ones when limited training data is available.
This could be due to the imbalance in the number of each relation
in the training set that this split introduces,
biasing the entity embeddings towards a better reconstruction of the 4 main
relations, to the detriment of the generalization over the 
13 remaining ones. Weighting the facts in accordance with the
preponderance of each relation in the dataset could improve
performances here.

\subsubsection{Family Split}

In this last split, all the \tt{mother}, \tt{father}, \tt{son} 
and \tt{daughter} are
in the train set for all families, but also all the 13 other relations of four
out of the five families. The value of $p$ corresponds
here to the amount of the 13 other relations
\emph{of the fifth family only} that are in the training set too.

\begin{figure}[t]
	\centering
	\includegraphics[width=0.422\linewidth]{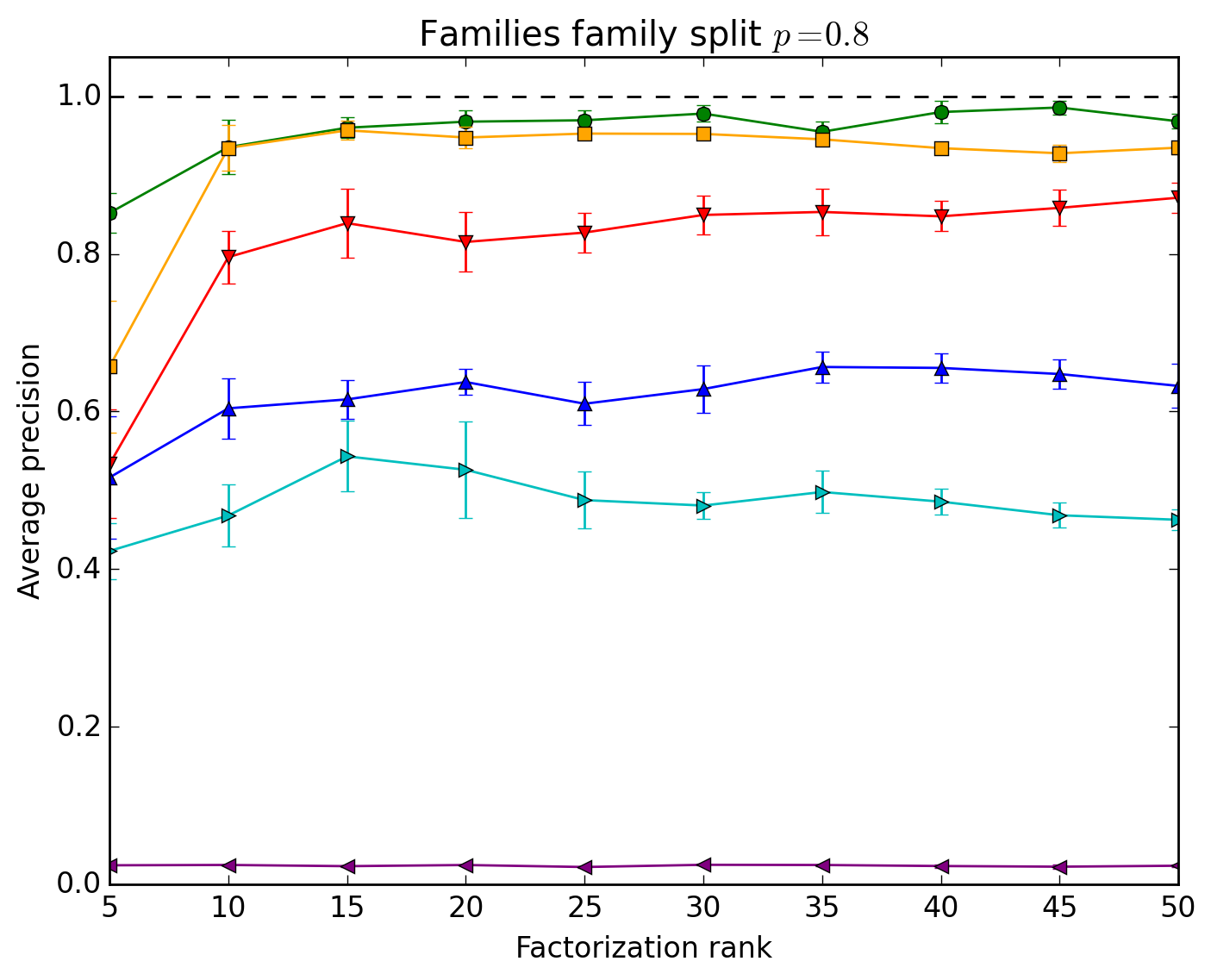}
	\includegraphics[width=0.56\linewidth]{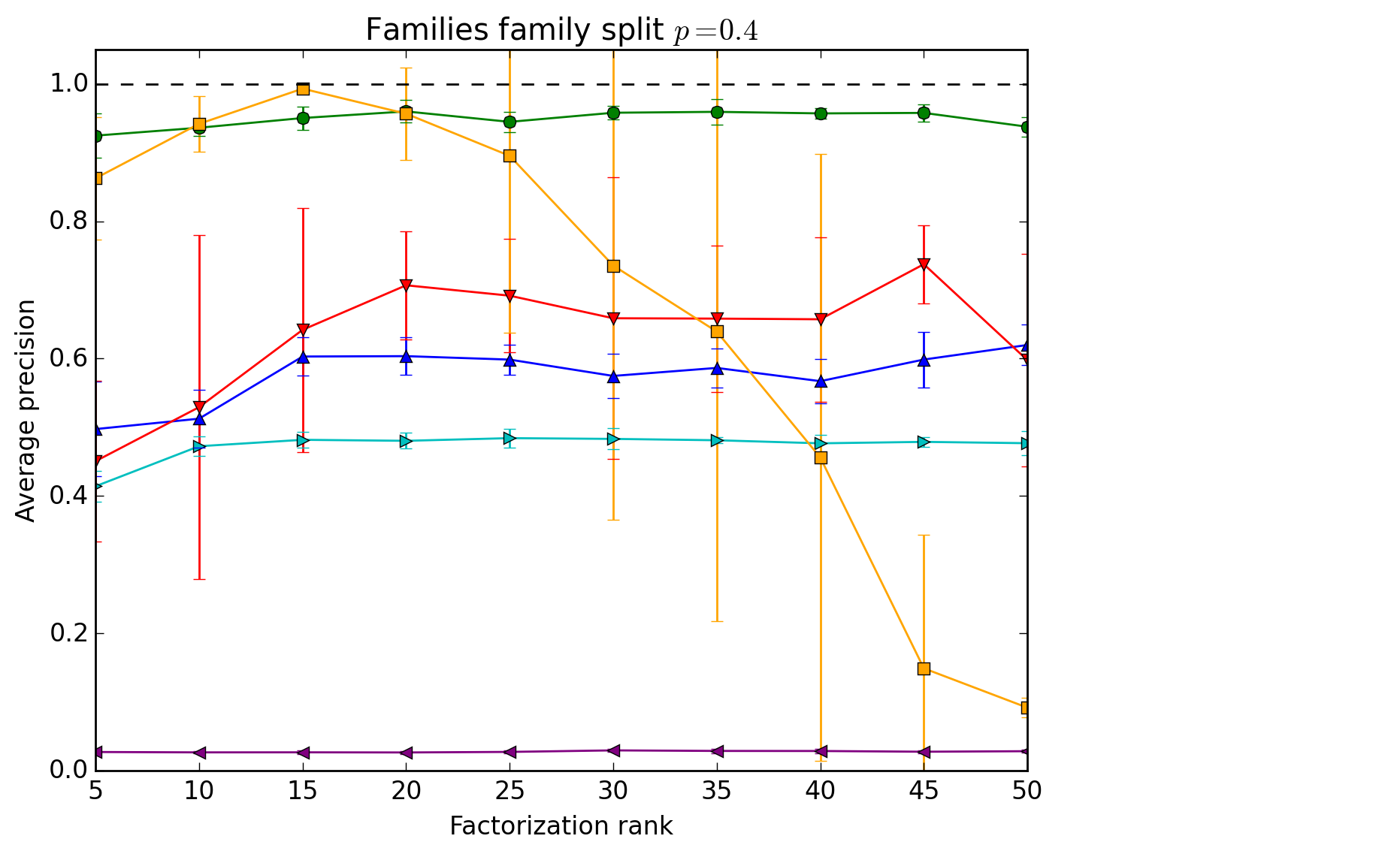}
	\includegraphics[width=0.422\linewidth]{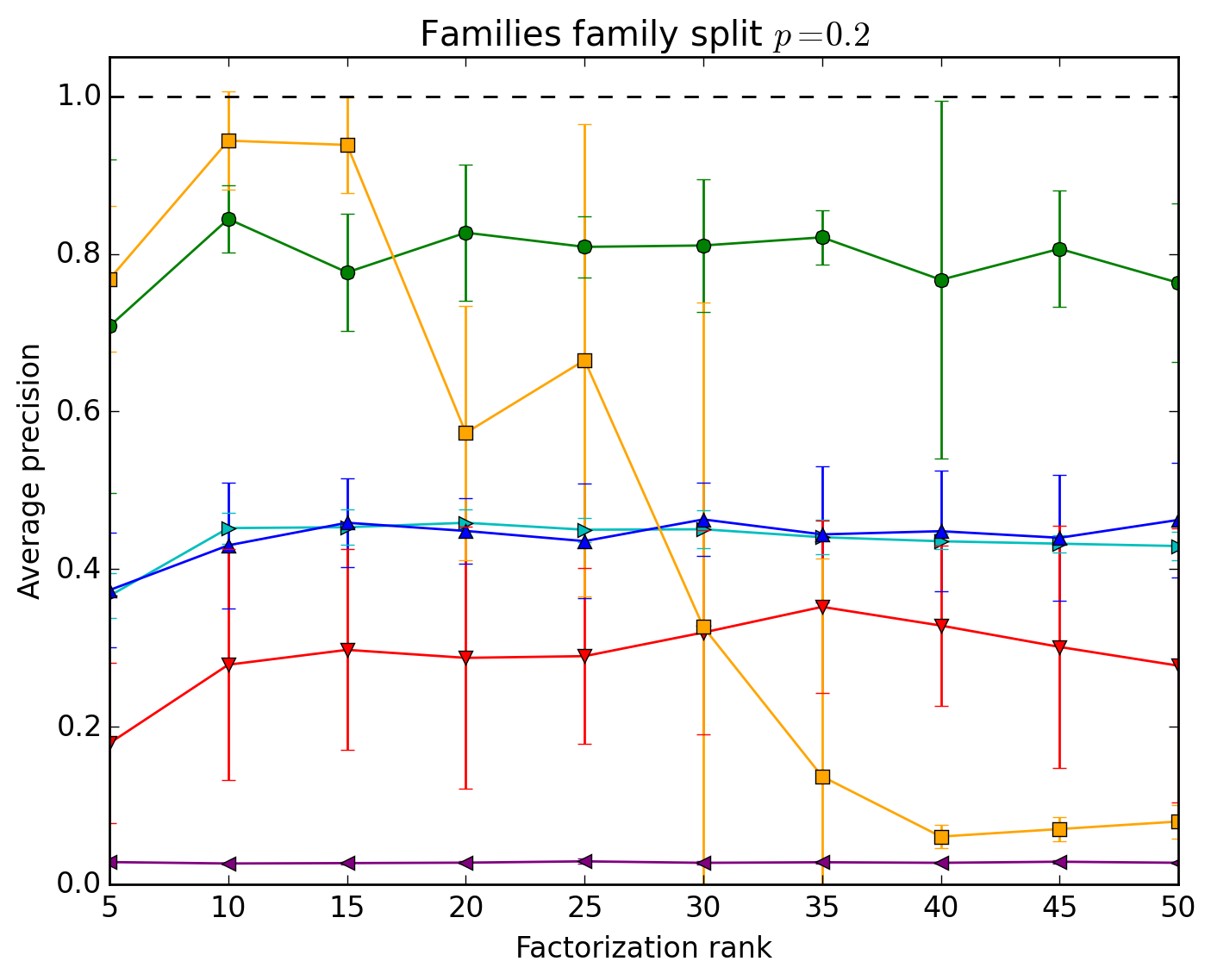}
	\includegraphics[width=0.56\linewidth]{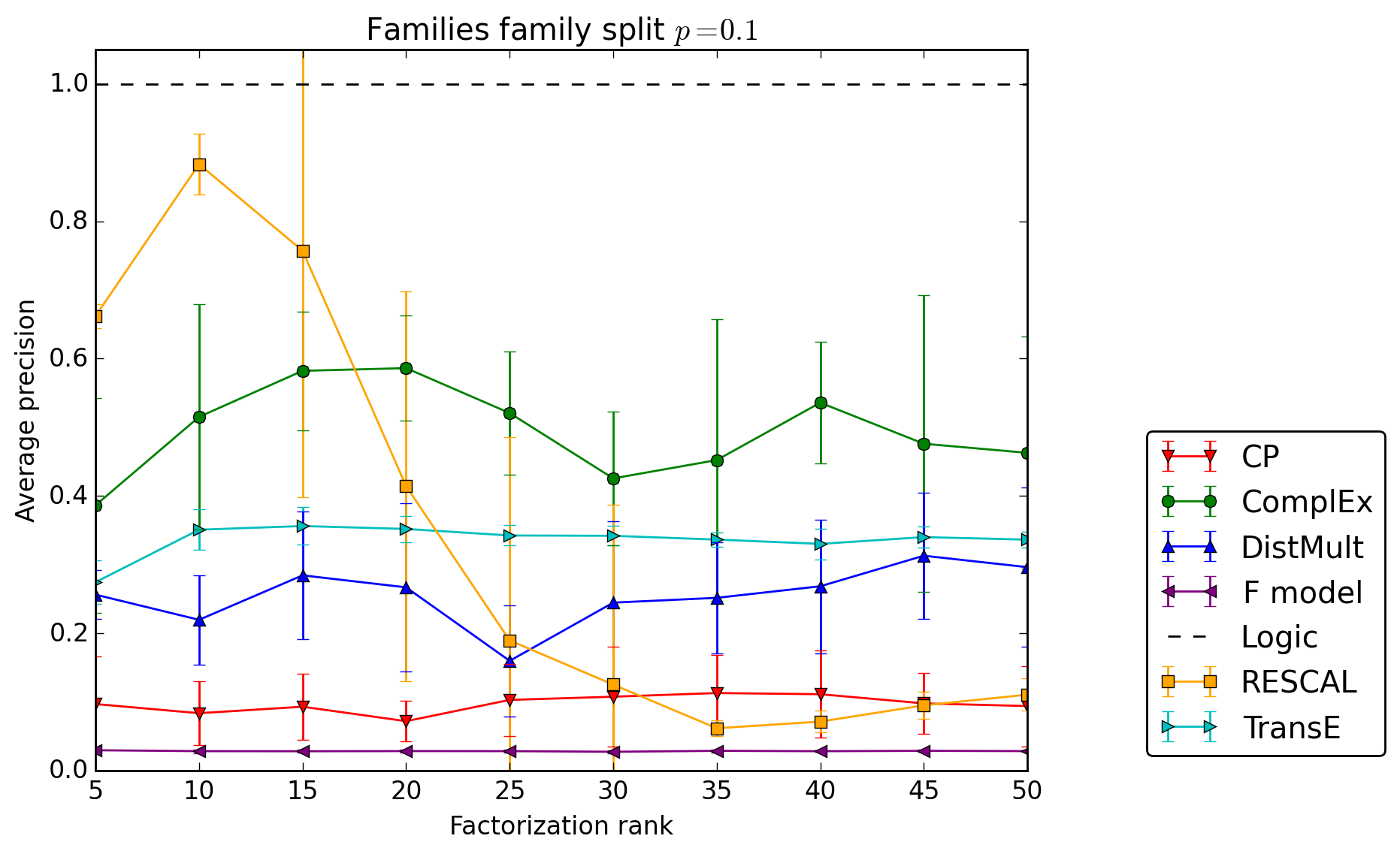}
	\includegraphics[width=0.56\linewidth]{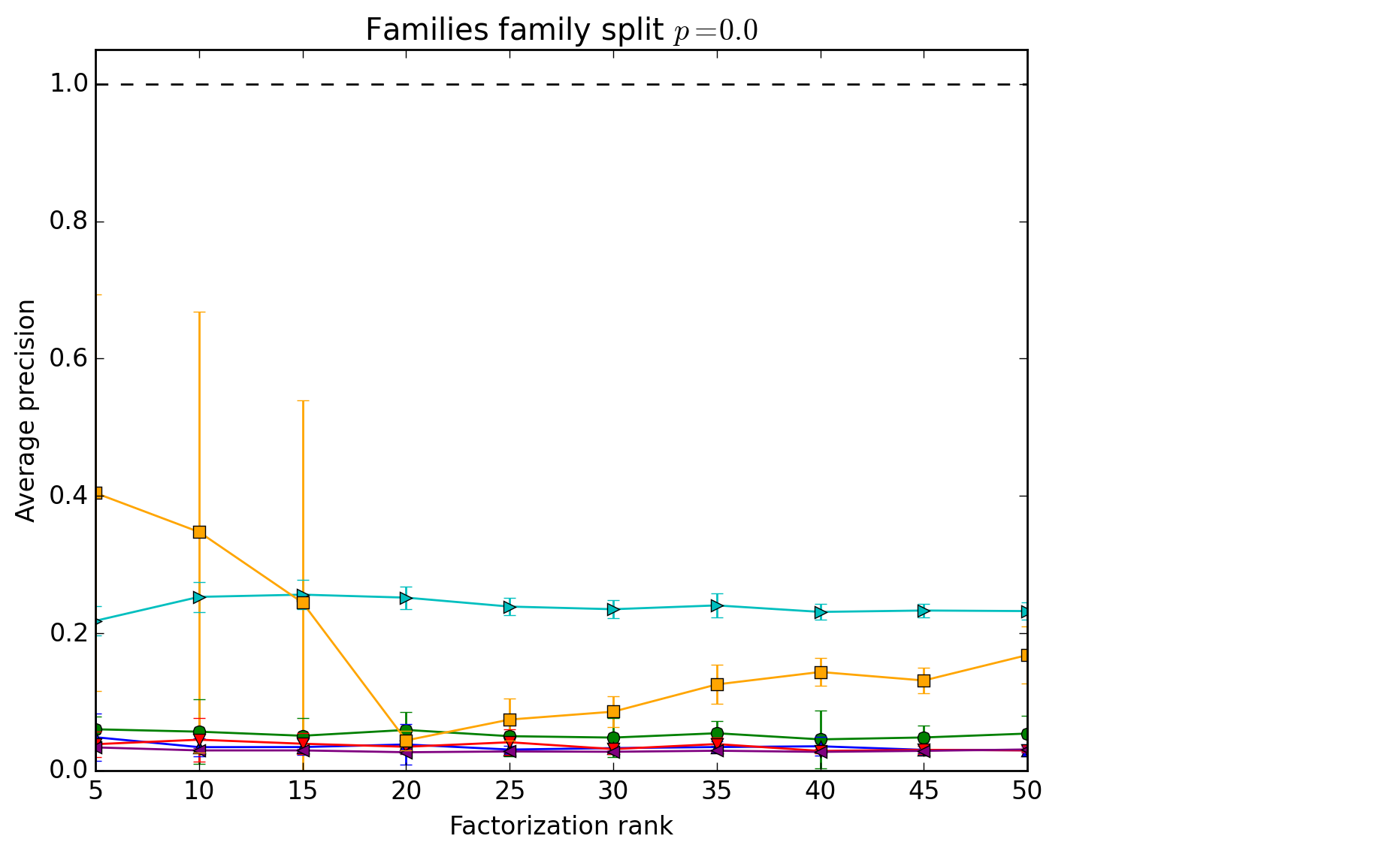}
	\caption{Average Precision for each factorization rank ranging from 5 to 50 of each model on the families experiment with the family split. Top-left: $p=0.8$, top-right: $p=0.4$, middle-left: $p=0.2$, middle-right: $p=0.1$, bottom: $p=0.0$.}
	\label{fig:exp_families_family}
\end{figure}

The curves in \Cref{fig:exp_families_family} show
a clear improvement over the previous ones in \Cref{fig:exp_families_evidence}.
\textsc{RESCAL} is again the best model as it 
reaches average precisions $ \geq 0.9$ even down
to $p=0.1$---with small ranks again. \textsc{ComplEx}
is in these cases the best with high ranks,
though much below \textsc{RESCAL}'s best scores when $p=0.1$.

Does that mean
these models were able to exploit the additional information?
Yes and no. We conjecture that the better results for
$p$ ranging from 0.8 to 0.1 are partly due to the relation imbalance
problem---explained in the previous split---being much smaller 
here, as all the relations of four families are given in
in the training set. 

To ensure that models indeed did not generalized
from the four perfectly informed families, 
we reduced the proportion $p$
of the 13 other relations of the fifth family that are in the training 
set to zero---which thus constitute the whole validation and test sets. 
And though the models are provided with four perfectly informed families,
and all the needed facts
to predict the missing ones in the fifth family, they fail
in this last setting as shown in the bottom plot of \Cref{fig:exp_families_evidence}. \textsc{RESCAL} and \textsc{TransE}
resist better than the other models again in this
last setting with $p=0$.

This is easily explained, as disconnected sets of entities, here families,
correspond to different blocks in the tensor $\ObsTensor$, as shown in
\Cref{fig:family_sets}.
Entities that are in different families $s,o \in \Omega^i$,
$s',o' \in \Omega^j$, $i \neq j$, are never involved together
in an observed fact: $((s,r,o'),y_{sro'}),((s',r,o),y_{s'ro}) \notin \Omega$, 
for any relation $r \in \RelationSpace$.
Thus when learning their embeddings $e_s,e_o$ and $e_{s'},e_{o'}$,
the only link they share is the embedding of the relation $r$
that is involved in the scoring functions $\phi(r,s,o)$ and $\phi(r,s',o')$.
This interpretation is also supported by \textsc{RESCAL} scores, which benefits from
its higher number of parameters of its relation 
representations $W_r \in \R^{\rk \times \rk}$,
which increases the amount of information shared across the
families.

To formally bring this problem to light, let us re-frame
the tensor approximation problem as a system of inequalities.
As $\proba(y_{rso}=1) = \sigma(\phi(r,s,o;\Theta))$,
true triples should have a score $\phi(r,s,o;\Theta) > 0$,
and false triples $\phi(r,s,o;\Theta) < 0$.
For the sake of the example, we will consider
factorizing the two relations \tt{sister} and \tt{grandfather},
with two families of 2 persons each, using the \textsc{DistMult} model.
Observing the true fact \tt{sister(a,b)} or the true fact \tt{sister(b,a)}
where $a,b \in \EntitySpace$,
allows us to deduce that $a$ and $b$ are not the grandfather of each other:
\begin{itemize}
    \item $\mathtt{sister(a,b) \Rightarrow \neg grandfather(a,b)}$
    \item $\mathtt{sister(a,b) \Rightarrow \neg grandfather(b,a)}$
    \item $\mathtt{sister(b,a) \Rightarrow \neg grandfather(a,b)}$
    \item $\mathtt{sister(b,a) \Rightarrow \neg grandfather(b,a)}$
\end{itemize}

Similarly to the family split with $p=0$, let us have both relations fully
observed for a first family that contains entities $a,b \in \EntitySpace^1$,
and only the facts of the relation $\mathtt{sister}$ observed 
for entities of a second family $c,d \in \EntitySpace^2$.
The resulting $2 \times 4 \times 4$ partially observed binary tensor
is:
\begin{equation}
    \mathtt{sister}:
    \begin{blockarray}{ccccc}
         & a & b & c & d \\
        \begin{block}{c(cccc)}
          a & -1 & \phantom{-}1 &  &   \\
          b & \phantom{-}1 & -1 &  &  \\
          c &  &  & -1 & \phantom{-}1 \\
          d &  &  & \phantom{-}1 & -1 \\
        \end{block}
    \end{blockarray}\,,
    \qquad
    \mathtt{grandfather}:
    \begin{blockarray}{ccccc}
         & a & b & c & d \\
        \begin{block}{c(cccc)}
          a & -1 & -1 &  &   \\
          b & -1 & -1 &  &  \\
          c &  &  & \cdot & \cdot \\
          d &  &  & \cdot & \cdot \\
        \end{block}
    \end{blockarray}
\end{equation}
where $\cdot$ and empty spaces are unobserved facts.
From the first, fully observed family we wish
to learn the above rules and the irreflexivity of the \tt{grandfather}
relation, to correctly complete the \tt{grandfather} facts between entities $c$ and $d$.

As the observed blocks---and the block we wish to recover---are
symmetric here, there is no expressiveness issue with
using \textsc{DistMult}.
Decomposing this tensor with the \textsc{DistMult}
model with $K=2$ such that true facts have probability $\proba(y_{rso}=1) > 0.5$
and false facts have probability $\proba(y_{rso}=1) < 0.5$, 
amounts to solving the following system of inequalities:
\begin{equation}
\begin{cases}
    w_{s1}^{\phantom{2}} e_{a1}^2 + w_{s2}^{\phantom{2}} e_{a2}^2< 0 \\
    w_{s1}^{\phantom{2}} e_{b1}^2 + w_{s2}^{\phantom{2}} e_{b2}^2< 0 \\
    w_{s1} e_{a1} e_{b1} + w_{s2} e_{a2} e_{b2} > 0 \\
    w_{g1}^{\phantom{2}} e_{a1}^2 + w_{g2}^{\phantom{2}} e_{a2}^2< 0 \\
    w_{g1}^{\phantom{2}} e_{b1}^2 + w_{g2}^{\phantom{2}} e_{b2}^2< 0 \\
    w_{g1} e_{a1} e_{b1} + w_{g2} e_{a2} e_{b2} < 0 \\\\
    w_{s1}^{\phantom{2}} e_{c1}^2 + w_{s2}^{\phantom{2}} e_{c2}^2< 0 \\
    w_{s1}^{\phantom{2}} e_{d1}^2 + w_{s2}^{\phantom{2}} e_{d2}^2< 0 \\
    w_{s1} e_{c1} e_{d1} + w_{s2} e_{c2} e_{d2} > 0 
\end{cases}
\label{eq:ineq_example_family}
\end{equation}
where $e_i \in \R^2$ is the embedding 
of entity $i \in \EntitySpace$,
$w_s \in \R^2$ is the embedding of the relation 
\tt{sister}, and $w_g \in \R^2$ is the embedding of the relation 
\tt{grandfather}. The six first inequalities involve the
entities $a$ and $b$, and the three lower ones involve
the entities $c$ and $d$.

Correctly reconstructing the \tt{grandfather} facts between $c$ and $d$
would thus require their embeddings to satisfy the same
three additional inequalities:
\begin{equation}
\begin{cases}
    w_{g1}^{\phantom{2}} e_{c1}^2 + w_{g2}^{\phantom{2}} e_{c2}^2< 0 \\
    w_{g1}^{\phantom{2}} e_{d1}^2 + w_{g2}^{\phantom{2}} e_{d2}^2< 0 \\
    w_{g1} e_{c1} e_{d1} + w_{g2} e_{c2} e_{d2} < 0 
\end{cases}
\,.
\label{eq:ineq_sol_example_family}
\end{equation}
However, it is easy to check that arbitrary solutions to the system (\ref{eq:ineq_example_family})
for $e_c$ and $e_d$ does not necessarily satisfy
the system (\ref{eq:ineq_sol_example_family}),
and hence does not necessarily predict the \tt{grandfather} facts between $c$ and $d$
correctly.
Also, this would be true even if we added
more families like $a$ and $b$ with both relations
fully observed, as this would not add more constraints
on $e_c$ and $e_d$.

This explains why all models fail in the family split
with $p=0$: nothing encourages less constrained entities
to have embeddings that resemble the ones of similar,
more constrained entities; and adding more examples
of more constrained entities does not help.\\

\textbf{Family Experiments Summary:}
\begin{itemize}
    \item \textsc{RESCAL} is the best model in all different splits, but overfits with a too big $K$.
    \item \textsc{RESCAL} and \textsc{TransE} are the most robust to missing data.
    \item \textsc{ComplEx} behaves well with more data and hardly overfits.
    \item Relation imbalance in the training set can be a problem when the test set is distributed differently, and could be easily fixed by weighting the facts accordingly.
    \item The absence of explicit parameter sharing between entity representations prevents knowledge transfer between disjoint sets of entities.
\end{itemize}

\section{Future Research Directions}
\label{sec:research_dir}

Overall, the \textsc{ComplEx} model proved to have the more stable 
generalization abilities across all the synthetic
experiments. 
Most models showed a good ability to learn basic
relation properties, except on antisymmetry where
only \textsc{ComplEx} succeeded. This said, when decreasing 
the size of the training set down to 10\% on joint learning
of the relation properties,
the best models were 10 points of average precision
behind the best possible score. 
Improving models towards learning basic binary relation properties from less data
thus seems a promising direction.

Some models showed their advantages in
some specific settings. \textsc{RESCAL} and \textsc{TransE} 
showed a good robustness
when a lot of data is missing in the family experiments,
thanks to the bilinear terms for \textsc{TransE}, and the
rich matrix relation representations of \textsc{RESCAL}.
The F model was not fit for these experiments,
but its pairwise terms are known to give it an advantage
for non-compositional pairs of entities \cite{Welbl2016}.

Different possible combinations seem promising.
The behaviour of \textsc{RESCAL} and \textsc{ComplEx} on symmetric and
antisymmetric experiments suggests that encoding these
patterns through complex conjugation is more stable
than using the non-commutative matrix product.
But \textsc{RESCAL}'s matrix representations of relations helped
a lot in the family experiments, as long as the rank was not too
high, suggesting that there might 
be a middle ground between $K$ and $K^2$ to be found 
for the parametric representation of the relations.
Using tridiagonal or pentadiagonal (or more) symmetric matrices 
for relation representations
within the \textsc{ComplEx} model could be an answer to these problems.

Combining the scoring functions of the \textsc{TransE} and F
models with \textsc{ComplEx} could also lead to a more 
robust model.
The combination of bilinear and trilinear terms
has already been explored within real-valued models \cite{garcia2014effective},
also with vectorial weights over each term \cite{Jenatton2012},
as well as combining different pairwise terms \cite{Welbl2016,Singh2015},
which yielded better performance in all cases.

The main defect of latent factor models that this experimental
survey points to is their low ability to transfer knowledge
between disjoint set of entities, as shown in the last
family split with $p=0$. Real knowledge graphs
might not have fully disjoint subsets, 
but rather some less-connected sub-graphs, between which
this effect is likely to appear too. We believe improving this ability
of latent factor models is key. 

One already-pursued way
to harness this problem is to enable latent factor models to make use
of logic rules \cite{Rocktaschel2015,demeester2016lifted}.
As already said, those rules are not always available, and thus
latent factor models should be improved in order to have this 
ability to learn from disjoint subsets, 
while still operating without rules.

Intuitively, sharing parameters across all entity representations
could also solve this issue, as used in Bayesian
clustered factorization models \cite{sutskever2009modelling}.
Though those models have known scalability issues.
A possible, more scalable way to implement a shared parametrization between
the entity embeddings $E \in \C^{\Ne \times \rk}$ is through
a nested factorization, where the matrix $E$ is itself expressed
as a low-rank factorization, as it has already been proposed for
the relation embeddings \cite{Jenatton2012}. Another one
could be a suited regularization over the whole matrix $E$:
in most proposals $E$ is regularized row-wise with $||e_i||^2_2$ for
all $i \in \EntitySpace$---as shown in
\Cref{eq:objective_func}.

Another linked limitation of latent factor models---that does not
require experiments to be shown---is their inability to generalize 
to new entities without retraining. Indeed for new facts involving a new entity
$i$, its embedding $e_i \in \C^\rk$ is unknown.
But in a logic-based setting, only the new facts involving the new entity
are necessary to infer other facts from known rules. 
Some recent works started tackling this problem:
\citet{verga2017generalizing} proposed a solution for the F model,
by expressing entity pair embeddings as combinations
of the relation embeddings in which they appear.
\citet{hamaguchi2017knowledge} used graph neural networks
to handle unseen entities at test time.


The evidence split in the family experiments also pointed
out a potential problem of imbalance in the distribution
of the relations across the facts when the train and test sets
are distributed differently. Correcting this imbalance
via down-weighting the facts involving the most frequent 
relations could be a solution, as well as sharing the 
parametrization between the relations.

A non-mentioned aspect of the problem in this paper is the theoretical
learnability of such logic formulas, a field that has been extensively
covered 
\cite{valiant1984theory,kearns1994cryptographic,muggleton1994inductive,dzeroski1994inductive}.
However logic learnability by latent factor models has not yet been
specifically studied. Recently established links between sign-matrices
complexity---specifically the sign-rank \cite{linial2007complexity}---and VC-dimension open the door to such
theoretical study \cite{alon2016sign}, and possible extensions
to the tensor case. This being said, theoretical guarantees
generally come under the condition that the training and test 
sets are drawn from the same distribution, which is not the case in the
last two splits of the family experiments: a theoretical analysis of the 
learnability of such cases might require a new theoretical framework 
for statistical learning.

\section{Conclusion}

We experimentally surveyed state-of-the-art latent factor models
for link prediction in knowledge graphs, in order
to assess their ability to learn (i) binary relation
properties, and (ii) genealogical relations, directly from observed facts,
as well as their robustness to missing data.
Latent factor models yield good performances in the first case,
while having more difficulties in the second one. Specifically,
we show that such models do not reason as it is generally
meant for logical inference engines, as they
are unable to transfer their predictive abilities 
between disjoint subsets of entities.
The different behaviors of the models in each experimental setup
suggest possible enhancements and research directions, 
including combining them, as well as
it exposes each model's advantages and limitations.



\section*{Acknowledgments}
This work was supported in part by the Association Nationale de la
Recherche et de la Technologie through the CIFRE grant 2014/0121.

\bibliographystyle{apalike}
\bibliography{nonauto_bib,complex_bib}

\begin{thebibliography}{}

\bibitem[Aaronson, 2013]{aaronson2011philosophers}
Aaronson, S. (2013).
\newblock Why philosophers should care about computational complexity.
\newblock In B.~J.~Copeland, C.~P. and Shagrir, O., editors, {\em
  Computability: Turing, G\"odel, Church, and Beyond}, pages 261--328. MIT
  Press.

\bibitem[Alon et~al., 2016]{alon2016sign}
Alon, N., Moran, S., and Yehudayoff, A. (2016).
\newblock Sign rank versus vc dimension.
\newblock In {\em Conference on Learning Theory}, pages 47--80.

\bibitem[Auer et~al., 2007]{dbpedia}
Auer, S., Bizer, C., Kobilarov, G., Lehmann, J., and Ives, Z. (2007).
\newblock {DBpedia}: A nucleus for a web of open data.
\newblock In {\em International Semantic Web Conference, Busan, Korea}, pages
  11--15. Springer.

\bibitem[Bollacker et~al., 2008]{Bollacker2008}
Bollacker, K., Evans, C., Paritosh, P., Sturge, T., and Taylor, J. (2008).
\newblock {Freebase: a collaboratively created graph database for structuring
  human knowledge}.
\newblock In {\em ACM SIGMOD International Conference on Management of Data},
  pages 1247--1250.

\bibitem[Bordes et~al., 2013a]{bordes2013irreflexive}
Bordes, A., Usunier, N., Garcia-Duran, A., Weston, J., and Yakhnenko, O.
  (2013a).
\newblock Irreflexive and hierarchical relations as translations.
\newblock {\em arXiv preprint arXiv:1304.7158}.

\bibitem[Bordes et~al., 2013b]{bordes2013translating}
Bordes, A., Usunier, N., Garcia-Duran, A., Weston, J., and Yakhnenko, O.
  (2013b).
\newblock Translating embeddings for modeling multi-relational data.
\newblock In {\em Advances in Neural Information Processing Systems}, pages
  2787--2795.

\bibitem[Bordes et~al., 2011]{bordes2011learning}
Bordes, A., Weston, J., Collobert, R., and Bengio, Y. (2011).
\newblock Learning structured embeddings of knowledge bases.
\newblock In {\em {AAAI} Conference on Artificial Intelligence}.

\bibitem[Bottou, 2014]{bottou2014machine}
Bottou, L. (2014).
\newblock From machine learning to machine reasoning.
\newblock {\em Machine Learning}, 94(2):133--149.

\bibitem[Bouchard et~al., 2015]{bouchard2015approximate}
Bouchard, G., Singh, S., and Trouillon, T. (2015).
\newblock On approximate reasoning capabilities of low-rank vector spaces.
\newblock {\em AAAI Spring Symposium on Knowledge Representation and Reasoning:
  Integrating Symbolic and Neural Approaches}.

\bibitem[Bowman et~al., 2015]{bowman2015recursive}
Bowman, S.~R., Potts, C., and Manning, C.~D. (2015).
\newblock Recursive neural networks can learn logical semantics.
\newblock In {\em ACL Workshop on Continuous Vector Space Models and their
  Compositionality}.

\bibitem[Cardano, 1545]{cardano1545}
Cardano, G. (1545).
\newblock {\em Artis Magn\ae, Sive de Regulis Algebraicis Liber Unus}.

\bibitem[Caroll and Chang, 1970]{cc-candecomp-1970}
Caroll, J.~D. and Chang, J.-J. (1970).
\newblock Analysis of individual differences in multidimensional scaling via
  $n$-way generalization of {Eckart--Young} decomposition.
\newblock {\em Psychometrika}, 35:283--319.

\bibitem[Cyganiak et~al., 2014]{cyganiak2014rdf}
Cyganiak, R., Wood, D., and Lanthaler, M. (2014).
\newblock {RDF} 1.1 concepts and abstract syntax.
\newblock {\em W3C Recommendation}.

\bibitem[Davis et~al., 1993]{davis1993knowledge}
Davis, R., Shrobe, H., and Szolovits, P. (1993).
\newblock What is a knowledge representation?
\newblock {\em AI magazine}, 14(1):17--33.

\bibitem[Demeester et~al., 2016]{demeester2016lifted}
Demeester, T., Rockt{\"a}schel, T., and Riedel, S. (2016).
\newblock Lifted rule injection for relation embeddings.
\newblock In {\em Empirical Methods in Natural Language Processing}, pages
  1389--1399.

\bibitem[Dong et~al., 2014]{dong2014_knowledgevault}
Dong, X., Gabrilovich, E., Heitz, G., Horn, W., Lao, N., Murphy, K., Strohmann,
  T., Sun, S., and Zhang, W. (2014).
\newblock Knowledge vault: A web-scale approach to probabilistic knowledge
  fusion.
\newblock In {\em ACM SIGKDD International Conference on Knowledge Discovery
  and Data Mining}, pages 601--610.

\bibitem[Duchi et~al., 2011]{duchi2011adaptive}
Duchi, J., Hazan, E., and Singer, Y. (2011).
\newblock Adaptive subgradient methods for online learning and stochastic
  optimization.
\newblock {\em Journal of Machine Learning Research}, 12:2121--2159.

\bibitem[Dzeroski and Lavrac, 1994]{dzeroski1994inductive}
Dzeroski, S. and Lavrac, N. (1994).
\newblock {\em Inductive logic programming: techniques and applications}.
\newblock Ellis Horwood, New York.

\bibitem[Friedman et~al., 1999]{friedman1999learning}
Friedman, N., Getoor, L., Koller, D., and Pfeffer, A. (1999).
\newblock Learning probabilistic relational models.
\newblock In {\em International Joint Conference on Artificial Intelligence},
  volume~99, pages 1300--1309.

\bibitem[Gal{\'a}rraga et~al., 2015]{galarraga2015fast}
Gal{\'a}rraga, L., Teflioudi, C., Hose, K., and Suchanek, F.~M. (2015).
\newblock Fast rule mining in ontological knowledge bases with amie+.
\newblock {\em The VLDB Journal}, 24(6):707--730.

\bibitem[Garc{\'\i}a-Dur{\'a}n et~al., 2014]{garcia2014effective}
Garc{\'\i}a-Dur{\'a}n, A., Bordes, A., and Usunier, N. (2014).
\newblock Effective blending of two and three-way interactions for modeling
  multi-relational data.
\newblock In {\em Joint European Conference on Machine Learning and Knowledge
  Discovery in Databases}, pages 434--449. Springer.

\bibitem[Getoor and Taskar, 2007]{Getoor2007}
Getoor, L. and Taskar, B. (2007).
\newblock {\em Introduction to Statistical Relational Learning}.
\newblock MIT Press.

\bibitem[Grefenstette, 2013]{grefenstette2013towards}
Grefenstette, E. (2013).
\newblock Towards a formal distributional semantics: Simulating logical calculi
  with tensors.
\newblock In {\em Joint Conference on Lexical and Computational Semantics}.

\bibitem[Halmos, 1998]{halmos1998naive}
Halmos, P.~R. (1998).
\newblock {\em Naive set theory}.
\newblock Springer Science \& Business Media.

\bibitem[Hamaguchi et~al., 2017]{hamaguchi2017knowledge}
Hamaguchi, T., Oiwa, H., Shimbo, M., and Matsumoto, Y. (2017).
\newblock Knowledge transfer for out-of-knowledge-base entities: A graph neural
  network approach.
\newblock {\em arXiv preprint arXiv:1706.05674}.

\bibitem[Harshman, 1970]{harshman-parafac-1970}
Harshman, R.~A. (1970).
\newblock Foundations of the {PARAFAC} procedure: models and conditions for an
  explanatory multimodal factor analysis.
\newblock {\em UCLA Working Papers in Phonetics}, 16:1--84.

\bibitem[Hayashi and Shimbo, 2017]{hayashi2017equivalence}
Hayashi, K. and Shimbo, M. (2017).
\newblock On the equivalence of holographic and complex embeddings for link
  prediction.
\newblock {\em arXiv preprint arXiv:1702.05563}.

\bibitem[Heckerman et~al., 2007]{heckerman2007probabilistic}
Heckerman, D., Meek, C., and Koller, D. (2007).
\newblock Probabilistic entity-relationship models, prms, and plate models.
\newblock {\em Introduction to statistical relational learning}, pages
  201--238.

\bibitem[Hinton, 1986]{Hinton1986}
Hinton, G.~E. (1986).
\newblock Learning distributed representation of concepts.
\newblock In {\em Conference of the Cognitive Science Society}.

\bibitem[Hitchcock, 1927]{hitchcock-sum-1927}
Hitchcock, F.~L. (1927).
\newblock The expression of a tensor or a polyadic as a sum of products.
\newblock {\em Journal of Mathematical Physics}, 6(1):164--189.

\bibitem[Jenatton et~al., 2012]{Jenatton2012}
Jenatton, R., Bordes, A., Le~Roux, N., and Obozinski, G. (2012).
\newblock A latent factor model for highly multi-relational data.
\newblock In {\em Advances in Neural Information Processing Systems}, pages
  3167--3175.

\bibitem[Kearns and Valiant, 1994]{kearns1994cryptographic}
Kearns, M. and Valiant, L. (1994).
\newblock Cryptographic limitations on learning boolean formulae and finite
  automata.
\newblock {\em Journal of the ACM}, 41(1):67--95.

\bibitem[Kersting and De~Raedt, 2001]{kersting2001towards}
Kersting, K. and De~Raedt, L. (2001).
\newblock Towards combining inductive logic programming with bayesian networks.
\newblock In {\em International Conference on Inductive Logic Programming},
  pages 118--131.

\bibitem[Koren, 2008]{koren2}
Koren, Y. (2008).
\newblock Factorization meets the neighborhood: A multifaceted collaborative
  filtering model.
\newblock In {\em ACM SIGKDD International Conference on Knowledge Discovery
  and Data Mining}, pages 426--434.

\bibitem[Koren et~al., 2009]{koren_netflix}
Koren, Y., Bell, R., and Volinsky, C. (2009).
\newblock Matrix factorization techniques for recommender systems.
\newblock {\em Computer}, 42(8):30--37.

\bibitem[Lewis and Steedman, 2013]{lewis2013combining}
Lewis, M. and Steedman, M. (2013).
\newblock Combining distributional and logical semantics.
\newblock {\em Transactions of the Association for Computational Linguistics},
  1:179--192.

\bibitem[Linial et~al., 2007]{linial2007complexity}
Linial, N., Mendelson, S., Schechtman, G., and Shraibman, A. (2007).
\newblock Complexity measures of sign matrices.
\newblock {\em Combinatorica}, 27(4):439--463.

\bibitem[Lisi, 2010]{lisi2010inductive}
Lisi, F.~A. (2010).
\newblock Inductive logic programming in databases: From datalog to.
\newblock {\em Theory and Practice of Logic Programming}, 10(3):331--359.

\bibitem[Ma et~al., 2015]{ma2015knowledge}
Ma, Y., Crook, P.~A., Sarikaya, R., and Fosler-Lussier, E. (2015).
\newblock Knowledge graph inference for spoken dialog systems.
\newblock In {\em IEEE International Conference on Acoustics, Speech and Signal
  Processing}, pages 5346--5350.

\bibitem[Minervini et~al., 2017]{minervini2017adversarial}
Minervini, P., Demeester, T., Rockt{\"a}schel, T., and Riedel, S. (2017).
\newblock Adversarial sets for regularising neural link predictors.
\newblock In {\em Conference on Uncertainty in Artificial Intelligence}.

\bibitem[Muggleton, 1995]{muggleton1995inverse}
Muggleton, S. (1995).
\newblock Inverse entailment and progol.
\newblock {\em New generation computing}, 13(3-4):245--286.

\bibitem[Muggleton and De~Raedt, 1994]{muggleton1994inductive}
Muggleton, S. and De~Raedt, L. (1994).
\newblock Inductive logic programming: Theory and methods.
\newblock {\em The Journal of Logic Programming}, 19:629--679.

\bibitem[Ngo and Haddawy, 1997]{ngo1997answering}
Ngo, L. and Haddawy, P. (1997).
\newblock Answering queries from context-sensitive probabilistic knowledge
  bases.
\newblock {\em Theoretical Computer Science}, 171(1):147--177.

\bibitem[Nickel et~al., 2016a]{nickel_2016_review}
Nickel, M., Murphy, K., Tresp, V., and Gabrilovich, E. (2016a).
\newblock A review of relational machine learning for knowledge graphs.
\newblock {\em Proceedings of the {IEEE}}, 104(1):11--33.

\bibitem[Nickel et~al., 2016b]{nickel_2016_holographic}
Nickel, M., Rosasco, L., and Poggio, T.~A. (2016b).
\newblock Holographic embeddings of knowledge graphs.
\newblock In {\em {AAAI} Conference on Artificial Intelligence}, pages
  1955--1961.

\bibitem[Nickel and Tresp, 2013]{nickel2013logistic}
Nickel, M. and Tresp, V. (2013).
\newblock Logistic tensor factorization for multi-relational data.
\newblock {\em arXiv preprint arXiv:1306.2084}.

\bibitem[Nickel et~al., 2011]{Nickel2011}
Nickel, M., Tresp, V., and Kriegel, H.-P. (2011).
\newblock A three-way model for collective learning on multi-relational data.
\newblock In {\em International Conference on Machine Learning}, pages
  809--816.

\bibitem[Popper, 1934]{popper1934}
Popper, K. (1934).
\newblock {\em Logik der Forschung}.
\newblock Mohr Siebeck.

\bibitem[Richardson and Domingos, 2006]{richardson2006markov}
Richardson, M. and Domingos, P. (2006).
\newblock Markov logic networks.
\newblock {\em Machine Learning}, 62(1-2):107--136.

\bibitem[Riedel et~al., 2013]{riedel_2013_univschema}
Riedel, S., Yao, L., McCallum, A., and Marlin, B.~M. (2013).
\newblock Relation extraction with matrix factorization and universal schemas.
\newblock In {\em North American Chapter of the Association of Computational
  Linguistics: Human Language Technologies}, pages 74--84.

\bibitem[Rockt{\"a}schel et~al., 2014]{rocktaschel2014low}
Rockt{\"a}schel, T., Bosnjak, M., Singh, S., and Riedel, S. (2014).
\newblock Low-dimensional embeddings of logic.
\newblock In {\em Workshop on semantic parsing at ACL}.

\bibitem[Rockt{\"a}schel and Riedel, 2016]{rocktaschel2016learning}
Rockt{\"a}schel, T. and Riedel, S. (2016).
\newblock Learning knowledge base inference with neural theorem provers.
\newblock {\em Workshop on Automated Knowledge Base Construction at NAACL-HLT},
  pages 45--50.

\bibitem[Rockt{\"a}schel et~al., 2015]{Rocktaschel2015}
Rockt{\"a}schel, T., Singh, S., and Riedel, S. (2015).
\newblock Injecting logical background knowledge into embeddings for relation
  extraction.
\newblock In {\em North American Chapter of the Association for Computational
  Linguistics: Human Language Technologies}, pages 1119--1129.

\bibitem[Singh et~al., 2015]{Singh2015}
Singh, S., Rockt\"{a}schel, T., and Riedel, S. (2015).
\newblock Towards combined matrix and tensor factorization for universal schema
  relation extraction.
\newblock In {\em Workshop on Vector Space Modeling for Natural Language
  Processing at NAACL-HLT}, pages 135--142.

\bibitem[Smolensky et~al., 2016]{smolensky2016basic}
Smolensky, P., Lee, M., He, X., Yih, W.-t., Gao, J., and Deng, L. (2016).
\newblock Basic reasoning with tensor product representations.
\newblock {\em arXiv preprint arXiv:1601.02745}.

\bibitem[Socher et~al., 2013]{socher2013reasoning}
Socher, R., Chen, D., Manning, C.~D., and Ng, A. (2013).
\newblock Reasoning with neural tensor networks for knowledge base completion.
\newblock In {\em Advances in Neural Information Processing Systems}, pages
  926--934.

\bibitem[Sutskever et~al., 2009]{sutskever2009modelling}
Sutskever, I., Tenenbaum, J.~B., and Salakhutdinov, R.~R. (2009).
\newblock Modelling relational data using bayesian clustered tensor
  factorization.
\newblock In {\em Advances in neural information processing systems}, pages
  1821--1828.

\bibitem[Trouillon et~al., 2017]{trouillon2017knowledge}
Trouillon, T., Dance, C.~R., Welbl, J., Riedel, S., Gaussier, {\'E}., and
  Bouchard, G. (2017).
\newblock Knowledge graph completion via complex tensor factorization.
\newblock {\em arXiv preprint arXiv:1702.06879}.
\newblock To appear in the Journal of Machine Learning Research.

\bibitem[Trouillon and Nickel, 2017]{trouillon2017comparison}
Trouillon, T. and Nickel, M. (2017).
\newblock Complex and holographic embeddings of knowledge graphs: a comparison.
\newblock {\em International Workshop on Statistical Relational AI}.

\bibitem[Trouillon et~al., 2016]{trouillon2016}
Trouillon, T., Welbl, J., Riedel, S., Gaussier, E., and Bouchard, G. (2016).
\newblock {Complex embeddings for simple link prediction}.
\newblock In {\em International Conference on Machine Learning}, volume~48,
  pages 2071--2080.

\bibitem[Valiant, 1984]{valiant1984theory}
Valiant, L.~G. (1984).
\newblock A theory of the learnable.
\newblock {\em Communications of the ACM}, 27(11):1134--1142.

\bibitem[Vapnik, 1995]{vapnik1995nature}
Vapnik, V.~N. (1995).
\newblock {\em The Nature of Statistical Learning Theory}.
\newblock Springer-Verlag New York, Inc.

\bibitem[Verga et~al., 2017]{verga2017generalizing}
Verga, P., Neelakantan, A., and McCallum, A. (2017).
\newblock Generalizing to unseen entities and entity pairs with row-less
  universal schema.
\newblock In {\em European Chapter of the Association of Computational
  Linguistics}.

\bibitem[Wang and Cohen, 2016]{wang2016learning}
Wang, W.~Y. and Cohen, W.~W. (2016).
\newblock Learning first-order logic embeddings via matrix factorization.
\newblock In {\em International Joint Conference on Artificial Intelligence},
  pages 2132--2138.

\bibitem[Welbl et~al., 2016]{Welbl2016}
Welbl, J., Bouchard, G., and Riedel, S. (2016).
\newblock A factorization machine framework for testing bigram embeddings in
  knowledge base completion.
\newblock In {\em Workshop on Automated Knowledge Base Construction at
  NAACL-HLT}, pages 103--107.

\bibitem[Wellman et~al., 1992]{wellman1992knowledge}
Wellman, M.~P., Breese, J.~S., and Goldman, R.~P. (1992).
\newblock From knowledge bases to decision models.
\newblock {\em The Knowledge Engineering Review}, 7(01):35--53.

\bibitem[Weston et~al., 2015]{weston2015towards}
Weston, J., Bordes, A., Chopra, S., Rush, A.~M., van Merri{\"e}nboer, B.,
  Joulin, A., and Mikolov, T. (2015).
\newblock Towards {AI}-complete question answering: A set of prerequisite toy
  tasks.
\newblock {\em arXiv preprint arXiv:1502.05698}.

\bibitem[Wikipedia, 2004]{wiki:binary_relations}
Wikipedia (2004).
\newblock Binary relation --- {W}ikipedia{,} the free encyclopedia.
\newblock \url{https://en.wikipedia.org/wiki/Binary_relation}.

\bibitem[Yang et~al., 2015]{Yang2015}
Yang, B., Yih, W.-t., He, X., Gao, J., and Deng, L. (2015).
\newblock {Embedding entities and relations for learning and inference in
  knowledge bases}.
\newblock In {\em International Conference on Learning Representations}.

\bibitem[Yoon et~al., 2016]{yoon2016translation}
Yoon, H.-G., Song, H.-J., Park, S.-B., and Park, S.-Y. (2016).
\newblock A translation-based knowledge graph embedding preserving logical
  property of relations.
\newblock In {\em North American Chapter of the Association of Computational
  Linguistics: Human Language Technologies}, pages 907--916.

\end{thebibliography}

\appendix
\newpage

\section{Learning Algorithm}
\label{app:algo}

\Cref{SGDC} describes the stochastic gradient descent algorithm
used to learn 
the evaluated models, with the AdaGrad learning-rate updates
\cite{duchi2011adaptive}. The parameters are 
initialized from a zero-mean normal distribution with unit
variance.
Squared gradients are accumulated to compute AdaGrad learning rates,
then gradients are updated.
Every $s$ iterations, the parameters $\Theta$ are evaluated
over the evaluation set $\Omega_v$,
through the {\it evaluate\_AP}$(\Omega_v;\Theta)$ function.
The optimization process is stopped when 
average precision decreases compared to the last evaluation (early stopping).
The {\it sample\_batch\_of\_size\_b}$(\Omega,b)$
function sample uniformly $b$ true and false triples uniformly at random
from the training set $\Omega$.

\begin{algorithm}[ht]
\caption{Stochastic gradient descent with AdaGrad}
\label{SGDC}
\begin{algorithmic}
\INPUT Training set $\Omega$, validation set $\Omega_v$, learning rate $\alpha\in \mathbb{R}_{++}$, rank $K\in \mathbb{Z}_{++}$, $L^2$ regularization factor $\lambda\in \mathbb{R}_{+}$, batch size $b\in \mathbb{Z}_{++}$, maximum iteration $m\in \mathbb{Z}_{++}$, validate every $s\in \mathbb{Z}_{++}$ iterations, AdaGrad regularizer $\epsilon = 10^{-8}$.
\OUTPUT Trained embeddings $\Theta$.
\STATE $\Theta_i\sim \mathcal{N}(\mathbf{0}^k, I^{k \times k})$ for each $i \in \mathcal{E}$
\STATE $\Theta_r \sim \mathcal{N}(\mathbf{0}^k, I^{k \times k})$ for each $r \in \mathcal{R}$
\STATE $g_{\Theta_i} \gets \mathbf{0}^k$  for each $i \in \mathcal{E}$
\STATE $g_{\Theta_r} \gets \mathbf{0}^k$  for each $r \in \mathcal{R}$
\STATE $previous\_score \gets 0$
\FOR{$i=1,\ldots,m$}
    \FOR {$j=1,\ldots,|\Omega|/b$}
        \STATE $\Omega_b \gets$ {\it sample\_batch\_of\_size\_b}$(\Omega,b)$
        \FOR {$((r,s,o),y_{rso})$ in $\Omega_b$}
            \FOR{$v$ in $\Theta$}
                \STATE // AdaGrad updates:
                \STATE $g_v \gets g_v + (\nabla_v \mathcal{L}(\{((r,s,o),y_{rso})\};\Theta))^2$
                \STATE // Gradient updates:
                \STATE $v \gets v - \frac{\alpha}{g_v + \epsilon} \nabla_v \mathcal{L}(\{((r,s,o),y_{rso})\};\Theta)$
            \ENDFOR
        \ENDFOR
    \ENDFOR
    \STATE // Early stopping
    \IF{$i \mod s = 0$}
        \STATE $current\_score \gets$ {\it evaluate\_AP}$(\Omega_v;\Theta)$ 
            \IF{$current\_score \leq previous\_score$}
                \STATE \textbf{break} 
            \ENDIF
            \STATE $previous\_score \gets current\_score$
    \ENDIF
\ENDFOR
\STATE \textbf{return} $\Theta$
\end{algorithmic}
\end{algorithm}

\section{Results with Reflexivity and Irreflexivity}
\label{app:refl_irrefl}

\begin{figure}[ht]
	\centering
	\includegraphics[width=0.422\linewidth]{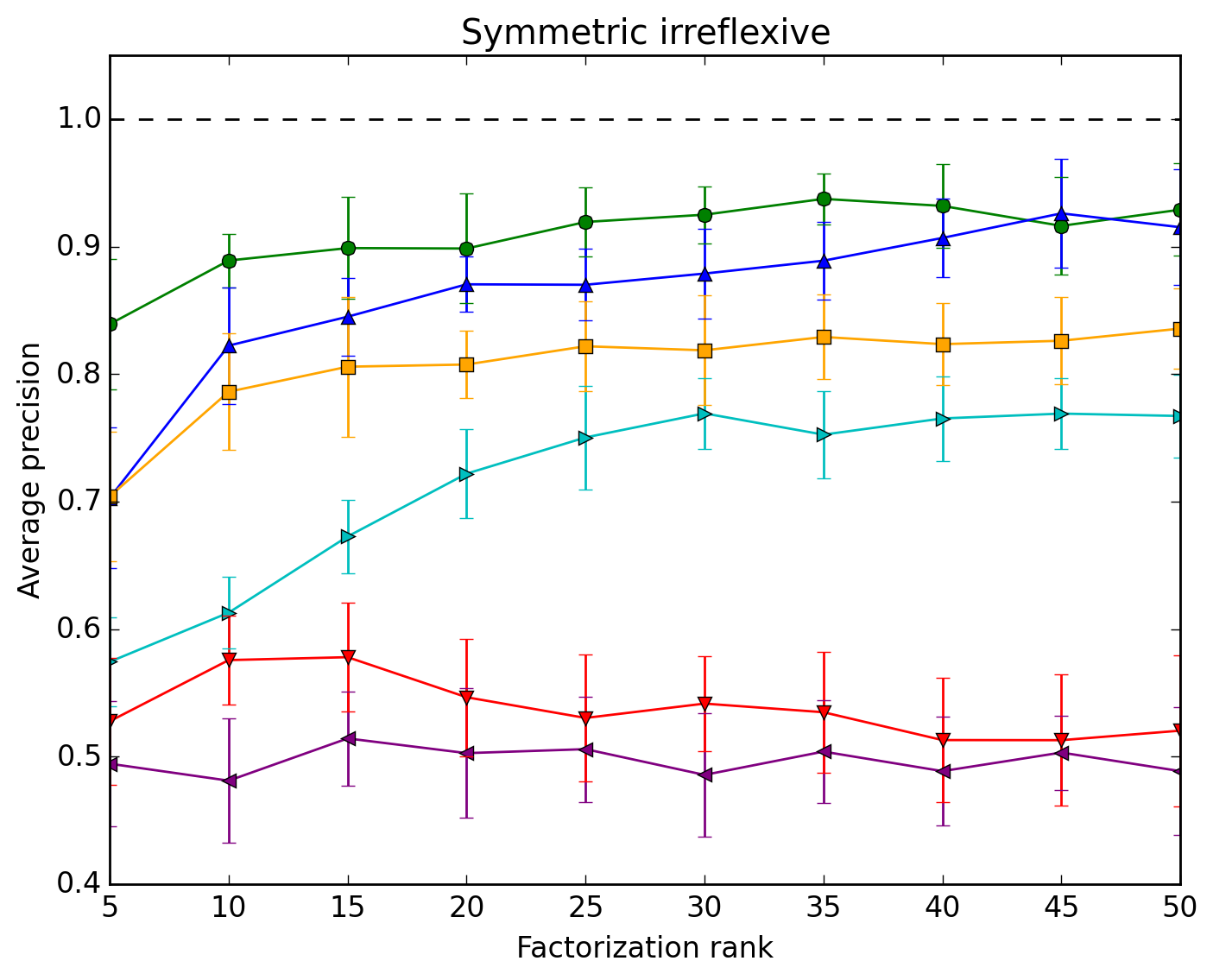}
	\includegraphics[width=0.56\linewidth]{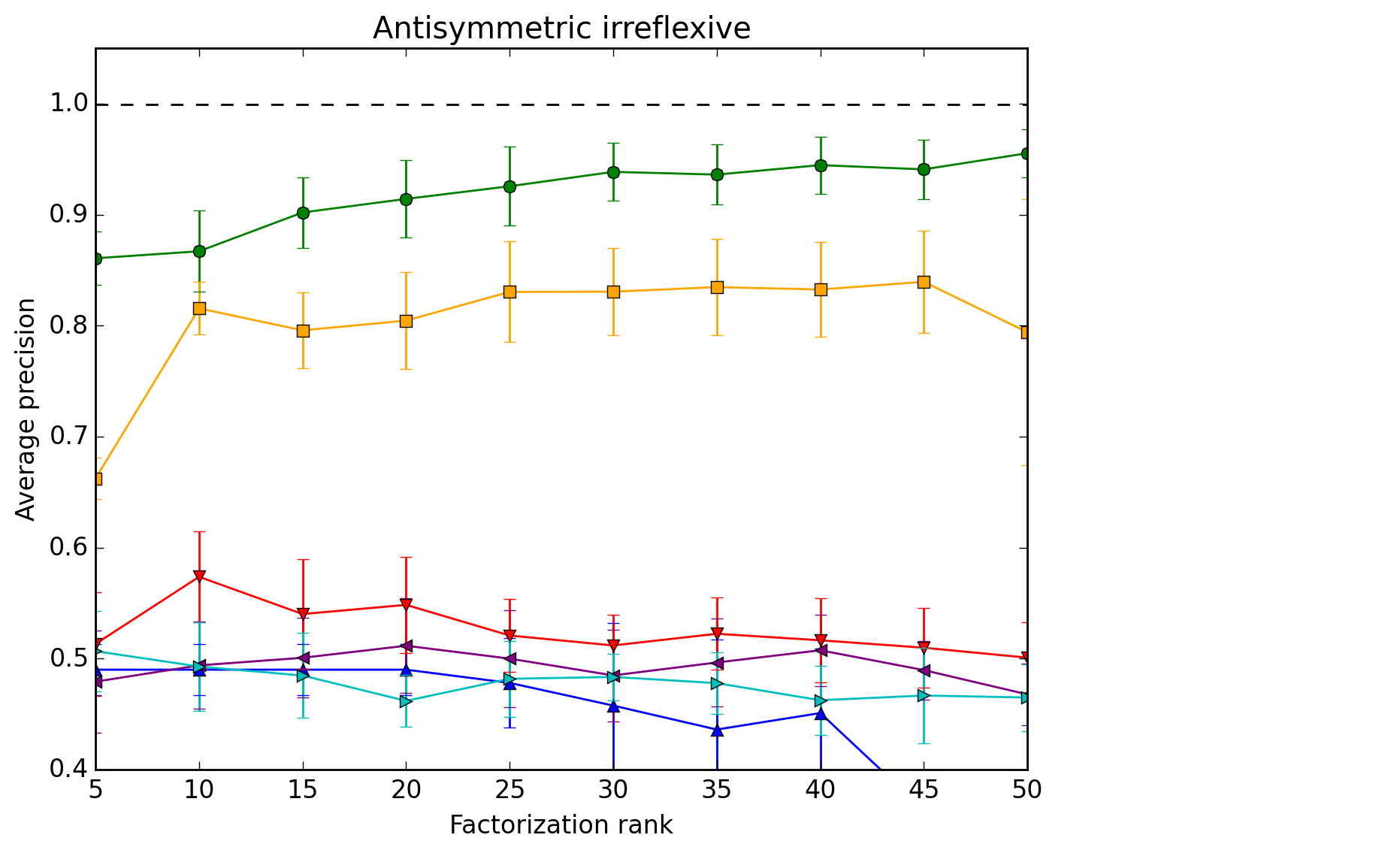}\\
	\includegraphics[width=0.56\linewidth]{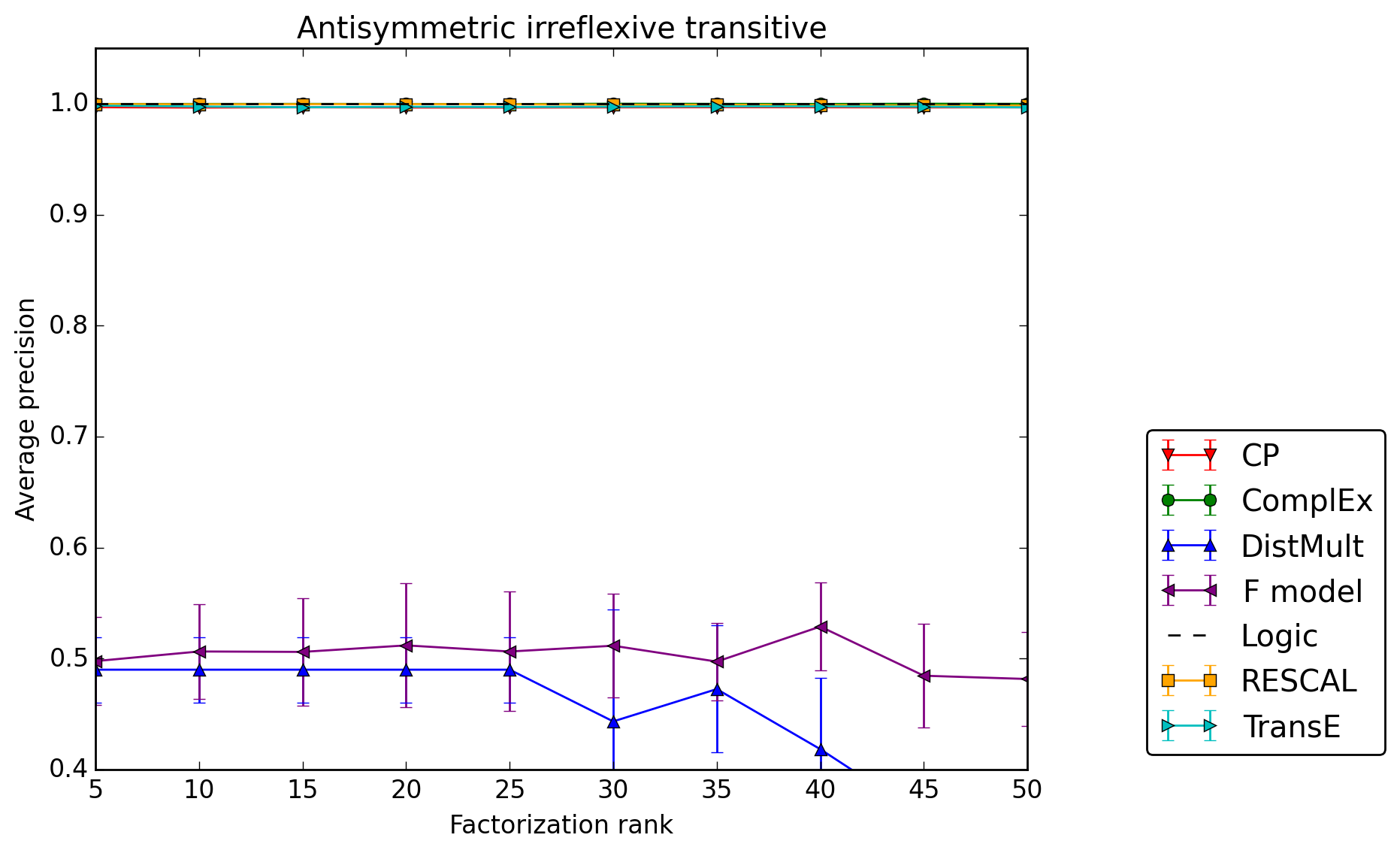}
	\caption{Generated irreflexive relations with 50 entities, combined with symmetry (top-left),
	antisymmetry (top-right) and antisymmetry and transitivity (bottom). 
	Average precision for each rank ranging from 5 to 50 for each model.}
	\label{fig:exp_irrefl}
\end{figure}

\begin{figure}[ht]
	\centering
	\includegraphics[width=0.422\linewidth]{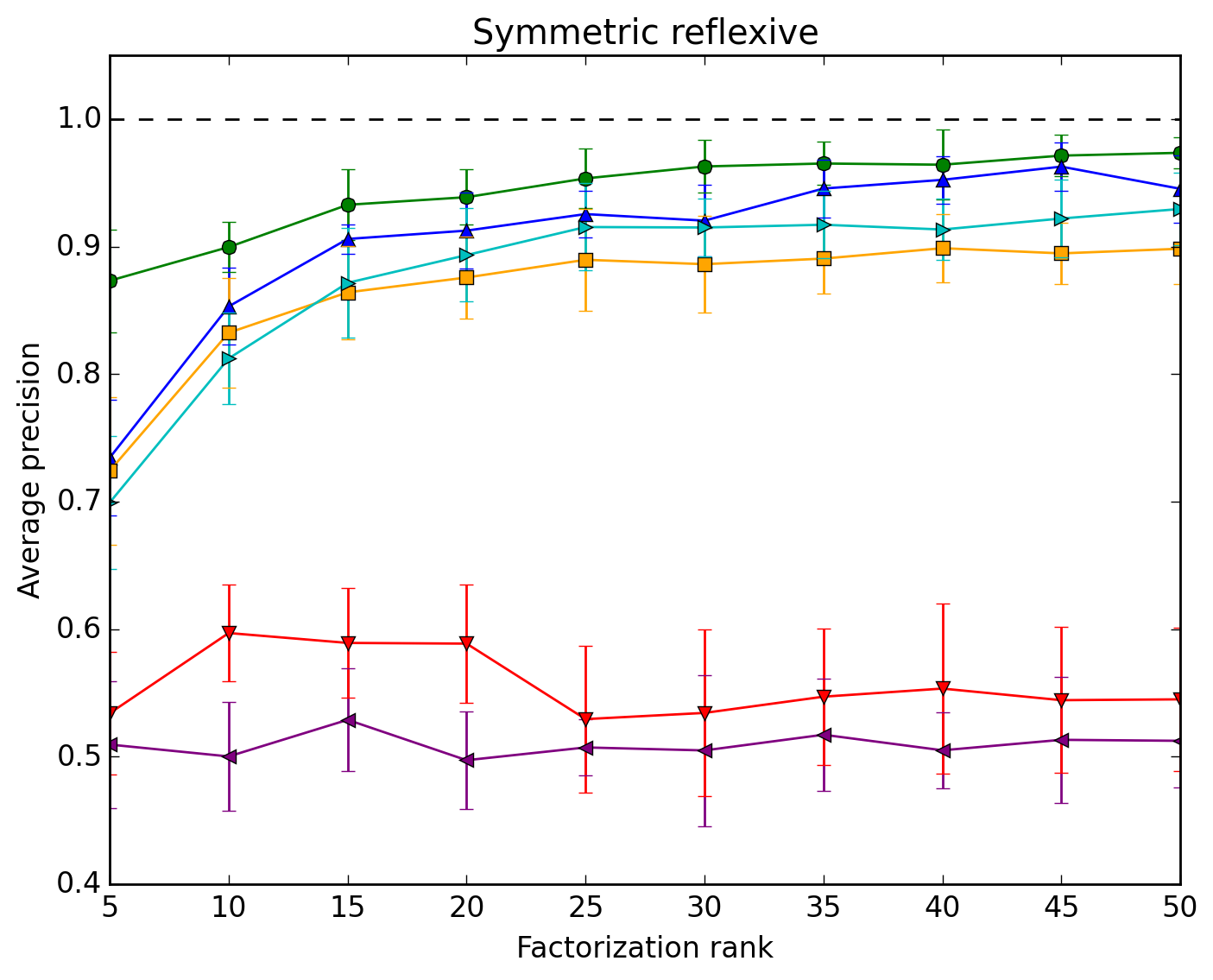}
	\includegraphics[width=0.56\linewidth]{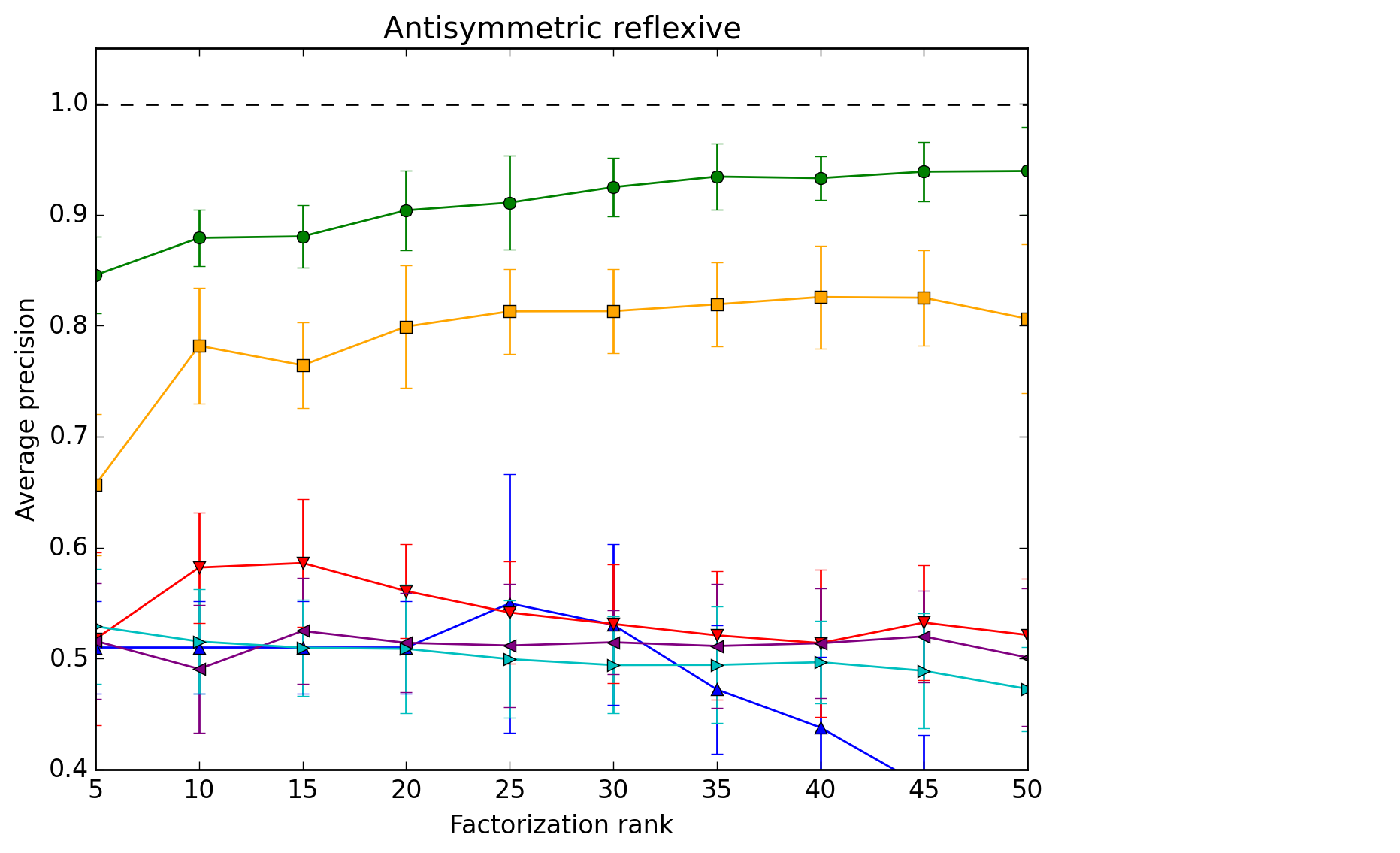}\\
	\includegraphics[width=0.56\linewidth]{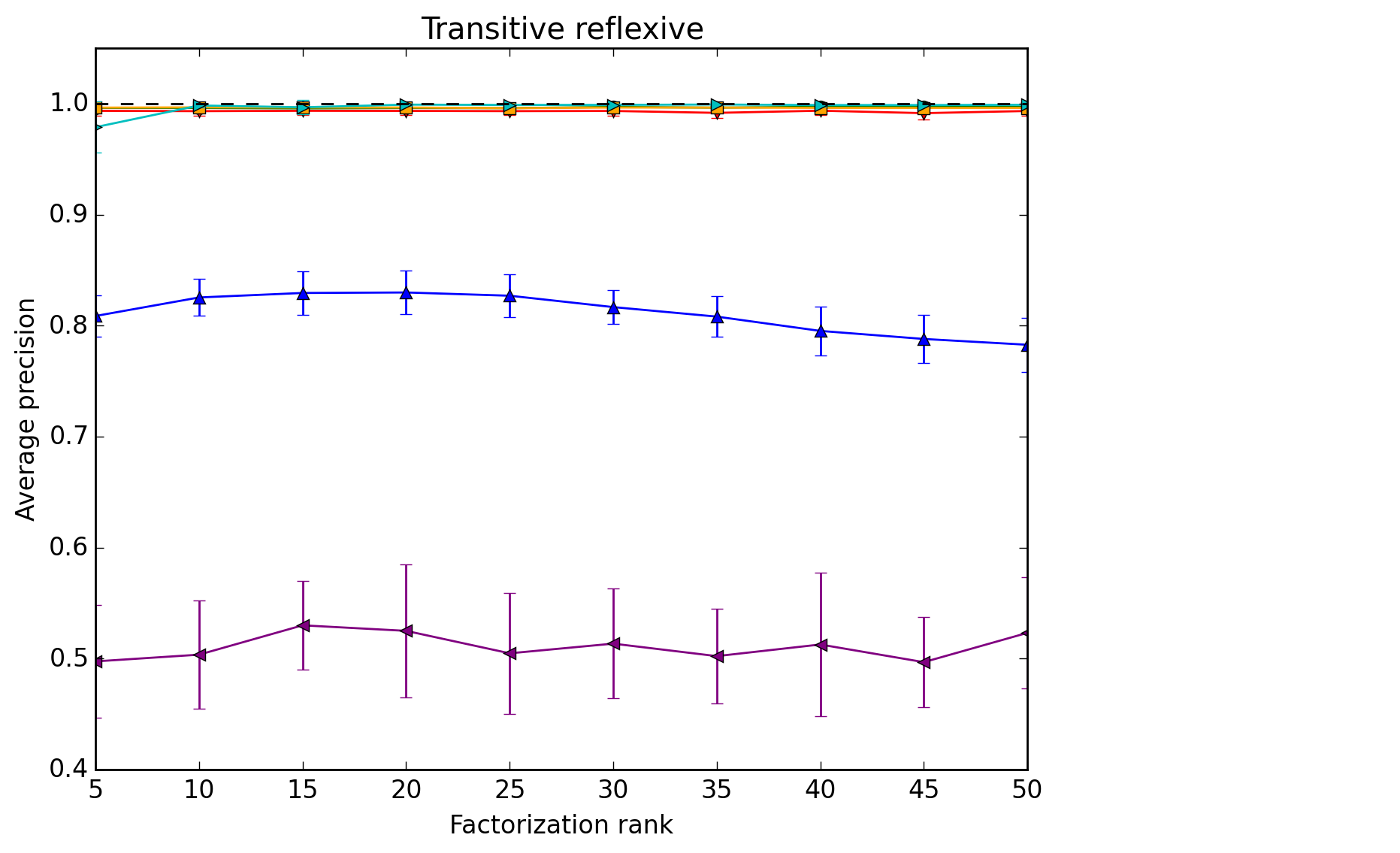}\\
	\includegraphics[width=0.422\linewidth]{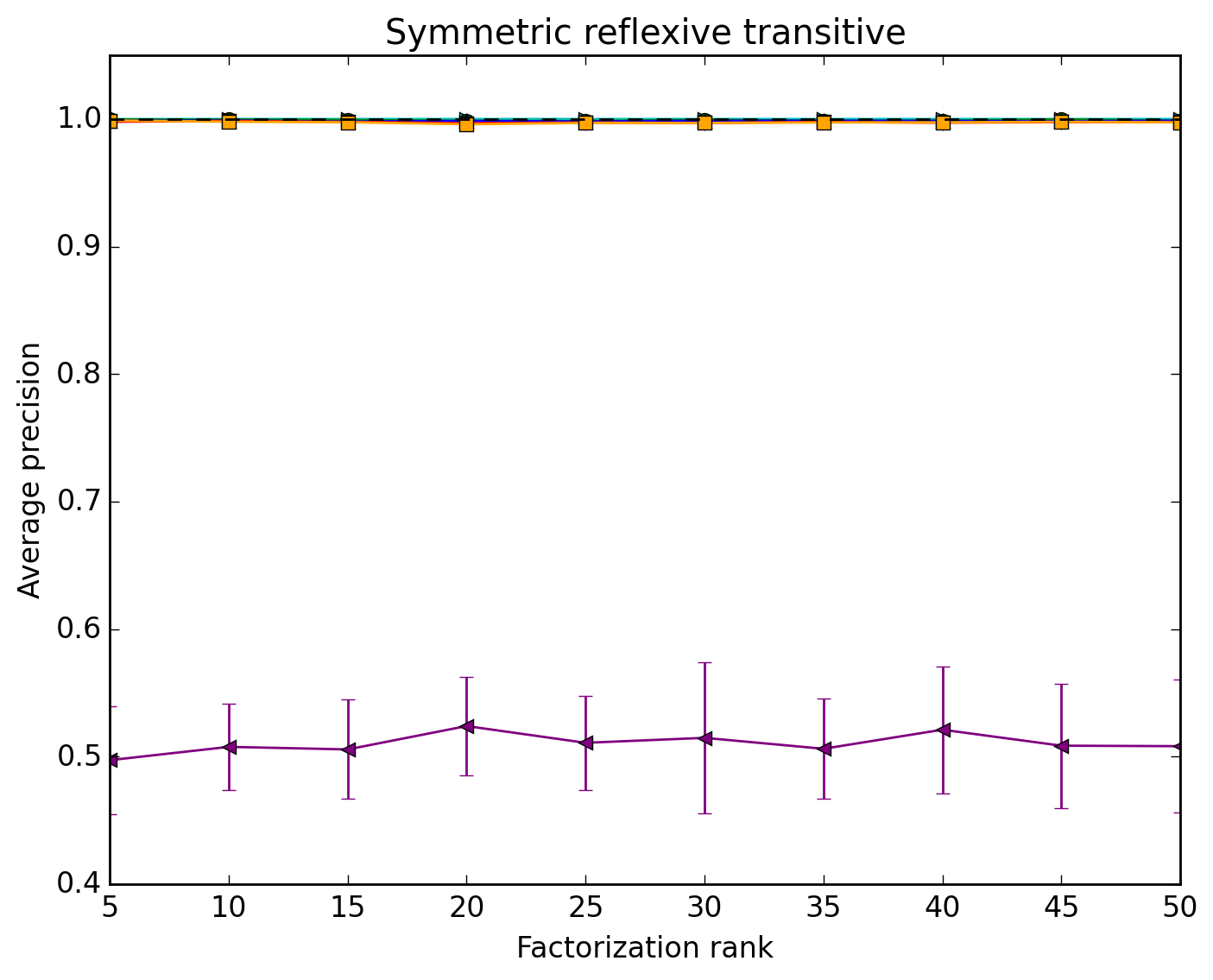}
	\includegraphics[width=0.56\linewidth]{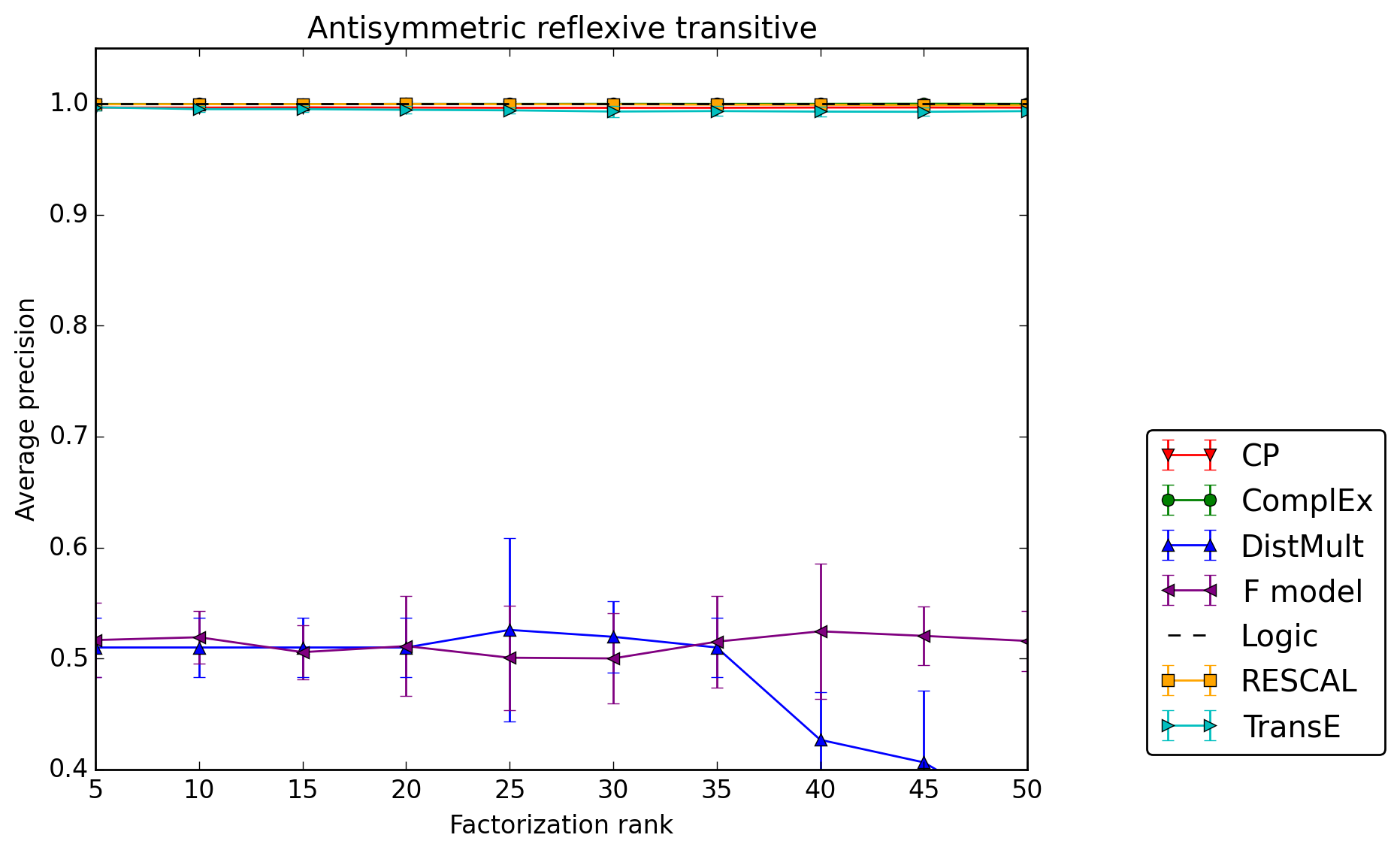}
	\caption{Generated reflexive relations with 50 entities, combined with symmetry (top-left),
	antisymmetry (top-right), transitivity (middle), symmetry and transitivity (bottom-left) and
	antisymmetry and transitivity (bottom-right). Average precision for each rank ranging from 5 to 50 for each model.}
	\label{fig:exp_refl}
\end{figure}

In this appendix we report results of the individual learning
of combinations of relation properties including reflexivity
and irreflexivity. Those results are included for completeness
as they are similar to the cases that are neither reflexive
nor irreflexive, reported in \Cref{sec:rel_props_indiv_res}.
\Cref{fig:exp_refl} shows results for the 5 combinations
with reflexivity, and \Cref{fig:exp_irrefl} for the 3 combinations
with irreflexivity. The irreflexive transitive case, and the irreflexive symmetric transitive case
are not reported as they are not consistent, as explained in \Cref{sec:rel_props_design}.
The single noticeable difference is in the symmetric
irreflexive case, where all models perform slightly worse compared 
to the symmetric and symmetric reflexive cases, especially \textsc{TransE}.

\end{document}